%% file: main.tex
\documentclass{article}

\usepackage{preprint,times}
\usepackage[T1]{fontenc}
\usepackage{amsmath,amssymb}
\usepackage{graphicx}
\usepackage{booktabs,tabularx,array,threeparttable}
\usepackage{placeins}
\usepackage{xcolor}
\usepackage{hyperref}
\usepackage{newtxtext}
\usepackage{newtxmath}
\usepackage{microtype}

\fussy
\definecolor{linkblue}{RGB}{0,40,110}
\hypersetup{hypertexnames=false,colorlinks=true,
  linkcolor=linkblue,citecolor=linkblue,urlcolor=linkblue,
  pdftitle={Fine Until Fine-Tuned Further: Repeated Solutions Make a Reasoning Adapter Fragile},
  pdfauthor={Ely Sheikh}}
\input{numbers.generated}
\input{numbers.revision2}

\newcommand{\think}{\texttt{\textbackslash n<think>\textbackslash n}}
\newcommand{\figfile}[2]{\includegraphics[width=#2]{#1}}

\title{Fine Until Fine-Tuned Further: Repeated Solutions\\Make a Reasoning Adapter Fragile}

\author{Ely Sheikh\\
  Aix-Marseille University\\
  \texttt{ely.sheikh1@gmail.com}}

\finalcopy
\begin{document}

\maketitle
\lhead{Preprint}
\suppressfloats[t]   

\begin{abstract}
Recipes such as s1 and LIMO teach a model to reason from a thousand worked solutions
or fewer, shown to it many times over. Judged when that training ends, the
repetition costs nothing. We study reasoning trained into a LoRA adapter, on a skill
the model already has, and find a cost that appears when the same adapter is trained
further: the reasoning becomes fragile, even to training on data unrelated to
reasoning. We fine-tuned Qwen3.5-9B-Base on its own correct solutions in three ways
that end equally accurate, at about 95\% of held-out competition problems. A drilled
model saw a few hundred solutions about eight times each, a once-trained model many
more once each, and a fresh-solution model saw the drilled problems as often but a new solution each time. After one further epoch of instruction tuning on the same
adapter, the drilled model had lost \nv{rs.ex.D} points more than the once-trained
one. The fresh-solution model, which saw the same few problems, lost nothing, so the
break comes from repeating the texts. On the same adapter the break recurs
across runs, a stronger model's traces, a 35B model, another model family and an
arithmetic task. Merging the adapter into the base weights and training a new one
protects the drilled model through that epoch. The lost reasoning is suppressed, not
erased: five updates of reasoning training bring nearly all of it back.
\end{abstract}

\input{sections/introduction}
\input{sections/setup}
\input{sections/fragile}
\input{sections/released}
\input{sections/signature}
\input{sections/suppressed}
\input{sections/remedies}
\input{sections/related}
\input{sections/discussion}

\input{sections/statements}

\subsubsection*{Acknowledgments}
We thank Thinking Machines Lab for a \$5,000 Tinker Research Grant, which supported the
compute for this work. We used an AI system (Claude, Anthropic) as a tool to write the code and
to draft parts of the text under our direction, and we take full responsibility for both.

\FloatBarrier
\bibliographystyle{preprint}
\bibliography{references}

\appendix
\input{sections/app-a-training}

\input{sections/app-b-results}
\input{sections/app-c-released}

\input{sections/app-d-failure}

\input{sections/app-e-divergence}

\input{sections/app-f-replay}

\input{sections/app-g-skill}

\input{sections/app-h-arithmetic}

\input{sections/app-i-ledger}
\end{document}

%% file: numbers.generated.tex
\makeatletter
\newcommand{\defnum}[2]{\expandafter\def\csname nv@#1\endcsname{#2}}
\newcommand{\nv}[1]{\ifcsname nv@#1\endcsname\csname nv@#1\endcsname\else\textbf{??}\PackageWarning{numbers}{Undefined number #1}\fi}
\makeatother

\defnum{d.rank}{32} 
\defnum{d.rank.alt}{128} 
\defnum{d.items}{192} 
\defnum{d.budget.xs}{2,048} 
\defnum{d.budget.short}{4,096} 
\defnum{d.budget.mid}{8,192} 
\defnum{d.budget.long}{16,384} 
\defnum{d.budget.aime}{32,768} 
\defnum{d.topp}{0.95} 
\defnum{d.T.updates}{30} 
\defnum{d.S.samples}{6,400} 
\defnum{d.S.perprompt}{8} 
\defnum{d.S.updates}{230} 
\defnum{d.S.passes}{1.15} 
\defnum{d.S.presentations}{9.2} 
\defnum{d.drill.n}{579} 
\defnum{d.drill.updates}{140} 
\defnum{d.upd.low}{20} 
\defnum{d.upd.mid}{60} 
\defnum{d.pass.low}{1.1} 
\defnum{d.pass.mid}{3.3} 
\defnum{d.pass.high}{7.7} 
\defnum{d.O.n}{4,480} 
\defnum{d.O.pass.mid}{0.43} 
\defnum{d.O.pass.high}{1.0} 
\defnum{d.P.draws}{24} 
\defnum{d.P.kept}{8} 
\defnum{d.P.passes}{1.0} 
\defnum{d.R.fit}{20} 
\defnum{d.RS.passes}{11.1} 
\defnum{mt.drill.req}{0.8} 
\defnum{mt.held.req}{20} 
\defnum{mt.p5.band}{0.01} 
\defnum{d.R2.updates}{60} 
\defnum{d.R1.match}{0.1} 
\defnum{d.R5.rows}{1,920} 
\defnum{d.acq.lr}{$10^{-4}$} 
\defnum{d.stage.lr}{$3\times10^{-4}$} 
\defnum{d.stage.lrlow}{$10^{-4}$} 
\defnum{d.stage.batch}{32} 
\defnum{d.stage.updates}{20} 
\defnum{d.norobots}{640} 
\defnum{d.anchor.extra}{32} 
\defnum{d.anchor.whigh}{1} 
\defnum{d.anchor.wlow}{0.25} 
\defnum{d.anchor.fit}{80} 
\defnum{d.anchor.parity}{5} 
\defnum{d.parity.math}{5} 
\defnum{d.math.pool}{36,447} 
\defnum{d.math.distinct}{58,674} 
\defnum{d.math.eval}{221} 
\defnum{d.math.evalall}{311} 
\defnum{d.math.aime}{59} 
\defnum{d.math.easy}{256} 
\defnum{d.math.cut}{2,048} 
\defnum{d.tces.cut}{1,024} 
\defnum{d.decision.tokens}{8} 
\defnum{d.horizon}{480} 
\defnum{d.active}{3} 
\defnum{d.reforce.times}{8} 
\defnum{d.reforce.tokens}{1,024} 
\defnum{d.RP.replay}{6.25} 
\defnum{d.RL.n}{640} 
\defnum{d.SH.temp}{0.5} 
\defnum{d.SH.draws}{4} 
\defnum{d.SK.cap}{2,048} 
\defnum{d.SK.items}{300} 
\defnum{d.SK.smoke}{5} 
\defnum{d.SD.shared}{91} 
\defnum{r4.req}{10} 
\defnum{r5.req}{5} 
\defnum{r5.req.sec}{10} 
\defnum{reg.MP1.thr}{5} 
\defnum{reg.M3.2.thr}{20} 
\defnum{reg.M2.1.thr}{10} 
\defnum{reg.T1.thr}{10} 
\defnum{reg.T2.thr}{3} 
\defnum{reg.T2.thr2}{10} 
\defnum{reg.T4.thr}{5} 
\defnum{reg.T5.thr}{10} 
\defnum{reg.R1.1.thr}{2} 
\defnum{reg.R1.2.thr}{77.4} 
\defnum{reg.R1.3.thr}{1.5} 
\defnum{reg.R1.4.thr}{10} 
\defnum{reg.R2.1.thr}{10} 
\defnum{reg.R3a.thr}{20} 
\defnum{reg.R3b.thr}{2} 
\defnum{reg.DP4.thr}{0.7} 
\defnum{reg.CT1.thr}{15} 
\defnum{reg.CT2.thr}{10} 
\defnum{reg.AN1.thr}{40} 
\defnum{reg.AN3.thr}{20} 
\defnum{reg.AN3.thr2}{0.06} 
\defnum{reg.AN4.thr}{20} 
\defnum{reg.C-3.thr}{10} 
\defnum{reg.C2.thr}{10} 
\defnum{reg.C3.thr}{3--6} 
\defnum{reg.C3.thr2}{1} 
\defnum{reg.RR-D3b.thr}{10} 
\defnum{reg.RR-D3b.thr2}{3} 
\defnum{reg.RR-D4.thr}{30} 
\defnum{reg.RL1.thr}{0.5} 
\defnum{reg.RL2.thr}{2} 
\defnum{reg.RL.suppressed}{0.75} 
\defnum{reg.RL.habit}{0.5} 
\defnum{reg.RP.prevent}{3} 
\defnum{reg.RP.upper}{6} 
\defnum{reg.RP.mitigate}{50} 
\defnum{reg.RS.D}{5} 
\defnum{reg.RS.P}{3} 
\defnum{reg.SH.gate}{0.05} 
\defnum{reg.SH.within}{3} 
\defnum{reg.SH.half}{36.5} 
\defnum{reg.SK.cal}{25} 
\defnum{reg.SK.gate}{30} 
\defnum{reg.SK.break}{10} 
\defnum{reg.SK.within}{3} 
\defnum{lit.kop.n}{400} 
\defnum{lit.kop.ratio}{128} 
\defnum{lit.s1.epochs}{5} 
\defnum{lit.limo.epochs}{15} 
\defnum{m.calib.direct}{83} 
\defnum{m.scorer.units}{30} 
\defnum{m.scorer.hand}{20} 
\defnum{aime.ci}{13--16} 
\defnum{m35.pre.mad}{0.0034} 
\defnum{m35.pre.two}{0.0030} 
\defnum{m35.pre.thr}{0.001} 
\defnum{m35.noise.mad}{0.0003} 
\defnum{mnem.pre.pos}{1,032} 
\defnum{mnem.tmpl.n}{7,047} 
\defnum{anc.samples.capped}{55} 
\defnum{mt.traces.P}{7.99} 
\defnum{r5.wordprob}{1--1.5} 
\defnum{comp.total}{2,437} 
\defnum{sk.text.mean}{871} 
\defnum{sk.text.max}{1,937} 
\defnum{reg.total}{70} 
\defnum{reg.held}{50} 
\defnum{reg.failed}{17} 
\defnum{reg.failed.partial}{7} 
\defnum{reg.unread}{3} 
\defnum{reg.math}{44} 
\defnum{reg.arith}{26} 
\defnum{reg.failed.arith}{12} 
\defnum{cal.l12.forced}{1.00} 
\defnum{cal.l12.free}{1.00} 
\defnum{cal.l12.direct}{0.58} 
\defnum{cal.l12.med}{850} 
\defnum{cal.l3.forced}{0.92} 
\defnum{cal.l3.free}{0.83} 
\defnum{cal.l3.direct}{0.42} 
\defnum{cal.l3.med}{1,326} 
\defnum{cal.l4.forced}{0.88} 
\defnum{cal.l4.free}{0.71} 
\defnum{cal.l4.direct}{0.17} 
\defnum{cal.l4.med}{954} 
\defnum{cal.l5.forced}{0.62} 
\defnum{cal.l5.free}{0.58} 
\defnum{cal.l5.direct}{0.08} 
\defnum{cal.l5.med}{2,928} 
\defnum{cal.oly.forced}{0.46} 
\defnum{cal.oly.free}{0.42} 
\defnum{cal.oly.direct}{0.00} 
\defnum{cal.oly.med}{6,144 (cap)} 
\defnum{cal.num.forced}{0.73} 
\defnum{cal.num.free}{0.67} 
\defnum{cal.num.direct}{0.19} 
\defnum{cal.num.med}{1,410} 
\defnum{tces4.M0.u0}{54.9} 
\defnum{tces4s.M0.u0}{13.8} 
\defnum{tces4.M0.u20}{45.1} 
\defnum{tces4s.M0.u20}{9.9} 
\defnum{tces4.M0.u80}{39.3} 
\defnum{tces4r.M0.u0}{52} 
\defnum{tces4free.M0.u0}{13.5} 
\defnum{tces4.T.u0}{62.2} 
\defnum{tces4s.T.u0}{22.1} 
\defnum{tces4.T.u20}{59.4} 
\defnum{tces4s.T.u20}{22.9} 
\defnum{tces4.T.u80}{21.4} 
\defnum{tces4r.T.u0}{48} 
\defnum{tces4free.T.u0}{44.0} 
\defnum{tces4.S.u0}{57.0} 
\defnum{tces4s.S.u0}{14.6} 
\defnum{tces4.S.u20}{57.6} 
\defnum{tces4s.S.u20}{8.6} 
\defnum{tces4r.S.u0}{51} 
\defnum{tces4free.S.u0}{37.8} 
\defnum{tces4.RP.u0}{61.2} 
\defnum{tces4s.RP.u0}{51.8} 
\defnum{tces4.RP.u20}{2.6} 
\defnum{tces4s.RP.u20}{0.5} 
\defnum{tces4r.RP.u0}{12} 
\defnum{tces4free.RP.u0}{43.2} 
\defnum{tces4.RS.u0}{58.9} 
\defnum{tces4s.RS.u0}{34.6} 
\defnum{tces4.RS.u20}{40.6} 
\defnum{tces4s.RS.u20}{22.7} 
\defnum{tces4r.RS.u0}{35} 
\defnum{tces4free.RS.u0}{39.3} 
\defnum{tces4.U20.u0}{59.9} 
\defnum{tces4s.U20.u0}{10.4} 
\defnum{tces4.U20.u20}{58.1} 
\defnum{tces4s.U20.u20}{25.3} 
\defnum{tces4r.U20.u0}{61} 
\defnum{tces4free.U20.u0}{43.8} 
\defnum{tces4.F20.u0}{58.6} 
\defnum{tces4s.F20.u0}{22.9} 
\defnum{tces4.F20.u20}{62.5} 
\defnum{tces4s.F20.u20}{22.1} 
\defnum{tces4r.F20.u0}{43} 
\defnum{tces4free.F20.u0}{51.8} 
\defnum{tces4.U60.u0}{64.8} 
\defnum{tces4s.U60.u0}{14.1} 
\defnum{tces4.U60.u20}{46.4} 
\defnum{tces4s.U60.u20}{7.6} 
\defnum{tces4r.U60.u0}{62} 
\defnum{tces4free.U60.u0}{46.1} 
\defnum{tces4.F60.u0}{66.1} 
\defnum{tces4s.F60.u0}{39.6} 
\defnum{tces4.F60.u20}{47.9} 
\defnum{tces4s.F60.u20}{28.4} 
\defnum{tces4r.F60.u0}{30} 
\defnum{tces4free.F60.u0}{65.6} 
\defnum{tces4.U140.u0}{53.6} 
\defnum{tces4s.U140.u0}{12.8} 
\defnum{tces4.U140.u20}{0.3} 
\defnum{tces4s.U140.u20}{0.3} 
\defnum{tces4r.U140.u0}{51} 
\defnum{tces4free.U140.u0}{33.9} 
\defnum{tces4.F140.u0}{61.7} 
\defnum{tces4s.F140.u0}{52.9} 
\defnum{tces4.F140.u20}{3.1} 
\defnum{tces4s.F140.u20}{2.3} 
\defnum{tces4r.F140.u0}{11} 
\defnum{tces4free.F140.u0}{55.7} 
\defnum{tces4s.handoff.lasting.max}{22.9} 
\defnum{tces4s.handoff.lasting.min}{10.4} 
\defnum{tces4.dose.M0}{$-$9.9} 
\defnum{tces4.dose.T}{$-$2.9} 
\defnum{tces4.dose.S}{+0.5} 
\defnum{tces4.dose.U.low}{$-$1.8} 
\defnum{tces4.dose.U.mid}{$-$18.5} 
\defnum{tces4.dose.U.high}{$-$53.4} 
\defnum{tces4.dose.F.low}{+3.9} 
\defnum{tces4.dose.F.mid}{$-$18.2} 
\defnum{tces4.dose.F.high}{$-$58.6} 
\defnum{tces4loss.RS}{18.2} 
\defnum{tces4loss.U60}{18.5} 
\defnum{tces4loss.F60}{18.2} 
\defnum{tces4g.U20}{61.5} 
\defnum{tces4gs.U20}{10.4} 
\defnum{tces4g.F20}{60.9} 
\defnum{tces4gs.F20}{21.6} 
\defnum{tces4g.U140}{29.9} 
\defnum{tces4gs.U140}{6.8} 
\defnum{tces4g.F140}{37.2} 
\defnum{tces4gs.F140}{28.1} 
\defnum{tces4i.M0}{53.6} 
\defnum{tces4is.M0}{26.3} 
\defnum{tces4i.U20}{58.9} 
\defnum{tces4is.U20}{26.8} 
\defnum{tces4i.F20}{57.8} 
\defnum{tces4is.F20}{37.0} 
\defnum{tces4i.U140}{34.9} 
\defnum{tces4is.U140}{27.6} 
\defnum{tces4i.F140}{36.7} 
\defnum{tces4is.F140}{31.0} 
\defnum{tces4ir.M0}{35} 
\defnum{tces4i.keep.max}{58.9} 
\defnum{tces4i.keep.min}{53.6} 
\defnum{tces4i.median.max}{139} 
\defnum{tces4i.median.min}{119} 
\defnum{reg.RR-D3b.obs}{37.8--41.7} 
\defnum{tces4.dmo.U.b}{$-$51.6} 
\defnum{tces4.dmo.U.b.ci}{[$-$60.4, $-$42.7]} 
\defnum{tces4.dmo.U.g}{$-$25.3} 
\defnum{tces4.dmo.U.i}{$-$17.7} 
\defnum{tces4.dmo.F.b}{$-$62.5} 
\defnum{tces4.dmo.F.b.ci}{[$-$70.1, $-$55.2]} 
\defnum{tces4.dmo.F.g}{$-$26.8} 
\defnum{tces4.dmo.F.i}{$-$24.2} 
\defnum{tces4.once.maxloss}{1.0} 
\defnum{tces4.cliff.F140}{45.1} 
\defnum{tces4.rerun.S}{41.9} 
\defnum{tces4.rerun.T}{53.9} 
\defnum{tces16.M0.u0}{77.6} 
\defnum{tces16s.M0.u0}{19.3} 
\defnum{tces16r.M0.u0}{67} 
\defnum{tces16w4.M0.u0}{50.0} 
\defnum{tces16w8.M0.u0}{65.1} 
\defnum{tces16.M0.u20}{73.4} 
\defnum{tces16s.M0.u20}{18.8} 
\defnum{tces16r.M0.u20}{61} 
\defnum{tces16w4.M0.u20}{50.8} 
\defnum{tces16w8.M0.u20}{64.8} 
\defnum{tces16cap.M0.u20}{22.1} 
\defnum{tces16over4.M0.u20}{46} 
\defnum{tces16.T.u0}{78.1} 
\defnum{tces16s.T.u0}{22.9} 
\defnum{tces16r.T.u0}{63} 
\defnum{tces16w4.T.u0}{58.6} 
\defnum{tces16w8.T.u0}{71.6} 
\defnum{tces16.T.u20}{74.7} 
\defnum{tces16s.T.u20}{34.6} 
\defnum{tces16r.T.u20}{45} 
\defnum{tces16w4.T.u20}{55.5} 
\defnum{tces16w8.T.u20}{65.6} 
\defnum{tces16cap.T.u20}{18.0} 
\defnum{tces16over4.T.u20}{41} 
\defnum{tces16.S.u0}{77.9} 
\defnum{tces16s.S.u0}{26.3} 
\defnum{tces16r.S.u0}{59} 
\defnum{tces16w4.S.u0}{54.2} 
\defnum{tces16w8.S.u0}{68.8} 
\defnum{tces16.S.u20}{75.8} 
\defnum{tces16s.S.u20}{10.9} 
\defnum{tces16r.S.u20}{81} 
\defnum{tces16w4.S.u20}{54.2} 
\defnum{tces16w8.S.u20}{65.6} 
\defnum{tces16cap.S.u20}{16.1} 
\defnum{tces16over4.S.u20}{41} 
\defnum{tces16.U140.u0}{75.5} 
\defnum{tces16s.U140.u0}{17.4} 
\defnum{tces16r.U140.u0}{66} 
\defnum{tces16w4.U140.u0}{52.9} 
\defnum{tces16w8.U140.u0}{65.4} 
\defnum{tces16.U140.u20}{6.0} 
\defnum{tces16s.U140.u20}{0.8} 
\defnum{tces16r.U140.u20}{35} 
\defnum{tces16w4.U140.u20}{1.8} 
\defnum{tces16w8.U140.u20}{2.9} 
\defnum{tces16cap.U140.u20}{93.2} 
\defnum{tces16over4.U140.u20}{98} 
\defnum{tces16.F140.u0}{81.5} 
\defnum{tces16s.F140.u0}{68.5} 
\defnum{tces16r.F140.u0}{14} 
\defnum{tces16w4.F140.u0}{60.4} 
\defnum{tces16w8.F140.u0}{73.4} 
\defnum{tces16.F140.u20}{11.7} 
\defnum{tces16s.F140.u20}{4.9} 
\defnum{tces16r.F140.u20}{20} 
\defnum{tces16w4.F140.u20}{4.2} 
\defnum{tces16w8.F140.u20}{7.3} 
\defnum{tces16cap.F140.u20}{87.5} 
\defnum{tces16over4.F140.u20}{95} 
\defnum{tces16.draws}{384} 
\defnum{tces16capn.U140}{358} 
\defnum{tces16scan.U140}{7} 
\defnum{tces16capn.F140}{336} 
\defnum{tces16scan.F140}{9} 
\defnum{tces16loss.U140}{69.5} 
\defnum{tces16loss.F140}{69.8} 
\defnum{tces16ex.U140}{67.4} 
\defnum{tces16ex.U140.ci}{[60.9, 74.0]} 
\defnum{tces16ex.F140}{67.7} 
\defnum{tces16ex.F140.ci}{[60.2, 75.3]} 
\defnum{tces16.lastingloss.max}{4.2} 
\defnum{tces16.lastingloss.min}{2.1} 
\defnum{tces16.handoff.spread}{6.0} 
\defnum{tces16.gain.max}{27.6} 
\defnum{tces16.gain.min}{19.5} 
\defnum{tces.hs.M0}{4.9} 
\defnum{tces.hs.T}{4.1} 
\defnum{tces.hs.S}{5.0} 
\defnum{tces.hs.RP}{1.7} 
\defnum{tces.hs.U140}{2.7} 
\defnum{tces.hs.rep.S}{4.81} 
\defnum{tces.hs.rep.T}{4.04} 
\defnum{tces.halves.RP}{10} 
\defnum{tces.frac.tokens}{37} 
\defnum{e1.M0.never}{1} 
\defnum{e1.RP.never}{0} 
\defnum{e1.RS.never}{2} 
\defnum{e2.M0}{54.2} 
\defnum{e2s.M0}{12.5} 
\defnum{e2.T}{61.5} 
\defnum{e2s.T}{17.4} 
\defnum{e2.S}{58.9} 
\defnum{e2s.S}{14.8} 
\defnum{e2.RP}{60.9} 
\defnum{e2s.RP}{51.6} 
\defnum{e2.RS}{62.2} 
\defnum{e2s.RS}{35.2} 
\defnum{lj.M0.u1}{10.45} 
\defnum{lj.M0.u2}{8.06} 
\defnum{lj.M0.u3}{9.31} 
\defnum{lj.T.u1}{10.46} 
\defnum{lj.T.u2}{7.89} 
\defnum{lj.T.u3}{9.10} 
\defnum{lj.S.u1}{10.43} 
\defnum{lj.S.u2}{7.74} 
\defnum{lj.S.u3}{10.10} 
\defnum{lj.RP.u1}{12.90} 
\defnum{lj.RP.u2}{13.12} 
\defnum{lj.RP.u3}{11.66} 
\defnum{lj.RS.u1}{21.01} 
\defnum{lj.RS.u2}{12.32} 
\defnum{lj.RS.u3}{9.17} 
\defnum{lj.U20.u1}{10.89} 
\defnum{lj.U20.u2}{8.05} 
\defnum{lj.U20.u3}{9.08} 
\defnum{lj.U60.u1}{10.56} 
\defnum{lj.U60.u2}{8.52} 
\defnum{lj.U60.u3}{9.38} 
\defnum{lj.U140.u1}{10.35} 
\defnum{lj.U140.u2}{19.50} 
\defnum{lj.U140.u3}{19.44} 
\defnum{lj.F20.u1}{10.97} 
\defnum{lj.F20.u2}{7.86} 
\defnum{lj.F20.u3}{9.11} 
\defnum{lj.F60.u1}{10.82} 
\defnum{lj.F60.u2}{8.69} 
\defnum{lj.F60.u3}{9.24} 
\defnum{lj.F140.u1}{11.77} 
\defnum{lj.F140.u2}{26.91} 
\defnum{lj.F140.u3}{16.19} 
\defnum{lj.gentle.U140}{12.38} 
\defnum{lj.gentle.F140}{15.93} 
\defnum{lj.gentle.once.max}{8.41} 
\defnum{lj.gentle.once.min}{8.37} 
\defnum{lj.it.U140}{52.56} 
\defnum{lj.it.F140}{52.82} 
\defnum{lj.it.others.max}{52.38} 
\defnum{lj.it.others.min}{52.05} 
\defnum{rho.B}{0.93} 
\defnum{rho.B.RS}{0.89} 
\defnum{rho.group}{0.92} 
\defnum{rho.passes}{0.78} 
\defnum{rho.fa}{0.23} 
\defnum{rho.ua}{0.05} 
\defnum{rho.loss1}{0.26} 
\defnum{rho.loss2}{0.94} 
\defnum{rho.within.low}{0.4} 
\defnum{rho.within.high}{0.0} 
\defnum{pearson.loss2}{$-$0.84} 
\defnum{b.T}{$-$0.0004} 
\defnum{b.S}{$-$0.0035} 
\defnum{b.U20}{$-$0.0027} 
\defnum{b.U60}{0.015} 
\defnum{b.U140}{0.107} 
\defnum{b.F20}{$-$0.0025} 
\defnum{b.F60}{0.016} 
\defnum{b.F140}{0.161} 
\defnum{b.RP}{0.061} 
\defnum{b.RS}{0.093} 
\defnum{b.offset.lo}{0.0033} 
\defnum{b.offset.hi}{0.051} 
\defnum{br.m0top}{0.55} 
\defnum{br.U140}{0.66} 
\defnum{br.F140}{0.69} 
\defnum{br.last.to}{0.58} 
\defnum{n1.b.T}{$-$0.0001} 
\defnum{n1.a.T}{1.3} 
\defnum{n1.alt.T}{0.98} 
\defnum{n1.b.S}{$-$0.0034} 
\defnum{n1.a.S}{2.2} 
\defnum{n1.alt.S}{0.91} 
\defnum{n1.b.U140}{0.097} 
\defnum{n1.a.U140}{7.6} 
\defnum{n1.alt.U140}{0.41} 
\defnum{n1.b.F140}{0.139} 
\defnum{n1.a.F140}{8.1} 
\defnum{n1.alt.F140}{0.30} 
\defnum{n1.b.U2}{0.154} 
\defnum{n1.a.U2}{8.2} 
\defnum{n1.alt.U2}{0.25} 
\defnum{n1.b.FA1}{0.0085} 
\defnum{n1.a.FA1}{3.1} 
\defnum{n1.alt.FA1}{0.95} 
\defnum{n1.b.FA025}{0.022} 
\defnum{n1.a.FA025}{4.9} 
\defnum{n1.alt.FA025}{0.87} 
\defnum{n1.b.FN}{0.013} 
\defnum{n1.a.FN}{4.1} 
\defnum{n1.alt.FN}{0.78} 
\defnum{n1x.b.U20}{$-$0.0027} 
\defnum{n1x.a.U20}{1.9} 
\defnum{n1x.alt.U20}{0.89} 
\defnum{n1x.b.U60}{0.012} 
\defnum{n1x.a.U60}{3.7} 
\defnum{n1x.alt.U60}{0.74} 
\defnum{n1x.b.F20}{$-$0.0024} 
\defnum{n1x.a.F20}{1.9} 
\defnum{n1x.alt.F20}{0.90} 
\defnum{n1x.b.F60}{0.013} 
\defnum{n1x.a.F60}{3.8} 
\defnum{n1x.alt.F60}{0.74} 
\defnum{n1x.b.RP}{0.047} 
\defnum{n1x.a.RP}{6.8} 
\defnum{n1x.alt.RP}{0.68} 
\defnum{n1x.b.RS}{0.068} 
\defnum{n1x.a.RS}{7.8} 
\defnum{n1x.alt.RS}{0.67} 
\defnum{n1.bfrac.M0}{0} 
\defnum{n1.bfrac.T}{0.0} 
\defnum{n1.bfrac.S}{$-$0.9} 
\defnum{n1.bfrac.U140}{25} 
\defnum{n1.bfrac.F140}{36} 
\defnum{n1.bfrac.U2}{39} 
\defnum{n1.bfrac.FA1}{2.2} 
\defnum{n1.bfrac.FA025}{5.7} 
\defnum{n1.bfrac.FN}{3.4} 
\defnum{n1x.bfrac.U60}{3.2} 
\defnum{n1x.bfrac.F60}{3.4} 
\defnum{n1.kept.F140}{68} 
\defnum{n1.kept.U2}{71} 
\defnum{n1.nll}{0.390} 
\defnum{n1x.b.gap}{0.046} 
\defnum{n1x.shrink.min}{8.2} 
\defnum{n1x.shrink.max}{27} 
\defnum{bprof.T.4-7}{0.0057} 
\defnum{bprof.T.8-15}{0.0003} 
\defnum{bprof.T.16-31}{$-$0.0010} 
\defnum{bprof.T.32-63}{$-$0.0017} 
\defnum{bprof.T.64-127}{$-$0.0004} 
\defnum{bprof.T.128-255}{$-$0.0006} 
\defnum{bprof.T.256-511}{$-$0.0003} 
\defnum{bprof.T.512-1024}{$-$0.0002} 
\defnum{bprof.S.4-7}{$-$0.018} 
\defnum{bprof.S.8-15}{$-$0.0040} 
\defnum{bprof.S.16-31}{$-$0.0033} 
\defnum{bprof.S.32-63}{$-$0.0044} 
\defnum{bprof.S.64-127}{$-$0.0044} 
\defnum{bprof.S.128-255}{$-$0.0039} 
\defnum{bprof.S.256-511}{$-$0.0033} 
\defnum{bprof.S.512-1024}{$-$0.0032} 
\defnum{bprof.U20.4-7}{0.0059} 
\defnum{bprof.U20.8-15}{$-$0.0026} 
\defnum{bprof.U20.16-31}{$-$0.0019} 
\defnum{bprof.U20.32-63}{$-$0.0030} 
\defnum{bprof.U20.64-127}{$-$0.0037} 
\defnum{bprof.U20.128-255}{$-$0.0029} 
\defnum{bprof.U20.256-511}{$-$0.0026} 
\defnum{bprof.U20.512-1024}{$-$0.0027} 
\defnum{bprof.F20.4-7}{0.0017} 
\defnum{bprof.F20.8-15}{$-$0.0021} 
\defnum{bprof.F20.16-31}{$-$0.0003} 
\defnum{bprof.F20.32-63}{$-$0.0025} 
\defnum{bprof.F20.64-127}{$-$0.0032} 
\defnum{bprof.F20.128-255}{$-$0.0030} 
\defnum{bprof.F20.256-511}{$-$0.0026} 
\defnum{bprof.F20.512-1024}{$-$0.0024} 
\defnum{bprof.U60.4-7}{0.025} 
\defnum{bprof.U60.8-15}{$-$0.0020} 
\defnum{bprof.U60.16-31}{$-$0.0008} 
\defnum{bprof.U60.32-63}{0.019} 
\defnum{bprof.U60.64-127}{0.020} 
\defnum{bprof.U60.128-255}{0.019} 
\defnum{bprof.U60.256-511}{0.017} 
\defnum{bprof.U60.512-1024}{0.013} 
\defnum{bprof.F60.4-7}{0.015} 
\defnum{bprof.F60.8-15}{0.0004} 
\defnum{bprof.F60.16-31}{0.0072} 
\defnum{bprof.F60.32-63}{0.0092} 
\defnum{bprof.F60.64-127}{0.014} 
\defnum{bprof.F60.128-255}{0.020} 
\defnum{bprof.F60.256-511}{0.016} 
\defnum{bprof.F60.512-1024}{0.016} 
\defnum{bprof.U140.4-7}{0.023} 
\defnum{bprof.U140.8-15}{$-$0.0014} 
\defnum{bprof.U140.16-31}{0.0003} 
\defnum{bprof.U140.32-63}{0.080} 
\defnum{bprof.U140.64-127}{0.114} 
\defnum{bprof.U140.128-255}{0.132} 
\defnum{bprof.U140.256-511}{0.123} 
\defnum{bprof.U140.512-1024}{0.099} 
\defnum{bprof.F140.4-7}{0.034} 
\defnum{bprof.F140.8-15}{0.014} 
\defnum{bprof.F140.16-31}{0.032} 
\defnum{bprof.F140.32-63}{0.103} 
\defnum{bprof.F140.64-127}{0.161} 
\defnum{bprof.F140.128-255}{0.207} 
\defnum{bprof.F140.256-511}{0.181} 
\defnum{bprof.F140.512-1024}{0.150} 
\defnum{bprof.RP.4-7}{0.030} 
\defnum{bprof.RP.8-15}{0.018} 
\defnum{bprof.RP.16-31}{0.0062} 
\defnum{bprof.RP.32-63}{0.081} 
\defnum{bprof.RP.64-127}{0.093} 
\defnum{bprof.RP.128-255}{0.086} 
\defnum{bprof.RP.256-511}{0.067} 
\defnum{bprof.RP.512-1024}{0.048} 
\defnum{bprof.RS.4-7}{0.185} 
\defnum{bprof.RS.8-15}{0.075} 
\defnum{bprof.RS.16-31}{0.043} 
\defnum{bprof.RS.32-63}{0.108} 
\defnum{bprof.RS.64-127}{0.145} 
\defnum{bprof.RS.128-255}{0.137} 
\defnum{bprof.RS.256-511}{0.093} 
\defnum{bprof.RS.512-1024}{0.075} 
\defnum{bprof.lasting.max}{0.004} 
\defnum{anc.h16.FA1}{75.3} 
\defnum{anc.h4.FA1}{51.0} 
\defnum{anc.a16.FA1}{48.2} 
\defnum{anc.a4.FA1}{25.8} 
\defnum{anc.cap.FA1}{49.0} 
\defnum{anc.strict.FA1}{13.0} 
\defnum{anc.free.FA1}{62.2} 
\defnum{anc.d16.FA1}{+36.5} 
\defnum{anc.d16ci.FA1}{[30.5, 42.2]} 
\defnum{anc.b.FA1}{0.0093} 
\defnum{anc.h16.FA025}{78.6} 
\defnum{anc.h4.FA025}{53.6} 
\defnum{anc.a16.FA025}{27.9} 
\defnum{anc.a4.FA025}{7.6} 
\defnum{anc.cap.FA025}{71.1} 
\defnum{anc.strict.FA025}{10.2} 
\defnum{anc.free.FA025}{61.7} 
\defnum{anc.d16.FA025}{+16.1} 
\defnum{anc.d16ci.FA025}{[10.9, 21.4]} 
\defnum{anc.b.FA025}{0.024} 
\defnum{anc.h16.FN}{85.7} 
\defnum{anc.h4.FN}{58.3} 
\defnum{anc.a16.FN}{25.3} 
\defnum{anc.a4.FN}{4.9} 
\defnum{anc.cap.FN}{72.4} 
\defnum{anc.strict.FN}{6.5} 
\defnum{anc.free.FN}{55.2} 
\defnum{anc.d16.FN}{+13.5} 
\defnum{anc.d16ci.FN}{[7.8, 19.3]} 
\defnum{anc.b.FN}{0.015} 
\defnum{anc.h16.U2}{70.6} 
\defnum{anc.h4.U2}{45.3} 
\defnum{anc.a16.U2}{6.5} 
\defnum{anc.a4.U2}{0.3} 
\defnum{anc.cap.U2}{92.2} 
\defnum{anc.strict.U2}{2.1} 
\defnum{anc.free.U2}{22.4} 
\defnum{anc.d16.U2}{$-$5.2} 
\defnum{anc.d16ci.U2}{[$-$9.6, $-$1.0]} 
\defnum{anc.b.U2}{0.172} 
\defnum{anc.h4.F140}{61.7} 
\defnum{anc.a4.F140}{4.2} 
\defnum{anc.free.F140}{55.7} 
\defnum{anc.gap16.FA1}{6.3} 
\defnum{anc.gap4.FA1}{10.7} 
\defnum{anc.gap16.FA025}{2.9} 
\defnum{anc.gap4.FA025}{8.1} 
\defnum{anc.gain.F140}{21.1} 
\defnum{anc.gain.FA1}{24.2} 
\defnum{anc.fit.FA1}{85} 
\defnum{anc.fit.FA025}{99} 
\defnum{anc.fit.FN}{36} 
\defnum{anc.loss16.FA1}{27.1} 
\defnum{m.h.base}{94.7} 
\defnum{m.h.D20}{95.4} 
\defnum{m.h.D60}{95.1} 
\defnum{m.h.D140}{95.4} 
\defnum{m.h.O140}{94.6} 
\defnum{m.it.base}{95.6} 
\defnum{m.it.D20}{94.5} 
\defnum{m.it.D60}{94.0} 
\defnum{m.it.D140}{59.3} 
\defnum{m.it.O140}{94.7} 
\defnum{m.b.base}{91.7} 
\defnum{m.b.D20}{94.3} 
\defnum{m.b.D60}{92.8} 
\defnum{m.b.D140}{22.5} 
\defnum{m.b.O140}{94.8} 
\defnum{mfree.h.base}{74.7} 
\defnum{mfree.h.D20}{95.0} 
\defnum{mfree.h.D60}{97.3} 
\defnum{mfree.h.D140}{95.2} 
\defnum{mfree.h.O140}{94.1} 
\defnum{mfree.b.O140}{74.2} 
\defnum{mfree.it.max}{33.5} 
\defnum{mfree.it.min}{22.4} 
\defnum{m.it.gapDO}{35.4} 
\defnum{mfree.it.gapDO}{9.7} 
\defnum{mex.O.it}{36.2} 
\defnum{mexci.O.it}{[31.9, 40.4]} 
\defnum{mex.O.b}{73.1} 
\defnum{mexci.O.b}{[69.3, 76.7]} 
\defnum{mloss.it.D60}{1.1} 
\defnum{mlossci.it.D60}{[$-$0.9, 2.9]} 
\defnum{mloss.b.D60}{2.4} 
\defnum{mlossci.b.D60}{[0.2, 4.5]} 
\defnum{mgain.it.base}{0.9} 
\defnum{mgain.it.O140}{0.1} 
\defnum{m.rho.b}{0.4} 
\defnum{m.rho.it}{1.0} 
\defnum{mb.D60.3}{0.016} 
\defnum{mb.D140.3}{0.111} 
\defnum{mb.nll}{0.459} 
\defnum{mb.D140.pdrop}{10} 
\defnum{mb.frac.D20}{0.1} 
\defnum{mb.frac.D60}{3.4} 
\defnum{mb.frac.D140}{24} 
\defnum{mb.frac.O140}{0.1} 
\defnum{m.bfrac.base}{0} 
\defnum{mb.kept}{63} 
\defnum{mb.alt.D140}{0.48} 
\defnum{aime.h.base}{75.8} 
\defnum{aime.h.D140}{69.9} 
\defnum{aime.h.O140}{69.5} 
\defnum{aime.it.base}{71.6} 
\defnum{aime.it.D140}{22.0} 
\defnum{aime.it.O140}{70.8} 
\defnum{aime.base.h16}{51.7} 
\defnum{aime.it.D140.cap}{28.8} 
\defnum{aime.it.D140.loop}{53} 
\defnum{aime.it.D140.unboxed}{36.4} 
\defnum{mbud.2048.h.base}{67.2} 
\defnum{mbud.4096.h.base}{79.6} 
\defnum{mbud.8192.h.base}{88.3} 
\defnum{mbud.16384.h.base}{94.7} 
\defnum{mcap.h.base}{5.1} 
\defnum{munbox.h.base}{3.8} 
\defnum{mtok.h.base}{3,340} 
\defnum{mbud.2048.h.D140}{71.5} 
\defnum{mbud.4096.h.D140}{81.6} 
\defnum{mbud.8192.h.D140}{89.5} 
\defnum{mbud.16384.h.D140}{95.4} 
\defnum{mcap.h.D140}{4.5} 
\defnum{munbox.h.D140}{3.6} 
\defnum{mtok.h.D140}{3,058} 
\defnum{mbud.2048.h.O140}{68.3} 
\defnum{mbud.4096.h.O140}{80.1} 
\defnum{mbud.8192.h.O140}{88.8} 
\defnum{mbud.16384.h.O140}{94.6} 
\defnum{mcap.h.O140}{5.4} 
\defnum{munbox.h.O140}{3.8} 
\defnum{mtok.h.O140}{3,370} 
\defnum{mbud.4096.h.D60}{80.2} 
\defnum{mbud.2048.it.base}{71.8} 
\defnum{mbud.4096.it.base}{83.7} 
\defnum{mbud.8192.it.base}{91.1} 
\defnum{mbud.16384.it.base}{95.6} 
\defnum{mcap.it.base}{2.5} 
\defnum{munbox.it.base}{1.9} 
\defnum{mtok.it.base}{2,712} 
\defnum{mbud.2048.it.D140}{51.4} 
\defnum{mbud.4096.it.D140}{55.2} 
\defnum{mbud.8192.it.D140}{58.3} 
\defnum{mbud.16384.it.D140}{59.3} 
\defnum{mcap.it.D140}{5.9} 
\defnum{munbox.it.D140}{20.1} 
\defnum{mtok.it.D140}{2,007} 
\defnum{mbud.2048.it.O140}{67.5} 
\defnum{mbud.4096.it.O140}{81.3} 
\defnum{mbud.8192.it.O140}{89.3} 
\defnum{mbud.16384.it.O140}{94.7} 
\defnum{mcap.it.O140}{4.8} 
\defnum{munbox.it.O140}{4.1} 
\defnum{mtok.it.O140}{3,196} 
\defnum{mbud.4096.it.D60}{83.3} 
\defnum{mbud.2048.b.base}{67.9} 
\defnum{mbud.4096.b.base}{78.5} 
\defnum{mbud.8192.b.base}{86.5} 
\defnum{mbud.16384.b.base}{91.7} 
\defnum{mcap.b.base}{7.0} 
\defnum{munbox.b.base}{5.7} 
\defnum{mtok.b.base}{3,608} 
\defnum{mbud.2048.b.D140}{20.4} 
\defnum{mbud.4096.b.D140}{21.8} 
\defnum{mbud.8192.b.D140}{22.5} 
\defnum{mbud.16384.b.D140}{22.5} 
\defnum{mcap.b.D140}{2.6} 
\defnum{munbox.b.D140}{40.3} 
\defnum{mtok.b.D140}{867} 
\defnum{mbud.2048.b.O140}{67.9} 
\defnum{mbud.4096.b.O140}{79.5} 
\defnum{mbud.8192.b.O140}{89.6} 
\defnum{mbud.16384.b.O140}{94.8} 
\defnum{mcap.b.O140}{4.5} 
\defnum{munbox.b.O140}{3.6} 
\defnum{mtok.b.O140}{3,162} 
\defnum{mbud.4096.b.D60}{80.8} 
\defnum{mcap.b.D20}{5.8} 
\defnum{mcap.b.D60}{4.2} 
\defnum{mf.b.D.acc}{22.5} 
\defnum{mf.b.D.int}{32.2} 
\defnum{mf.b.D.cap}{2.6} 
\defnum{mf.b.D.unbox}{40.3} 
\defnum{mf.b.D.closes}{0.9} 
\defnum{mf.b.D.ends}{97} 
\defnum{mf.b.D.short}{49} 
\defnum{mf.b.D.med}{103} 
\defnum{mf.b.O.acc}{94.8} 
\defnum{mf.b.O.int}{95.0} 
\defnum{mf.b.O.cap}{4.5} 
\defnum{mf.b.O.unbox}{3.6} 
\defnum{mf.b.O.closes}{96} 
\defnum{mf.b.O.ends}{0.0} 
\defnum{mf.b.O.short}{0.0} 
\defnum{mf.b.O.med}{1,294} 
\defnum{mf.b.base.acc}{91.7} 
\defnum{mf.b.base.int}{92.2} 
\defnum{mf.b.base.cap}{7.0} 
\defnum{mf.b.base.unbox}{5.7} 
\defnum{mf.b.base.closes}{84} 
\defnum{mf.b.base.ends}{11} 
\defnum{mf.b.base.short}{1.0} 
\defnum{mf.b.base.med}{1,474} 
\defnum{mf.it.D.acc}{59.3} 
\defnum{mf.it.D.int}{67.2} 
\defnum{mf.it.D.cap}{5.9} 
\defnum{mf.it.D.unbox}{20.1} 
\defnum{mf.it.D.closes}{29} 
\defnum{mf.it.D.ends}{78} 
\defnum{mf.it.D.short}{7.5} 
\defnum{mf.it.D.med}{451} 
\defnum{mf.credit.b}{9.7} 
\defnum{mf.lenient.b}{64.0} 
\defnum{mf.credit.it}{7.9} 
\defnum{mf.lenient.it}{29.1} 
\defnum{mf.n}{884} 
\defnum{mf.closes.n}{8} 
\defnum{mf.short.tokens}{100} 
\defnum{m.src.rows}{64,312} 
\defnum{m.contam}{91} 
\defnum{m.scr.nonint}{12,493} 
\defnum{m.scr.multi}{2,614} 
\defnum{m.scr.dup}{107} 
\defnum{m.scr.fig}{331} 
\defnum{m.scr.long}{28} 
\defnum{m.scr.guess}{6,563} 
\defnum{m.scr.prompt}{512} 
\defnum{m.ref.rows}{200} 
\defnum{m.ref.agree}{93} 
\defnum{m35.h.base}{96.0} 
\defnum{m35.h.D20}{96.0} 
\defnum{m35.h.D60}{96.3} 
\defnum{m35.h.D140}{94.9} 
\defnum{m35.h.O140}{96.7} 
\defnum{m35.it.base}{48.1} 
\defnum{m35.it.D20}{93.4} 
\defnum{m35.it.D60}{91.6} 
\defnum{m35.it.D140}{50.8} 
\defnum{m35.it.O140}{91.0} 
\defnum{m35.b.base}{93.7} 
\defnum{m35.b.D20}{96.0} 
\defnum{m35.b.D60}{79.6} 
\defnum{m35.b.D140}{21.5} 
\defnum{m35.b.O140}{95.4} 
\defnum{m35.aime.h.base}{79.7} 
\defnum{m35.aime.h.D140}{72.0} 
\defnum{m35.aime.h.O140}{78.8} 
\defnum{m35.aime.it.base}{5.9} 
\defnum{m35.aime.it.D140}{11.4} 
\defnum{m35.aime.it.O140}{77.5} 
\defnum{m35.free.h.base}{82.6} 
\defnum{m35.cap.b.base}{4.0} 
\defnum{m35.free.h.D140}{95.2} 
\defnum{m35.cap.b.D140}{26.6} 
\defnum{m35.free.h.O140}{97.1} 
\defnum{m35.cap.b.O140}{5.1} 
\defnum{m35.nll}{0.499} 
\defnum{m35.B.D140}{0.160} 
\defnum{m35.B.O140}{0.0009} 
\defnum{m35.bfrac.D60}{4.3} 
\defnum{m35.bfrac.D140}{32} 
\defnum{m35.bfrac.O140}{0.2} 
\defnum{m35.kept}{57} 
\defnum{m35.ex.it}{38.3} 
\defnum{m35.ex.it.ci}{[33.7, 43.0]} 
\defnum{m35.ex.b}{72.1} 
\defnum{m35.ex.b.ci}{[68.0, 75.9]} 
\defnum{m35.parity}{1.8} 
\defnum{m35.rho.b}{0.7} 
\defnum{m35.rho.it}{$-$0.1} 
\defnum{mnem.h.base}{93.6} 
\defnum{mnem.h.D20}{93.1} 
\defnum{mnem.h.D60}{95.6} 
\defnum{mnem.h.D140}{93.4} 
\defnum{mnem.h.O140}{93.4} 
\defnum{mnem.it.base}{83.6} 
\defnum{mnem.it.D20}{88.8} 
\defnum{mnem.it.D60}{69.8} 
\defnum{mnem.it.D140}{22.3} 
\defnum{mnem.it.O140}{81.4} 
\defnum{mnem.b.base}{94.1} 
\defnum{mnem.b.D20}{90.4} 
\defnum{mnem.b.D60}{92.4} 
\defnum{mnem.b.D140}{54.2} 
\defnum{mnem.b.O140}{94.2} 
\defnum{mnem.aime.h.base}{72.9} 
\defnum{mnem.aime.h.D140}{70.8} 
\defnum{mnem.aime.h.O140}{74.6} 
\defnum{mnem.aime.it.base}{67.8} 
\defnum{mnem.aime.it.D140}{2.5} 
\defnum{mnem.aime.it.O140}{64.4} 
\defnum{mnem.free.h.base}{72.2} 
\defnum{mnem.cap.b.base}{7.6} 
\defnum{mnem.free.h.D140}{93.2} 
\defnum{mnem.cap.b.D140}{8.5} 
\defnum{mnem.free.h.O140}{93.7} 
\defnum{mnem.cap.b.O140}{7.1} 
\defnum{mnem.nll}{0.640} 
\defnum{mnem.B.D140}{0.114} 
\defnum{mnem.B.O140}{0.0002} 
\defnum{mnem.bfrac.D60}{0.8} 
\defnum{mnem.bfrac.D140}{18} 
\defnum{mnem.bfrac.O140}{0.0} 
\defnum{mnem.kept}{61} 
\defnum{mnem.ex.it}{59.2} 
\defnum{mnem.ex.it.ci}{[54.5, 63.8]} 
\defnum{mnem.ex.b}{40.0} 
\defnum{mnem.ex.b.ci}{[35.6, 44.6]} 
\defnum{mnem.rho.b}{0.9} 
\defnum{mnem.rho.it}{0.7} 
\defnum{m35.unbox.b.D140}{13.8} 
\defnum{m35.loss.b.D60}{16.6} 
\defnum{m35.loss.b.D60.ci}{[13.5, 19.8]} 
\defnum{mnem.unbox.it.D140}{50.5} 
\defnum{mnem.loss.it.D60}{25.8} 
\defnum{mnem.loss.it.D60.ci}{[21.9, 29.8]} 
\defnum{m35.loop.b.D140}{98} 
\defnum{m35.med.b.D140}{270} 
\defnum{m35.short.b.D140}{62} 
\defnum{mnem.cap.it.D140}{1.2} 
\defnum{mnem.tok.it.D140}{1,138} 
\defnum{m35.noise.B}{0.0000} 
\defnum{m35.noise.ci}{[$-$0.00002, 0.00002]} 
\defnum{mm.P.h}{94.6} 
\defnum{mm.P.it}{94.9} 
\defnum{mm.P.b}{94.2} 
\defnum{mm.P.free}{94.8} 
\defnum{mm.P.cap.b}{5.0} 
\defnum{mm.P.bfrac}{0.1} 
\defnum{mm.D2.h}{93.9} 
\defnum{mm.D2.it}{81.4} 
\defnum{mm.D2.b}{5.5} 
\defnum{mm.D2.free}{95.2} 
\defnum{mm.D2.cap.b}{2.1} 
\defnum{mm.D2.med.b}{26} 
\defnum{mm.D2.bfrac}{22} 
\defnum{mm.R.h}{95.6} 
\defnum{mm.R.it}{56.8} 
\defnum{mm.R.b}{12.9} 
\defnum{mm.R.free}{94.1} 
\defnum{mm.R.cap.b}{2.5} 
\defnum{mm.R.med.b}{59} 
\defnum{mm.R.bfrac}{26} 
\defnum{mm.P.B}{0.0004} 
\defnum{mm.P.aime.h}{72.0} 
\defnum{mm.P.aime.it}{75.4} 
\defnum{mm.P.distinct}{577} 
\defnum{mm.P.repeat}{6} 
\defnum{mm.slots}{4,480} 
\defnum{mm.D2O.it}{12.6} 
\defnum{mm.D2O.it.ci}{[8.8, 16.3]} 
\defnum{mm.RO.it}{38.9} 
\defnum{mm.RO.it.ci}{[35.0, 42.9]} 
\defnum{mm.PO.it}{$-$0.2} 
\defnum{mm.PO.it.ci}{[$-$2.0, 1.5]} 
\defnum{mm.DP.it}{36.4} 
\defnum{mm.DP.it.ci}{[32.4, 40.4]} 
\defnum{mm.D2O.b}{88.6} 
\defnum{mm.D2O.b.ci}{[85.5, 91.5]} 
\defnum{mm.RO.b}{82.9} 
\defnum{mm.RO.b.ci}{[79.6, 86.1]} 
\defnum{mm.PO.b}{0.6} 
\defnum{mm.PO.b.ci}{[$-$1.7, 2.8]} 
\defnum{mm.DP.b}{72.5} 
\defnum{mm.DP.b.ci}{[68.8, 76.0]} 
\defnum{mm.R.nll.base}{0.453} 
\defnum{mm.R.nll.r32}{0.277} 
\defnum{mm.R.nll.r128}{0.270} 
\defnum{mm.R.fitdiff}{3.8} 
\defnum{mm.R.ratio.b}{1.14} 
\defnum{mm.R.ratio.it}{1.08} 
\defnum{rfa.D}{95.4} 
\defnum{rfa.D.it}{69.0} 
\defnum{rfa.D.b}{37.6} 
\defnum{rfa.O}{94.6} 
\defnum{rfa.O.it}{94.7} 
\defnum{rfa.O.b}{94.8} 
\defnum{rfa.base}{94.7} 
\defnum{rfa.base.it}{95.7} 
\defnum{rfa.base.b}{92.3} 
\defnum{rfa.ex.it}{26.5} 
\defnum{rfa.ex.it.ci}{[22.3, 30.5]} 
\defnum{rfa.ex.b}{58.0} 
\defnum{rfa.ex.b.ci}{[53.7, 62.2]} 
\defnum{rfb.D}{93.7} 
\defnum{rfb.D.it}{71.7} 
\defnum{rfb.D.b}{26.1} 
\defnum{rfb.O}{92.9} 
\defnum{rfb.O.it}{92.2} 
\defnum{rfb.O.b}{93.7} 
\defnum{rfb.base}{92.8} 
\defnum{rfb.base.it}{93.3} 
\defnum{rfb.base.b}{82.5} 
\defnum{rfb.ex.it}{21.3} 
\defnum{rfb.ex.it.ci}{[16.9, 25.7]} 
\defnum{rfb.ex.b}{68.3} 
\defnum{rfb.ex.b.ci}{[64.3, 72.2]} 
\defnum{rfb.cont.D.it}{74} 
\defnum{rfb.cont.D.b}{91} 
\defnum{rfb.med.D.b}{6,688} 
\defnum{rfb.cap.D.b}{47.4} 
\defnum{rf.cost.min}{1.1} 
\defnum{rf.cost.max}{2.5} 
\defnum{rf.cost.baseb}{9.3} 
\defnum{rf.share.it}{41} 
\defnum{rf.share.b}{6.5} 
\defnum{r1.d.D.u2}{0.067} 
\defnum{r1.d.O.u2}{0.0005} 
\defnum{r1.d.P.u2}{0.0004} 
\defnum{reg.R1.3.obs}{0.74--1.18} 
\defnum{reg.R1.2.obs}{81.8} 
\defnum{reg.R3b.obs}{$-$0.0068} 
\defnum{r1.lossspread.max}{4.0} 
\defnum{reg.R1.4.obs.min}{1.7} 
\defnum{r1.own.base}{0.357} 
\defnum{r1.own.D20}{0.403} 
\defnum{r2.ex20}{12.6} 
\defnum{r2.ex}{13.1} 
\defnum{r2.ex.ci}{[9.8, 16.5]} 
\defnum{r2.examples}{1,920} 
\defnum{r3.ex60}{55.2} 
\defnum{r3.ex60.ci}{[50.9, 59.4]} 
\defnum{r3.f.D.u40}{41.1} 
\defnum{r3.f.D.u60}{41.6} 
\defnum{r3.f.O.u60}{96.0} 
\defnum{r3b.D.approx}{0.006} 
\defnum{r2.f.D.u20}{82.8} 
\defnum{r2.f.D.u60}{82.2} 
\defnum{r2.f.O.u20}{94.6} 
\defnum{r2.f.O.u60}{94.6} 
\defnum{r4.f.DT.u20}{89.7} 
\defnum{r4.f.DT.u60}{86.8} 
\defnum{r4.loss.DT.u1-5}{1.85} 
\defnum{r4.loss.DT.u56-60}{1.74} 
\defnum{r4.f.OT.u20}{96.6} 
\defnum{r4.f.OT.u60}{94.9} 
\defnum{r4.loss.OT.u1-5}{1.80} 
\defnum{r4.loss.OT.u56-60}{1.74} 
\defnum{r5.f.D}{85.4} 
\defnum{r5.loss.D.u1-5}{1.30} 
\defnum{r5.loss.D.u56-60}{0.97} 
\defnum{r5.f.O}{94.9} 
\defnum{r5.loss.O.u1-5}{1.18} 
\defnum{r5.loss.O.u56-60}{0.96} 
\defnum{r5.f.DT}{85.3} 
\defnum{r5.loss.DT.u1-5}{1.30} 
\defnum{r5.loss.DT.u56-60}{0.97} 
\defnum{r5.f.OT}{95.8} 
\defnum{r5.loss.OT.u1-5}{1.24} 
\defnum{r5.loss.OT.u56-60}{0.96} 
\defnum{r4.ex60}{6.9} 
\defnum{r4.ex60.ci}{[4.0, 10.0]} 
\defnum{r5.ex.D}{10.3} 
\defnum{r5.ex.D.ci}{[7.5, 13.1]} 
\defnum{r5.ex.DT}{9.3} 
\defnum{r5.ex.DT.ci}{[6.1, 12.2]} 
\defnum{r5.lost.O}{$-$0.3} 
\defnum{r5.lost.OT}{0.5} 
\defnum{r5.mix.aya}{361} 
\defnum{r5.mix.flan}{356} 
\defnum{r5.mix.codealpaca}{344} 
\defnum{r5.mix.wildjailbreak}{203} 
\defnum{r5.mix.wildchat}{164} 
\defnum{r5.mix.wildguard}{182} 
\defnum{r5.mix.phcode}{119} 
\defnum{r5.mix.phif}{98} 
\defnum{r5.mix.coconot}{35} 
\defnum{r5.mix.sciriff}{28} 
\defnum{r5.mix.oasst}{21} 
\defnum{r5.mix.tablegpt}{9} 
\defnum{r5.mathlike}{10.4} 
\defnum{r5.mathlike.n}{200} 
\defnum{r45.med.D.min}{597} 
\defnum{r45.med.D.max}{749} 
\defnum{r45.close.D.min}{29} 
\defnum{r45.close.D.max}{38} 
\defnum{r45.unbox.D.min}{4.4} 
\defnum{r45.unbox.D.max}{7.0} 
\defnum{r45.med.O.min}{1,166} 
\defnum{r45.med.O.max}{1,407} 
\defnum{r45.close.O.min}{96} 
\defnum{r45.unbox.O.min}{2.4} 
\defnum{r45.unbox.O.max}{3.3} 
\defnum{r45.cap.min}{2.3} 
\defnum{r45.cap.max}{4.5} 
\defnum{mt.n.D}{575} 
\defnum{mt.n.O}{4,441} 
\defnum{mt.pass.D}{7.8} 
\defnum{mt.pass.D20}{1.1} 
\defnum{mt.D.free}{93.9} 
\defnum{mt.D.h}{95.0} 
\defnum{mt.D.it}{33.7} 
\defnum{mt.D.b}{1.1} 
\defnum{mt.cap.b.D}{0.9} 
\defnum{mt.bfrac.D}{37} 
\defnum{mt.D20.free}{95.0} 
\defnum{mt.D20.h}{96.2} 
\defnum{mt.D20.it}{95.4} 
\defnum{mt.D20.b}{94.3} 
\defnum{mt.cap.b.D20}{3.6} 
\defnum{mt.bfrac.D20}{4.6} 
\defnum{mt.O.free}{95.7} 
\defnum{mt.O.h}{96.3} 
\defnum{mt.O.it}{94.6} 
\defnum{mt.O.b}{94.8} 
\defnum{mt.cap.b.O}{4.6} 
\defnum{mt.bfrac.O}{5.3} 
\defnum{mt.P.free}{95.5} 
\defnum{mt.P.h}{96.0} 
\defnum{mt.P.it}{95.4} 
\defnum{mt.P.b}{95.4} 
\defnum{mt.cap.b.P}{4.0} 
\defnum{mt.bfrac.P}{5.8} 
\defnum{mt.med.b.D}{22} 
\defnum{mt.med.it.D}{277} 
\defnum{mt.unbox.it.D}{39.6} 
\defnum{mt.ex.O.it}{59.6} 
\defnum{mt.ex.O.it.ci}{[55.5, 63.7]} 
\defnum{mt.ex.O.b}{92.4} 
\defnum{mt.ex.O.b.ci}{[89.4, 95.1]} 
\defnum{mt.PO.it}{$-$1.0} 
\defnum{mt.PO.it.ci}{[$-$2.6, 0.5]} 
\defnum{mt.PO.b}{$-$0.8} 
\defnum{mt.PO.b.ci}{[$-$2.7, 1.1]} 
\defnum{mt.aime.D.h}{73.7} 
\defnum{mt.aime.D.it}{11.0} 
\defnum{mt.aime.O.h}{72.9} 
\defnum{mt.aime.O.it}{78.0} 
\defnum{mt.held.drop}{1.2} 
\defnum{mt.drill.ratio}{0.69} 
\defnum{mt.d.D.own}{0.058} 
\defnum{mt.d.D.base}{$-$0.026} 
\defnum{mt.d.O.own}{0.0004} 
\defnum{mt.d.O.base}{0.0056} 
\defnum{mt.d.P.base}{0.0054} 
\defnum{mt.d.D.toward}{0.026} 
\defnum{mt.d.ratio}{141} 
\defnum{rp.D.h}{95.4} 
\defnum{rp.D.a}{94.3} 
\defnum{rp.O.h}{94.6} 
\defnum{rp.O.a}{94.7} 
\defnum{rp.DT.h}{95.0} 
\defnum{rp.DT.a}{95.5} 
\defnum{rp.OT.h}{96.3} 
\defnum{rp.OT.a}{95.8} 
\defnum{rp.ex.D}{1.1} 
\defnum{rp.ex.D.ci}{[$-$0.9, 3.3]} 
\defnum{rp.cut.D}{91} 
\defnum{rp.ex.DT}{$-$0.9} 
\defnum{rp.ex.DT.ci}{[$-$2.8, 0.8]} 
\defnum{rs.updates}{267} 
\defnum{rs.rows}{8,544} 
\defnum{rs.base.h}{94.7} 
\defnum{rs.base.a}{94.5} 
\defnum{rs.loss.base}{0.2} 
\defnum{rs.loss.base.ci}{[$-$1.2, 1.6]} 
\defnum{rs.D.h}{95.4} 
\defnum{rs.D.a}{86.0} 
\defnum{rs.loss.D}{9.4} 
\defnum{rs.loss.D.ci}{[6.8, 12.1]} 
\defnum{rs.O.h}{94.6} 
\defnum{rs.O.a}{94.8} 
\defnum{rs.loss.O}{$-$0.2} 
\defnum{rs.loss.O.ci}{[$-$1.5, 1.0]} 
\defnum{rs.P.h}{94.6} 
\defnum{rs.P.a}{95.0} 
\defnum{rs.loss.P}{$-$0.5} 
\defnum{rs.loss.P.ci}{[$-$1.6, 0.7]} 
\defnum{rs.ex.D}{9.6} 
\defnum{rs.ex.D.ci}{[6.9, 12.4]} 
\defnum{rs.ex.P}{$-$0.2} 
\defnum{rs.ex.P.ci}{[$-$1.9, 1.6]} 
\defnum{rl.D.it.u0}{59.3} 
\defnum{rl.D.it.u1}{84.7} 
\defnum{rl.D.it.u2}{90.3} 
\defnum{rl.D.it.u5}{93.9} 
\defnum{rl.D.it.R5}{0.98} 
\defnum{rl.D.it.R5.ci}{[0.94, 1.01]} 
\defnum{rl.D.it.u20}{94.3} 
\defnum{rl.D.it.hab}{94.0} 
\defnum{rl.D.it.habR}{0.98} 
\defnum{rl.D.b.u0}{22.5} 
\defnum{rl.D.b.u1}{51.0} 
\defnum{rl.D.b.u2}{79.5} 
\defnum{rl.D.b.u5}{92.8} 
\defnum{rl.D.b.R5}{0.97} 
\defnum{rl.D.b.R5.ci}{[0.95, 1.00]} 
\defnum{rl.D.b.u20}{93.3} 
\defnum{rl.D.b.hab}{92.3} 
\defnum{rl.D.b.habR}{0.97} 
\defnum{rl.DT.it.u0}{33.7} 
\defnum{rl.DT.it.u1}{76.4} 
\defnum{rl.DT.it.u2}{87.1} 
\defnum{rl.DT.it.u5}{93.1} 
\defnum{rl.DT.it.R5}{0.98} 
\defnum{rl.DT.it.R5.ci}{[0.94, 1.00]} 
\defnum{rl.DT.it.u20}{93.3} 
\defnum{rl.DT.b.u0}{1.1} 
\defnum{rl.DT.b.u1}{21.6} 
\defnum{rl.DT.b.u2}{70.5} 
\defnum{rl.DT.b.u5}{91.5} 
\defnum{rl.DT.b.R5}{0.96} 
\defnum{rl.DT.b.R5.ci}{[0.94, 0.99]} 
\defnum{rl.DT.b.u20}{93.0} 
\defnum{rl.ref}{94.6--94.8} 
\defnum{rl.O.it.u0}{94.7} 
\defnum{rl.O.it.u1}{95.2} 
\defnum{rl.O.it.u2}{94.2} 
\defnum{rl.O.it.u5}{94.3} 
\defnum{rl.O.it.u20}{94.6} 
\defnum{rl.ceil.max}{0.6} 
\defnum{sh.B}{0.082} 
\defnum{sh.B.ci}{[0.077, 0.087]} 
\defnum{sh.texts}{4,453} 
\defnum{sh.B.D}{0.111} 
\defnum{sh.B.O}{0.0004} 
\defnum{sh.h}{93.1} 
\defnum{sh.it}{95.0} 
\defnum{sh.b}{92.4} 
\defnum{sh.ex.it}{$-$1.8} 
\defnum{sh.ex.it.ci}{[$-$4.1, 0.3]} 
\defnum{sh.ex.b}{0.9} 
\defnum{sh.ex.b.ci}{[$-$1.6, 3.4]} 
\defnum{sd.s2.D.h}{94.8} 
\defnum{sd.s2.D.it}{32.7} 
\defnum{sd.s2.D.lr}{80.9} 
\defnum{sd.s2.O.h}{96.2} 
\defnum{sd.s2.O.it}{95.1} 
\defnum{sd.s2.O.lr}{95.4} 
\defnum{sd.s2.P.h}{95.6} 
\defnum{sd.s2.P.it}{94.6} 
\defnum{sd.s2.P.lr}{94.0} 
\defnum{sd.s2.ex.it}{61.1} 
\defnum{sd.s2.ex.it.ci}{[57.0, 65.2]} 
\defnum{sd.s2.ex.lr}{13.1} 
\defnum{sd.s2.ex.lr.ci}{[10.0, 16.3]} 
\defnum{sd.s3.D.h}{94.6} 
\defnum{sd.s3.D.it}{45.9} 
\defnum{sd.s3.D.lr}{76.1} 
\defnum{sd.s3.O.h}{94.9} 
\defnum{sd.s3.O.it}{95.1} 
\defnum{sd.s3.O.lr}{94.2} 
\defnum{sd.s3.P.h}{94.2} 
\defnum{sd.s3.P.it}{94.5} 
\defnum{sd.s3.P.lr}{94.9} 
\defnum{sd.s3.ex.it}{48.9} 
\defnum{sd.s3.ex.it.ci}{[45.2, 52.5]} 
\defnum{sd.s3.ex.lr}{17.8} 
\defnum{sd.s3.ex.lr.ci}{[14.1, 21.5]} 
\defnum{sd.PO.absmax}{1.4} 
\defnum{sd.pool.it.mean}{39.7} 
\defnum{sd.pool.it.min}{12.6} 
\defnum{sd.pool.it.max}{61.1} 
\defnum{sd.pool.it.sd}{20.7} 
\defnum{sd.pool.lr.mean}{14.7} 
\defnum{sd.pool.lr.min}{13.1} 
\defnum{sd.pool.lr.max}{17.8} 
\defnum{sd.out.SD1}{held} 
\defnum{sd.out.SD3}{held} 
\defnum{sk.smoke8.b5}{75.0} 
\defnum{sk.smoke8.b6}{55.0} 
\defnum{sk.smoke8.w12}{85.0} 
\defnum{sk.smoke8.w16}{60.0} 
\defnum{sk.cal}{0.2} 
\defnum{sk.gate}{66.1} 
\defnum{sk.gate.ci}{[63.3, 68.8]} 
\defnum{sk.gate.cap}{0.1} 
\defnum{sk.base}{0.2} 
\defnum{sk.D.h}{37.9} 
\defnum{sk.D.lr}{26.6} 
\defnum{sk.D.it}{0.1} 
\defnum{sk.O.h}{66.3} 
\defnum{sk.O.lr}{50.8} 
\defnum{sk.O.it}{0.4} 
\defnum{sk.P.h}{58.1} 
\defnum{sk.P.lr}{38.7} 
\defnum{sk.P.it}{0.4} 
\defnum{sk.PD.h}{20.2} 
\defnum{sk.PD.h.ci}{[16.2, 24.1]} 
\defnum{sk.fix.h}{$-$8.2} 
\defnum{sk.loss.D.lr}{11.3} 
\defnum{sk.loss.D.lr.ci}{[7.8, 14.8]} 
\defnum{sk.loss.O.lr}{15.5} 
\defnum{sk.loss.O.lr.ci}{[11.7, 19.3]} 
\defnum{sk.loss.P.lr}{19.4} 
\defnum{sk.loss.P.lr.ci}{[15.4, 23.3]} 
\defnum{sk.ex.lr}{$-$4.2} 
\defnum{sk.ex.lr.ci}{[$-$9.3, 0.9]} 
\defnum{sk.OD.lr}{24.2} 
\defnum{sk.OD.lr.ci}{[20.4, 27.9]} 
\defnum{sk.fix.lr}{$-$12.1} 
\defnum{sk.OD.it}{0.3} 
\defnum{sk.OD.it.ci}{[0.0, 0.8]} 
\defnum{sk.share.D.lr}{29.9} 
\defnum{sk.share.O.lr}{23.4} 
\defnum{sk.share.P.lr}{33.4} 

%% file: numbers.revision2.tex
\defnum{rb.rh.it.with}{91.5} 
\defnum{rb.rh.it.without}{59.3} 
\defnum{rb.rh.it.loss}{3.8} 
\defnum{rb.rh.it.loss.ci}{[1.8, 5.9]} 
\defnum{rb.rh.it.loss.without}{36.1} 
\defnum{rb.rh.it.cut}{89} 
\defnum{rb.rh.b.with}{55.8} 
\defnum{rb.rh.b.without}{22.5} 
\defnum{rb.rh.b.loss}{39.6} 
\defnum{rb.rh.b.loss.ci}{[35.7, 43.6]} 
\defnum{rb.rh.b.loss.without}{72.9} 
\defnum{rb.rh.b.cut}{46} 
\defnum{rb.wc.s2-P-u140}{0.60} 
\defnum{rb.wc.D-T-u20}{0.27} 
\defnum{rb.wc.s2-D-u140}{2.22} 
\defnum{rb.wc.D-r128-u140}{2.05} 
\defnum{rb.wc.O-T-u140}{0.80} 
\defnum{rb.wc.O-u60}{0.40} 
\defnum{rb.wc.O-sharp-u140}{0.61} 
\defnum{rb.wc.D-T-u140}{2.81} 
\defnum{rb.wc.D-u20}{0.25} 
\defnum{rb.wc.s2-O-u140}{0.63} 
\defnum{rb.wc.D2-u140}{2.50} 
\defnum{rb.wc.s3-O-u140}{0.62} 
\defnum{rb.wc.D-u60}{0.75} 
\defnum{rb.wc.s3-D-u140}{2.27} 
\defnum{rb.wc.s3-P-u140}{0.68} 
\defnum{rb.wc.P-T-u140}{1.17} 
\defnum{rb.wc.D-u140}{2.40} 
\defnum{rb.wc.O-u140}{0.69} 
\defnum{rb.wc.P-u140}{0.65} 
\defnum{rb.wc.D-u91}{1.35} 
\defnum{rb.wc.drilled.lossmin}{12} 
\defnum{rb.wc.drilled.lossmax}{62} 
\defnum{rb.wc.other.lossmax}{1.7} 
\defnum{rb.wc.drilled.min}{2.0} 
\defnum{rb.wc.drilled.max}{2.8} 
\defnum{rb.wc.other.max}{1.2} 
\defnum{rb.wc.judged.n}{18} 
\defnum{rb.wc.ratio}{3.5} 
\defnum{rb.am.D.it.u1}{0.22} 
\defnum{rb.am.D.it.u2}{0.39} 
\defnum{rb.am.D.it.u5}{0.80} 
\defnum{rb.am.O.it.u1}{0.11} 
\defnum{rb.am.O.it.u2}{0.18} 
\defnum{rb.am.O.it.u5}{0.33} 
\defnum{rb.am.P.it.u1}{0.11} 
\defnum{rb.am.P.it.u2}{0.18} 
\defnum{rb.am.P.it.u5}{0.33} 
\defnum{rb.am.D.gentle.u20}{0.76} 
\defnum{rb.am.D.gentle.u60}{1.31} 
\defnum{rb.am.O.gentle.u20}{0.28} 
\defnum{rb.am.O.gentle.u60}{0.53} 
\defnum{rb.am.D.normA}{59.2} 
\defnum{rb.am.D.normB}{27.8} 
\defnum{rb.am.O.normA}{52.0} 
\defnum{rb.am.O.normB}{14.2} 
\defnum{rb.am.P.normA}{52.0} 
\defnum{rb.am.P.normB}{13.5} 
\defnum{rb.am.D.it.u5.nohead}{0.54} 
\defnum{rb.am.O.it.u5.nohead}{0.31} 
\defnum{rb.am.ratio5.nohead}{1.7} 
\defnum{rb.am.ratio5.head}{3.6} 
\defnum{rb.am.mean.ratio5}{1.6} 
\defnum{rb.am.mean.ratio1}{1.5} 
\defnum{rb.am.mean.ratio60}{1.6} 
\defnum{rb.am.mean.ratioP5}{1.00} 
\defnum{rb.fn.s2-P-u140}{7.7} 
\defnum{rb.fn.D-T-u20}{3.9} 
\defnum{rb.fn.s2-D-u140}{15.2} 
\defnum{rb.fn.O-T-u140}{8.2} 
\defnum{rb.fn.O-sharp-u140}{7.9} 
\defnum{rb.fn.D-T-u140}{14.4} 
\defnum{rb.fn.D-u20}{3.7} 
\defnum{rb.fn.s2-O-u140}{7.7} 
\defnum{rb.fn.D2-u140}{15.1} 
\defnum{rb.fn.s3-O-u140}{7.7} 
\defnum{rb.fn.D-u60}{8.5} 
\defnum{rb.fn.s3-D-u140}{15.0} 
\defnum{rb.fn.s3-P-u140}{7.7} 
\defnum{rb.fn.P-T-u140}{8.6} 
\defnum{rb.fn.D-u140}{14.9} 
\defnum{rb.fn.O-u140}{7.7} 
\defnum{rb.fn.P-u140}{7.7} 
\defnum{rb.fn.D-u91}{11.7} 
\defnum{rb.fn.ratio}{1.9} 
\defnum{rb.fn.drilled.min}{14.4} 
\defnum{rb.fn.drilled.max}{15.2} 
\defnum{rb.fn.other.max}{8.6} 
\defnum{rb.am.ratioB}{2.0} 
\defnum{rb.am.ratio5}{2.4} 
\defnum{rb.am.ratio1}{2.0} 
\defnum{rb.am.ratio60}{2.5} 
\defnum{rb.rc.h.base}{94.3} 
\defnum{rb.rc.h.D}{95.7} 
\defnum{rb.rc.h.O}{95.0} 
\defnum{rb.rc.h.P}{93.6} 
\defnum{rb.rc.tinker.base}{94.7} 
\defnum{rb.rc.tinker.D}{95.4} 
\defnum{rb.rc.tinker.O}{94.6} 
\defnum{rb.rc.tinker.P}{94.6} 
\defnum{rb.rc.epoch.D}{96.2} 
\defnum{rb.rc.epoch.O}{95.7} 
\defnum{rb.rc.epoch.P}{94.5} 
\defnum{rb.rc.epoch.ex.D}{0.2} 
\defnum{rb.rc.epoch.ex.D.ci}{[$-$2.0, 2.5]} 
\defnum{rb.rc.epoch.ex.P}{$-$0.2} 
\defnum{rb.rc.epoch-const.D}{96.2} 
\defnum{rb.rc.epoch-const.loss.D}{$-$0.5} 
\defnum{rb.rc.epoch-const.loss.D.ci}{[$-$2.0, 1.1]} 
\defnum{rb.rc.epoch-const.O}{95.2} 
\defnum{rb.rc.epoch-const.ex.D}{$-$0.2} 
\defnum{rb.rc.epoch-full.D}{93.8} 
\defnum{rb.rc.epoch-full.loss.D}{1.9} 
\defnum{rb.rc.epoch-full.loss.D.ci}{[0.1, 3.7]} 
\defnum{rb.rc.epoch-full.O}{94.6} 
\defnum{rb.rc.epoch-full.loss.O}{0.5} 
\defnum{rb.rc.epoch-full.loss.O.ci}{[$-$1.2, 2.0]} 
\defnum{rb.rc.epoch-full.ex.D}{1.5} 
\defnum{rb.rc.epoch-full.ex.D.ci}{[$-$0.9, 4.0]} 
\defnum{rb.rc.it-stress.D}{86.4} 
\defnum{rb.rc.it-stress.loss.D}{9.3} 
\defnum{rb.rc.it-stress.loss.D.ci}{[6.7, 11.9]} 
\defnum{rb.rc.it-stress.O}{95.1} 
\defnum{rb.rc.it-stress.loss.O}{$-$0.1} 
\defnum{rb.rc.it-stress.loss.O.ci}{[$-$1.4, 1.0]} 
\defnum{rb.rc.it-stress.P}{94.3} 
\defnum{rb.rc.it-stress.loss.P}{$-$0.8} 
\defnum{rb.rc.it-stress.loss.P.ci}{[$-$2.5, 0.8]} 
\defnum{rb.rc.it-stress.ex.D}{9.4} 
\defnum{rb.rc.it-stress.ex.D.ci}{[6.4, 12.3]} 
\defnum{rb.rc.b-stress.D}{3.5} 
\defnum{rb.rc.b-stress.loss.D}{92.2} 
\defnum{rb.rc.b-stress.O}{13.3} 
\defnum{rb.rc.b-stress.loss.O}{81.7} 
\defnum{rb.rc.b-stress.P}{35.4} 
\defnum{rb.rc.b-stress.loss.P}{58.1} 
\defnum{rb.rc.b-stress.lessP}{23.5} 
\defnum{rb.rc.b-stress.ex.D}{10.5} 
\defnum{rb.rc.b-stress.ex.D.ci}{[7.4, 13.7]} 
\defnum{rb.rc.it-stress-full.D}{31.8} 
\defnum{rb.rc.it-stress-full.loss.D}{63.9} 
\defnum{rb.rc.it-stress-full.loss.D.ci}{[60.2, 67.6]} 
\defnum{rb.rc.it-stress-full.O}{50.0} 
\defnum{rb.rc.it-stress-full.loss.O}{45.0} 
\defnum{rb.rc.it-stress-full.loss.O.ci}{[41.0, 49.1]} 
\defnum{rb.rc.it-stress-full.P}{61.1} 
\defnum{rb.rc.it-stress-full.loss.P}{32.5} 
\defnum{rb.rc.it-stress-full.loss.P.ci}{[28.6, 36.2]} 
\defnum{rb.rc.it-stress-full.lessP}{12.6} 
\defnum{rb.rc.it-stress-full.ex.D}{18.9} 
\defnum{rb.rc.it-stress-full.ex.D.ci}{[14.1, 23.5]} 
\defnum{rb.rc.it-stress-full.ex.P}{$-$12.6} 
\defnum{rb.rc.epoch-cont-peft.D}{89.0} 
\defnum{rb.rc.epoch-cont-peft.loss.D}{6.7} 
\defnum{rb.rc.epoch-cont-peft.loss.D.ci}{[4.3, 9.2]} 
\defnum{rb.rc.epoch-const-head.D}{97.2} 
\defnum{rb.rc.epoch-const-head.loss.D}{$-$1.5} 
\defnum{rb.rc.epoch-const-head.loss.D.ci}{[$-$3.1, $-$0.1]} 
\defnum{rb.rc.move.epoch.D}{0.40} 
\defnum{rb.rc.move.epoch.O}{0.38} 
\defnum{rb.rc.move.epoch.ratio}{1.05} 
\defnum{rb.rc.move.it-stress.D}{0.64} 
\defnum{rb.rc.move.it-stress.O}{0.62} 
\defnum{rb.rc.move.it-stress.ratio}{1.03} 
\defnum{rb.sl.D-u140}{36.1} 
\defnum{rb.sl.D-u60}{1.1} 
\defnum{rb.sl.D-u20}{0.9} 
\defnum{rb.sl.O-u140}{$-$0.1} 
\defnum{rb.sl.D2-u140}{12.4} 
\defnum{rb.sl.D-r128-u140}{38.8} 
\defnum{rb.sl.P-u140}{$-$0.3} 
\defnum{rb.sl.D-T-u140}{61.3} 
\defnum{rb.sl.D-T-u20}{0.8} 
\defnum{rb.sl.O-T-u140}{1.7} 
\defnum{rb.sl.P-T-u140}{0.7} 
\defnum{rb.sl.O-sharp-u140}{$-$1.9} 
\defnum{rb.sl.s2-D-u140}{62.1} 
\defnum{rb.sl.s2-O-u140}{1.0} 
\defnum{rb.sl.s2-P-u140}{1.0} 
\defnum{rb.sl.s3-D-u140}{48.6} 
\defnum{rb.sl.s3-O-u140}{$-$0.2} 
\defnum{rb.sl.s3-P-u140}{$-$0.2} 
\defnum{rb.open.base}{39} 
\defnum{rb.open.trained}{100} 
\defnum{rb.reg.total}{17} 
\defnum{rb.reg.failed}{7} 
\defnum{rb.co.outcome}{held} 
\defnum{rb.comp.tinker}{51} 
\defnum{rb.comp.modal}{90} 
\defnum{rb.full.lr}{$10^{-5}$} 
\defnum{rb.p5.passes}{5.0} 
\defnum{rb.is.D.n}{575} 
\defnum{rb.is.O.n}{4,441} 
\defnum{rb.p5.pred}{1.4} 
\defnum{rb.p5.loss}{9.5} 
\defnum{rb.p5.loss.ci}{[6.6, 12.4]} 
\defnum{rb.p5.ex}{9.6} 
\defnum{rb.cs.constant.loss}{9.4} 
\defnum{rb.cs.constant.loss.ci}{[6.8, 12.0]} 
\defnum{rb.cs.standard}{86.2} 
\defnum{rb.cs.standard.loss}{9.2} 
\defnum{rb.cs.standard.loss.ci}{[6.4, 11.9]} 
\defnum{rb.cs.once-standard}{94.6} 
\defnum{rb.cs.once-standard.loss}{0.0} 
\defnum{rb.cs.once-standard.loss.ci}{[$-$1.4, 1.5]} 
\defnum{rb.cs.standard.ex}{9.2} 
\defnum{rb.cs.standard.ex.ci}{[6.2, 12.2]} 
\defnum{rb.rc.check.D-u140}{0.015} 
\defnum{rb.rc.check.M0}{0.008} 
\defnum{rb.rc.check.O-u140}{0.008} 
\defnum{rb.rc.check.P-u140}{0.008} 
\defnum{rb.em.it.plain}{57.7} 
\defnum{rb.em.it.masked}{74.1} 
\defnum{rb.em.it.share}{0.43} 
\defnum{rb.em.it.pct}{43} 
\defnum{rb.em.it.share.ci}{[0.35, 0.51]} 
\defnum{rb.em.it.first}{71.5} 
\defnum{rb.em.it.share.first}{0.36} 
\defnum{rb.em.it.share.firstall}{0.74} 
\defnum{rb.em.it.share.firstall.ci}{[0.58, 0.91]} 
\defnum{rb.em.it.handoff.first}{77.8} 
\defnum{rb.em.it.closes}{91} 
\defnum{rb.em.it.closes.plain}{26} 
\defnum{rb.em.b.plain}{23.5} 
\defnum{rb.em.b.masked}{19.9} 
\defnum{rb.em.b.share}{$-$0.05} 
\defnum{rb.em.b.pct}{$-$5} 
\defnum{rb.em.b.share.ci}{[$-$0.11, 0.01]} 
\defnum{rb.em.b.first}{57.5} 
\defnum{rb.em.b.share.first}{0.47} 
\defnum{rb.em.b.share.first.ci}{[0.42, 0.52]} 
\defnum{rb.em.b.share.firstall}{0.64} 
\defnum{rb.em.b.share.firstall.ci}{[0.58, 0.72]} 
\defnum{rb.em.b.closes}{3} 
\defnum{rb.em.b.boxes}{180} 
\defnum{rb.em.b.capped}{97} 
\defnum{rb.is.student.free}{69.5} 
\defnum{rb.is.student.forced}{69.6} 
\defnum{rb.is.student.epoch}{25.3} 
\defnum{rb.is.student.epoch.loss}{44.2} 
\defnum{rb.is.student.epoch.loss.ci}{[39.6, 48.9]} 
\defnum{rb.is.student.epochfree}{27.3} 
\defnum{rb.is.student.epochfree.loss}{42.2} 
\defnum{rb.is.student.epochfree.loss.ci}{[37.8, 46.6]} 
\defnum{rb.is.forced.h.D}{65.2} 
\defnum{rb.is.forced.h.O}{61.4} 
\defnum{rb.is.forced.h.P}{64.8} 
\defnum{rb.is.forced.epoch.D}{32.2} 
\defnum{rb.is.forced.epoch.O}{28.8} 
\defnum{rb.is.forced.epoch.ex.D}{0.3} 
\defnum{rb.is.forced.epoch.ex.D.ci}{[$-$4.3, 5.1]} 
\defnum{rb.is.forced.it-stress.D}{27.6} 
\defnum{rb.is.forced.it-stress.O}{26.8} 
\defnum{rb.is.forced.it-stress.P}{22.6} 
\defnum{rb.is.forced.it-stress.ex.D}{2.9} 
\defnum{rb.is.forced.it-stress.ex.D.ci}{[$-$1.7, 7.6]} 
\defnum{rb.is.forced.it-stress.ex.P}{7.6} 
\defnum{rb.is.forced.it-stress.ex.P.ci}{[2.8, 12.3]} 
\defnum{rb.is.free.h.D}{64.9} 
\defnum{rb.is.free.h.O}{63.2} 
\defnum{rb.is.free.epoch.D}{25.0} 
\defnum{rb.is.free.epoch.O}{20.5} 
\defnum{rb.is.free.epoch.ex.D}{$-$2.8} 
\defnum{rb.is.free.epoch.ex.D.ci}{[$-$7.1, 1.5]} 
\defnum{rb.is.free.it-stress.D}{21.3} 
\defnum{rb.is.free.it-stress.O}{21.5} 
\defnum{rb.is.free.it-stress.P}{21.9} 
\defnum{rb.is.free.it-stress.ex.D}{1.9} 
\defnum{rb.is.free.it-stress.ex.D.ci}{[$-$2.4, 6.2]} 
\defnum{rb.ic.h.D}{66.9} 
\defnum{rb.ic.h.O}{63.0} 
\defnum{rb.ic.it-stress.D}{1.6} 
\defnum{rb.ic.it-stress.O}{5.3} 
\defnum{rb.ic.it-stress.ex.D}{7.6} 
\defnum{rb.ic.it-stress.ex.D.ci}{[4.1, 11.2]} 
\defnum{rb.is.gain.O}{$-$8.0} 

%% file: sections/introduction.tex
\section{Introduction}
\label{sec:intro}

\begin{figure}[t]
\centering
\figfile{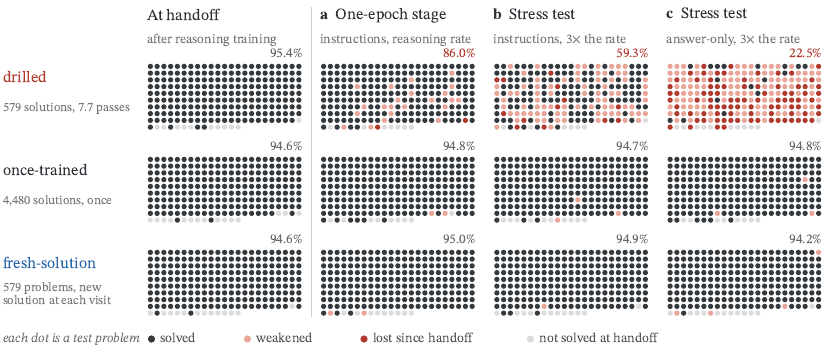}{0.86\linewidth}
\caption{Equal at handoff, unequal after a later stage. Each dot is one of the
\nv{d.math.eval} held-out MATH-500 problems, in the same place in every grid (easiest
first); dark: solved in at least 3 of 4 samples; percentages: forced accuracy. Each
later stage continues the model's reasoning adapter on data that does not teach math.
The once-trained and fresh-solution models still solve the problems. The drilled
model loses many (red: lost since handoff, solved in none of 4 samples; light red:
weakened, solved in 1 or 2), the more so the harsher the stage: (a)~the one-epoch
stage; (b)~the instruction stress test; (c)~the answer-only stress test
(Table~\ref{tab:stages}).}
\label{fig:teaser}
\end{figure}

Teaching a language model to reason can take very little data. Recipes
such as s1 and LIMO fine-tune a model on a thousand worked solutions or fewer, and
make up for the small number by going over them again and again: s1 trains on its
set \nv{lit.s1.epochs} times and LIMO \nv{lit.limo.epochs} times
\citep{muennighoff2025s1,ye2025limo}. Judged at \emph{handoff}, as we will call the
point where this training ends, the repetition looks free and sometimes helpful:
\citet{kopiczko2026repetition} found that \nv{lit.kop.n} examples repeated many
times over beat \nv{lit.kop.ratio} times as many examples seen once. But handoff is
rarely the end. A reasoning model is usually trained again, by its developers in a
later stage or by whoever adapts it next, often on data that has nothing to do with
reasoning. A handoff score cannot say how much of the reasoning survives that. We
asked whether repeating the same solutions makes the reasoning more fragile to the
next stage.

To isolate the repetition we hold everything else fixed. We take Qwen3.5-9B-Base,
which already solves \nv{m.h.base}\% of our held-out MATH-500 problems when it
reasons, and train it with a LoRA adapter on its own correct solutions to other
problems. The training therefore teaches no new mathematics, and any later difference
between the trained models comes from how the texts were presented. We present such solutions in two ways, for the same number of updates. The \emph{once-trained} model sees
\nv{d.O.n} solutions, once each. The \emph{drilled} model sees \nv{d.drill.n}
solutions, about eight times each. At handoff their accuracy cannot tell the two apart. Both
solve about 95\% of the test problems when we open their reasoning block for them,
the block in which the model reasons before it answers (\emph{forced} accuracy).
Both also open it by themselves in \nv{rb.open.trained}\% of their answers, where the
base does so in \nv{rb.open.base}\%. So on these problems the training changed when the model
reasons, not how well.

We then train each model a little further on something unrelated, continuing the
same adapter. If repetition were harmless, the two should come through alike. They
do not. After one epoch of everyday instructions at the learning rate of the
reasoning training (the \emph{one-epoch stage}), the once-trained model is where it
was. The drilled model has fallen from \nv{rs.D.h}\% to \nv{rs.D.a}\%
(Figure~\ref{fig:teaser}a), under either learning-rate schedule we tried. Two harsher
stages, twenty updates at three times that rate on the same instructions or on
puzzles with bare answers (the instruction and answer-only \emph{stress tests}), take it to \nv{m.it.D140}\% and \nv{m.b.D140}\%. But the drilled model also
saw far fewer problems. Is that why? A third, \emph{fresh-solution} model visits the
drilled model's problems as often, with a new correct solution at each visit. It came
through every one of these stages unharmed. So the break comes from seeing the same
texts again, not from seeing few problems. The pattern recurs with a stronger model's
traces, in two more runs, at 35B, in another model family and on an arithmetic task
(Section~\ref{sec:math}).

Continuing the adapter is what happens when several stages are trained in one
adapter, or when a released adapter is trained further. A downstream user often does
something else: they receive the weights with the adapter merged in, and train a new
adapter on them or fine-tune all of them (Section~\ref{sec:released}). Started that
way, the drilled model survives the one-epoch stage, but with a new adapter the
instruction stress test still costs it \nv{rb.rc.it-stress.ex.D} points more than the once-trained model. So
one run suggests that the merged drilled weights are fragile in themselves, and that
continuing the drilled adapter amplifies this. Section~\ref{sec:signature} offers an
account of the amplification, consistent with the measurements but not tested
causally. An adapter adds the product of two small matrices to each weight matrix it
adapts, and each update is scaled by their size. Drilling makes them larger, so later
updates move the drilled weights \nv{rb.am.mean.ratio5} times as far as the once-trained
ones. A new adapter starts equally small in every model, so it moves both alike.

What the later stage takes is suppressed rather than erased
(Section~\ref{sec:lost}): forbidding the drilled model to end its turn, or nudging it
on with ``Wait'', wins back less than half of the loss, while five updates of
reasoning training win back nearly all of it. The break can also be prevented or much
reduced (Section~\ref{sec:remedies}). A new solution at every visit prevented it in
every stage that continued the 9B adapter. Replaying \nv{d.RP.replay}\% of the
original solutions prevented it in a gentler instruction stage and cut it by
\nv{rb.rh.it.cut}\% and \nv{rb.rh.b.cut}\% in the two stress tests. All of this concerns
reasoning the model could already do, which s1 and LIMO aim to draw out. On a
synthetic skill the base model lacked, repetition mainly cost learning, and in a later
stage the fresh-solution model lost the most.

%% file: sections/setup.tex
\section{Method}
\label{sec:setup}

We train three copies of one base model on its own correct solutions to math
problems for the same number of updates, varying only how often each problem and
each solution text is seen. We then give them and the untrained base model the same
later stage of training on data not chosen to teach math. 

\subsection{Models and training}

Most experiments use Qwen3.5-9B-Base \citep{qwen2026qwen35}. Except for full fine-tuning in Section~\ref{sec:released}, all training, the later stages included, learns a LoRA adapter \citep{hu2022lora} of rank \nv{d.rank}
(\nv{d.rank.alt} in one repeat) while the base weights stay fixed. Unless we say
otherwise it runs on Tinker \citep{tml2025tinker}, a hosted training service. We repeated the main design
on Qwen3.5-35B-A3B-Base and on Nemotron~3 Nano \citep{nvidia2025nemotron3nano}, an
already post-trained hybrid Mamba--Transformer from another family, both mixtures of
experts with about \nv{d.active}B active parameters.

\subsection{Three ways to train on the same solutions}

The training texts are the base model's own correct solutions, sampled with the
reasoning block open, to math problems from NuminaMath-TIR in the T\"ulu~3 mixture
\citep{numina2024tir,lambert2024tulu3} (Appendix~\ref{app:mathdata}). All three
models train for \nv{d.drill.updates} updates of \nv{d.stage.batch} examples, \nv{d.O.n} in all, at
learning rate \nv{d.acq.lr}, which we
call the \emph{reasoning rate} (Table~\ref{tab:design}). The \emph{drilled} model cycles through \nv{d.drill.n}
solutions to as many problems and so sees each solution \nv{d.pass.high} times on
average; we also keep it after about one and three passes. The \emph{once-trained}
model sees \nv{d.O.n} solutions to as many problems, each once.

These two differ both in how often each text is seen and in how many problems there
are. The \emph{fresh-solution} model separates the two: it visits the drilled
model's problems in the same order and as often, with a different correct solution
at each visit, so it sees each text about once (Table~\ref{tab:arms} lists every
model).

\begin{table}[t]
\centering
\caption{The three ways of training on the base model's own solutions, each for
\nv{d.drill.updates} updates of \nv{d.stage.batch} examples. Visits: times each
problem is seen; passes: times each solution text is seen, on average.}
\label{tab:design}
\footnotesize
\begin{tabular}{@{}l l r r r r@{}}
\toprule
Model & Solutions & Problems & Distinct texts & Visits & Passes \\
\midrule
Drilled & one per problem & \nv{d.drill.n} & \nv{d.drill.n} & \nv{d.pass.high} & \nv{d.pass.high} \\
Once-trained & one per problem & \nv{d.O.n} & \nv{d.O.n} & \nv{d.O.pass.high} & \nv{d.O.pass.high} \\
Fresh-solution & a new one at each visit & \nv{d.drill.n} & up to \nv{d.P.kept} per problem & \nv{d.pass.high} & \nv{d.P.passes} \\
\bottomrule
\end{tabular}

\vspace{6pt}
\caption{The later stages. None trains the model to solve math problems.}
\label{tab:stages}
\begin{tabular}{@{}l l l r l@{}}
\toprule
Stage & Data & Rate & Updates & Why \\
\midrule
One-epoch stage & \nv{rs.rows} No Robots instructions & \nv{d.acq.lr} & \nv{rs.updates} & plausible next stage \\
Instruction stress test & \nv{d.norobots} No Robots instructions & \nv{d.stage.lr} & \nv{d.stage.updates} & magnifies differences \\
Answer-only stress test & puzzles with bare answers & \nv{d.stage.lr} & \nv{d.stage.updates} & rewards skipping reasoning \\
Gentler stage & \nv{r2.examples} No Robots instructions & \nv{d.stage.lrlow} & \nv{d.R2.updates} & same dose, reasoning rate \\
\bottomrule
\end{tabular}
\end{table}

\subsection{Later stages}

After handoff, each trained model and the untrained base model get a \emph{later
stage}: a second round of training on data not chosen to teach math
(Table~\ref{tab:stages}). Each trained model continues its own adapter, and the base
starts from a new one. The \emph{one-epoch stage}, our realistic case, is one epoch of
human-written instructions from No Robots \citep{rajani2023norobots} at the reasoning
rate. Two \emph{stress tests} run twenty updates at three times that rate, to make
the differences large enough to study: the \emph{instruction stress test} on the same
instructions, and the \emph{answer-only stress test} on program-synthesis puzzles
with bare answers as targets, which rewards answering without thinking first. The
\emph{gentler stage} runs \nv{d.R2.updates} updates at the reasoning rate, a third of the
stress tests' rate for three times as many updates (Appendix~\ref{app:stages}).

\subsection{What we measure}

The test set is the \nv{d.math.eval} problems of MATH-500 at difficulty levels 3 to
5 that have integer answers \citep{hendrycks2021math,lightman2024verify}, with
\nv{d.math.aime} AIME 2025 and 2026 problems from MathArena as a harder check
\citep{balunovic2025matharena,dekoninck2026matharena}. We sample four answers per
problem. \emph{Forced accuracy} is the share of correct answers when we start the
response with \think, the opening of the reasoning block, so that the model always
reasons first; \emph{free accuracy} is the share when the model completes the prompt
however it likes. Forced accuracy is our main score because a later stage can make a
model stop choosing to reason without removing its ability to, and free accuracy
then hides that ability (Section~\ref{sec:lost}).

A model's \emph{loss} in a later stage is its drop in forced accuracy from its own
handoff, and its \emph{extra loss} is how much more it ($X$) lost than a comparison model ($Y$):
$(h_X - e_X) - (h_Y - e_Y)$ points, where $h$ is forced accuracy at handoff and $e$
after the stage. Unless we say otherwise, $Y$ is the once-trained model after the same
stage. In the one-epoch stage, for example, the drilled model fell from
\nv{rs.D.h}\% to \nv{rs.D.a}\% while the once-trained model rose from \nv{rs.O.h}\%
to \nv{rs.O.a}\%, so the drilled model's extra loss is \nv{rs.ex.D} points. Intervals are 95\% paired
bootstrap intervals over test problems; they cover the choice of test problems but not
another training run, which is why we repeated the 9B design.
Appendix~\ref{app:ledger} lists the predictions, with thresholds, that we registered
in advance: \nv{reg.total} for the original experiments, of which \nv{reg.failed}
failed, and \nv{rb.reg.total} for those added in revision, of which
\nv{rb.reg.failed} failed.

%% file: sections/fragile.tex
\section{Continuing a drilled adapter breaks the reasoning}
\label{sec:math}

\begin{table}[t]
\centering
\caption{The models are nearly equal at handoff; only the models drilled for about eight
passes fall. Forced accuracy (\%) of the 9B models at handoff and
after each later stage (Table~\ref{tab:stages}). Passes: times the average training
text was seen; the drilled rows at fewer passes are the same run saved earlier.
Sharpened: trained like the once-trained model on solutions sampled at temperature
\nv{d.SH.temp} (Section~\ref{sec:signature}). Dashes: not run. Intervals, free
accuracy and the other models: Tables~\ref{tab:excess} and~\ref{tab:full}.}
\label{tab:main}
\small
\begin{tabular}{@{}l l r r r r r@{}}
\toprule
& & & & One-epoch & \multicolumn{2}{c}{Stress tests} \\
\cmidrule(l){6-7}
Training texts & Model & Passes & Handoff & stage & Instruction & Answer-only \\
\midrule
None & Base & -- & \nv{m.h.base} & \nv{rs.base.a} & \nv{m.it.base} & \nv{m.b.base} \\
\addlinespace
Own solutions & Drilled & \nv{d.pass.low} & \nv{m.h.D20} & -- & \nv{m.it.D20} & \nv{m.b.D20} \\
& Drilled & \nv{d.pass.mid} & \nv{m.h.D60} & -- & \nv{m.it.D60} & \nv{m.b.D60} \\
& Drilled & \nv{d.pass.high} & \nv{m.h.D140} & \nv{rs.D.a} & \nv{m.it.D140} & \nv{m.b.D140} \\
& Once-trained & \nv{d.O.pass.high} & \nv{m.h.O140} & \nv{rs.O.a} & \nv{m.it.O140} & \nv{m.b.O140} \\
& Fresh-solution & \nv{d.P.passes} & \nv{mm.P.h} & \nv{rs.P.a} & \nv{mm.P.it} & \nv{mm.P.b} \\
& Sharpened & \nv{d.O.pass.high} & \nv{sh.h} & -- & \nv{sh.it} & \nv{sh.b} \\
\addlinespace
gpt-oss-120b & Drilled & \nv{mt.pass.D} & \nv{mt.D.h} & -- & \nv{mt.D.it} & \nv{mt.D.b} \\
traces & Once-trained & \nv{d.O.pass.high} & \nv{mt.O.h} & -- & \nv{mt.O.it} & \nv{mt.O.b} \\
& Fresh-solution & \nv{d.P.passes} & \nv{mt.P.h} & -- & \nv{mt.P.it} & \nv{mt.P.b} \\
\bottomrule
\end{tabular}
\end{table}

\paragraph{Handoff and the later stages.}
At handoff the models barely differ: in no run or model on mathematics did the drilled
and once-trained models differ by more than \nv{m35.parity} points of forced accuracy
(Table~\ref{tab:main}). Later training tells them apart. In the one-epoch stage the base,
once-trained and fresh-solution models end within a point of where they started,
while the drilled model's extra loss is \nv{rs.ex.D} points (interval
\nv{rs.ex.D.ci}). The stress tests raise it to \nv{mex.O.it} points in the
instruction stress test and \nv{mex.O.b} in the answer-only stress test. On AIME, a harder test, the instruction stress test took the drilled model from
\nv{aime.h.D140}\% to \nv{aime.it.D140}\% and the once-trained one from
\nv{aime.h.O140}\% to \nv{aime.it.O140}\% (only \nv{d.math.aime} problems, so descriptive; Appendix~\ref{app:aime}).

\paragraph{Repeated texts or few problems?}
The drilled model differs from the once-trained one in two ways: it repeats its
texts, and it has far fewer problems. If having few problems were the trouble, the
fresh-solution model, which has exactly the drilled model's problems and visits,
would break too. It does not. After the one-epoch stage and both stress tests it
stays within a point of the once-trained model, with intervals that include zero. In
the stress tests the drilled model loses \nv{mm.DP.it} and \nv{mm.DP.b} points more
than the fresh-solution model. Both gaps clear the bar we had fixed in advance for
naming the repeated texts as the cause (\nv{reg.M3.2.thr} points, interval above
zero). By that criterion, the break comes from repeating the texts.

\begin{figure}[t]
\centering
\figfile{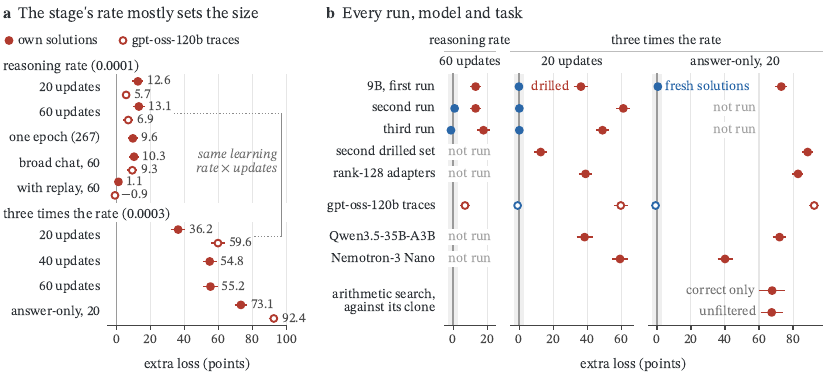}{0.86\linewidth}
\caption{The drilled model's extra loss over the once-trained model,
in points with 95\% intervals (positive: the drilled model lost more). (a)~The 9B
model after each later stage, on instructions unless named; filled: own solutions,
hollow: gpt-oss-120b's traces; replay: Section~\ref{sec:remedies}. (b)~Every run and
model on mathematics, and the arithmetic task, where the comparison is the teacher's
clone and the drilled models saw all of the teacher's samples (unfiltered) or the
correct ones only; at 35B and on Nemotron the stress tests also cost the base and
once-trained models accuracy. Blue: fresh-solution models; grey: the margin of
$\pm$\nv{reg.RS.P} points fixed in advance.}
\label{fig:replication}
\end{figure}

\paragraph{How many passes, and how harsh a stage.}
\label{sec:dose}
At 9B the break grows with the number of passes: the drilled checkpoints saved after
one and three passes lose no more than about three points in either stress test, and
a drilled model stopped after five passes, the number of epochs s1 uses, breaks only
partly (Section~\ref{sec:signature}). At 35B and on Nemotron the three-pass
checkpoint already broke in one stress test (Appendix~\ref{app:scale}). On instruction
data, the size of the break depends mostly on the learning rate: at the reasoning rate
the extra loss stays at about 10 to 13 points from twenty updates to a full epoch, on
the instructions and on a \emph{broad chat mix} from non-math sources of T\"ulu~3,
against \nv{mex.O.it} at three times the rate
(Figure~\ref{fig:replication}a; Appendix~\ref{app:stage}).

\paragraph{A stronger model's traces.}
s1 and LIMO repeat a stronger model's long traces, not the model's own solutions. So
we trained the three models again on gpt-oss-120b's traces
\citep{openai2025gptoss} for problems from the same source. In the stress tests the
result is the same, and larger: only the drilled model breaks (an extra loss of
\nv{mt.ex.O.it} points in the instruction stress test, against \nv{mex.O.it} on its own
solutions), and a fresh trace at every visit prevents it. In the gentler stage it was
smaller: \nv{r4.ex60} points, short of the \nv{r4.req} we had predicted
(Appendix~\ref{app:scale}).

\paragraph{Other runs and an arithmetic task.}
Our intervals do not cover the randomness of training, so
we ran the 9B design twice more, with new drilled problems, orders and
initializations. In both runs, after the instruction stress test and the gentler
stage, the drilled model again lost more than the other two, with every interval
above zero, and the fresh-solution model stayed within a point and a half of the
once-trained one (Appendix~\ref{app:runs}). The direction is stable; the size is
not. In the instruction stress test the extra loss ran from \nv{mm.D2O.it} points for
a second set of drilled problems in the first run to \nv{sd.s2.ex.it} in a new run, almost
fivefold, and nothing we measured at handoff predicts which. We therefore claim the direction
of the effect and not its size.

The break also appears outside mathematics. On an arithmetic search task, a variant
of the Countdown game \citep{gandhi2024sos}, the training texts are samples from a
teacher trained by reinforcement learning. Models drilled on these samples, on all of
them or on the correct ones only, lose \nv{tces16loss.U140} and
\nv{tces16loss.F140} points in the answer-only stress test. The teacher, its starting
model and a clone of the teacher that saw each sample about once (this task's once-trained
model) lose at most
\nv{tces16.lastingloss.max} (Appendix~\ref{app:tces}). In every run and model of
Figure~\ref{fig:replication}b, all of which continue the adapter, the drilled model's
interval lies above zero.

%% file: sections/released.tex
\section{Starting again from released weights}
\label{sec:released}

\begin{table}[t]
\centering
\caption{Continuing the drilled adapter breaks it under either schedule; a new
adapter protects it in the one-epoch stage but not in the instruction stress test.
Entries: forced accuracy (\%) after each later stage.  \emph{New adapter}: the
reasoning adapter is merged into the base weights and a newly initialized one, or
every weight of the text model, is trained, in different software
(Appendix~\ref{app:released}); these models start from \nv{rb.rc.h.D}\% (drilled),
\nv{rb.rc.h.O}\% (once-trained) and \nv{rb.rc.h.P}\% (fresh-solution).  Extra: the drilled model's
extra loss, in points. $^{\dagger}$Every model loses much of its accuracy and the two
models without repetition differ, so these stages cannot isolate repetition.
$^{\ast}$One run on the standard schedule at peak $3\times10^{-5}$, rerun after the
registered constant-rate optimizer diverged (Appendix~\ref{app:released}).
Dashes: not run.}
\label{tab:released}
\footnotesize
\begin{tabular}{@{}l rrrr rrrr@{}}
\toprule
& \multicolumn{4}{c}{Continued adapter} & \multicolumn{4}{c}{New adapter} \\
\cmidrule(lr){2-5}\cmidrule(l){6-9}
Later stage & Drilled & Once & Fresh & Extra & Drilled & Once & Fresh & Extra \\
\midrule
One epoch, constant rate & \nv{rs.D.a} & \nv{rs.O.a} & \nv{rs.P.a} & \nv{rs.ex.D} & \nv{rb.rc.epoch-const.D} & \nv{rb.rc.epoch-const.O} & -- & \nv{rb.rc.epoch-const.ex.D} \\
One epoch, standard schedule & \nv{rb.cs.standard} & \nv{rb.cs.once-standard} & -- & \nv{rb.cs.standard.ex} & \nv{rb.rc.epoch.D} & \nv{rb.rc.epoch.O} & \nv{rb.rc.epoch.P} & \nv{rb.rc.epoch.ex.D} \\
Instruction stress test & \nv{m.it.D140} & \nv{m.it.O140} & \nv{mm.P.it} & \nv{mex.O.it} & \nv{rb.rc.it-stress.D} & \nv{rb.rc.it-stress.O} & \nv{rb.rc.it-stress.P} & \nv{rb.rc.it-stress.ex.D} \\
Answer-only stress test & \nv{m.b.D140} & \nv{m.b.O140} & \nv{mm.P.b} & \nv{mex.O.b} & \nv{rb.rc.b-stress.D} & \nv{rb.rc.b-stress.O} & \nv{rb.rc.b-stress.P} & \nv{rb.rc.b-stress.ex.D}$^{\dagger}$ \\
\addlinespace
\multicolumn{5}{@{}l}{\emph{Full fine-tuning of the merged weights}} & & & & \\
One epoch, standard schedule & -- & -- & -- & -- & \nv{rb.rc.epoch-full.D} & \nv{rb.rc.epoch-full.O} & -- & \nv{rb.rc.epoch-full.ex.D} \\
Instruction stress test$^{\ast}$ & -- & -- & -- & -- & \nv{rb.rc.it-stress-full.D} & \nv{rb.rc.it-stress-full.O} & \nv{rb.rc.it-stress-full.P} & \nv{rb.rc.it-stress-full.ex.D}$^{\dagger}$ \\
\bottomrule
\end{tabular}
\end{table}

Every later stage so far continued the adapter that the reasoning training had
produced, at a constant learning rate. A downstream user often does something else:
they receive the weights with the adapter already merged in, and fine-tune them with
a new adapter or with all weights, usually with warm-up, cosine decay and gradient
clipping (the \emph{standard schedule}). If the break needs the drilled adapter itself, or the missing warm-up, such
a user is safe. To find out, we merged each of the three 9B adapters into the base
weights, as a released checkpoint would be, and checked that the merged weights
reproduce Tinker's per-token log-probabilities to within \nv{rb.rc.check.D-u140} nats
and the handoff accuracy to within about a point (Appendix~\ref{app:released}); we call the base weights
with the drilled adapter merged in the \emph{drilled weights}. Each set of weights
then went through the same later stages with a newly initialized rank-\nv{d.rank}
LoRA adapter, or with full fine-tuning. These runs use different training and
sampling software from the earlier ones, and the new adapter covers slightly
different modules; two controls below test whether that matters.

The answer depends on the adapter, not on the schedule (Table~\ref{tab:released}).
Continuing its own adapter under the standard schedule, the drilled model fell to
\nv{rb.cs.standard}\%, just as at a constant rate, while the once-trained model lost
nothing: an extra loss of \nv{rb.cs.standard.ex} points for the drilled model
(interval \nv{rb.cs.standard.ex.ci}). Warm-up and clipping did not protect the
continued drilled adapter. A new adapter does. Started from the drilled weights with a
new adapter, the model came through the one-epoch stage unharmed under either
schedule: its extra loss was \nv{rb.rc.epoch.ex.D} points (interval \nv{rb.rc.epoch.ex.D.ci}). We had registered a prediction of at least 5, and it
failed. Full fine-tuning for the
same epoch gave an extra loss of only \nv{rb.rc.epoch-full.ex.D} points (interval
\nv{rb.rc.epoch-full.ex.D.ci}).

Could the different software, and not the new adapter, be what protects? Two controls
on the drilled model say no. Continuing its own adapter in the new software, it still
lost \nv{rb.rc.epoch-cont-peft.loss.D} points (interval
\nv{rb.rc.epoch-cont-peft.loss.D.ci}), against \nv{rb.cs.constant.loss} on Tinker,
where nothing was frozen (here the adapter's linear-attention query, key and value
deltas had to be). With a new adapter that also covers the model's output head, as
the continued one does, it stayed at \nv{rb.rc.epoch-const-head.D}\%, above its
start. In the one-epoch stage, then, it is the new adapter
that protects the drilled model.

A new adapter does not remove the fragility, though: a harsher stage still finds it.
In the instruction stress test with a new adapter, the once-trained and
fresh-solution models lost nothing, and the drilled model's extra loss was
\nv{rb.rc.it-stress.ex.D} points (interval \nv{rb.rc.it-stress.ex.D.ci}). This is one run of one drilled set, so we
take it as an indication.

The two harshest stages tell us less: answer-only training with a new adapter, and
twenty updates of full fine-tuning (one run, Appendix~\ref{app:released}). In them the
drilled model lost \nv{rb.rc.b-stress.ex.D} and \nv{rb.rc.it-stress-full.ex.D} points
more than the once-trained one (intervals \nv{rb.rc.b-stress.ex.D.ci} and
\nv{rb.rc.it-stress-full.ex.D.ci}). But a stage isolates repetition only if the two
models without it agree, and here the fresh-solution model lost
\nv{rb.rc.b-stress.lessP} and \nv{rb.rc.it-stress-full.lessP} points less than the
once-trained one. Both stages also cost every model much of its
accuracy, and no untrained base went through them in the new software.

Our reading, from one run, is that the drilled weights may be fragile in themselves,
and that continuing the drilled adapter turns this into a large loss even in the
one-epoch stage.

%% file: sections/signature.tex
\section{Why continuing the drilled adapter breaks it: an account}
\label{sec:signature}

Why does continuing the drilled adapter
turn a fragility of the weights into a large loss in the one-epoch stage?

\paragraph{The account.} A LoRA adapter adds a
product $UV$ to a weight matrix, with $U$ starting at zero. One update changes the
weights by about $U\,\Delta V + \Delta U\, V$: each factor's step is scaled by the
other factor. Adam takes steps of a similar size for every parameter, so the larger
the factors, the further each update moves the weights. Drilling makes them larger:
the drilled adapter's $U$ factors have twice the norm of the once-trained adapter's
(\nv{rb.am.D.normB} against \nv{rb.am.O.normB} with the output head;
Appendix~\ref{app:released}). So the same later update should move the drilled
weights further; before measuring it, we registered at least twice as far after five
updates of the instruction stress test. They moved \nv{rb.am.ratio5} times as far:
\nv{rb.am.D.it.u5}\% of the weights' norm, against \nv{rb.am.O.it.u5}\% for both the once-trained and the
fresh-solution models. Much of the difference is in the output head; without it the
ratio is \nv{rb.am.ratio5.nohead}, below the twofold we had registered for all the
weights (Appendix~\ref{app:released}).

A new adapter starts from $U=0$ and a freshly drawn $V$ in every model, so its updates
should be the same size in every model, and they are. Measured the same way for both
(the mean over matrices, without the head; post hoc), later updates moved the continued
drilled adapter's weights
\nv{rb.am.mean.ratio5} times as far as the once-trained one's (after five stress-test updates
and after sixty of the gentler stage alike), and a new adapter's at
most \nv{rb.rc.move.epoch.ratio} times as far. Why repetition enlarges $U$ we can
only guess: the same texts give the same gradients, which add up over passes instead
of partly cancelling.

This is an account, not a demonstrated cause, and it leaves two things unexplained.
First, the drilled model's
extra loss stays at about 10 to 13 points from twenty updates to a full epoch while
its weights keep moving, so the amplification does not set the size of the loss.
Second, a new adapter moves both models alike, yet the drilled one loses more
in the stress test: the fragility itself does not need the amplification.

\begin{figure}[t]
\centering
\figfile{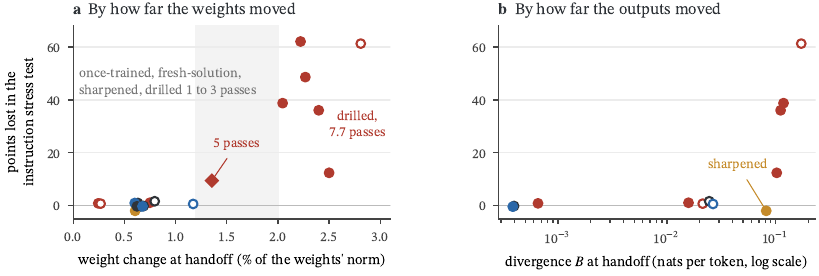}{0.84\linewidth}
\caption{Weight change and output divergence both separate the drilled models, as
the number of passes does; only the sharpened model (gold) is far out on $B$ and
robust. Each dot is one 9B model, placed by its own loss in the instruction stress
test (continuing its adapter) and (a)~by how far its adapter moved the base weights
at handoff, or (b)~by how far its outputs moved, $B$ (log scale; models without a $B$
are omitted). Red: drilled (filled: own solutions, hollow: gpt-oss-120b's traces),
from 1.1 to 7.7 passes; black: once-trained; blue: fresh-solution. The grey band in
(a) is the gap between the robust and the fragile models.}
\label{fig:weights}
\end{figure}

\paragraph{What else differs at handoff: the outputs.} Accuracy at handoff misses the drilled model. A first candidate to catch it is how far
training moved the model's outputs: how much less likely the base model's own forced
solutions become under the trained model, in nats per token. We write this $B$
(Appendix~\ref{app:topp} gives the formula). $B$ is zero if training has left the base
model's reasoning untouched; it estimates the forward divergence
$\mathrm{KL}(p_{\mathrm{base}}\,\|\,p_\theta)$, which \citet{shenfeld2026razor} relate
to forgetting. Under the drilled 9B model the base model's solutions lose
\nv{mb.D140.3} nats per token, about \nv{mb.D140.pdrop}\% of each token's
probability, and almost nothing under the once-trained and fresh-solution models
(Figure~\ref{fig:weights}b; every replication in Table~\ref{tab:full}). Most of it
is \emph{over-sharpening}: \nv{mb.kept}\% of the drilled model's $B$ comes from
positions where it keeps the base model's likeliest token but is even more sure of
it. Over-sharpening could explain the break, so we trained a \emph{sharpened} model
exactly like the once-trained model, but on the base model's first correct sample at
temperature \nv{d.SH.temp}, which concentrates on its likeliest choices
(Appendix~\ref{app:sharpen}, Figure~\ref{fig:bflag}). It moved its outputs three-quarters as far as the
drilled model did, yet after both stress tests its extra loss was under two points
either way, with intervals that include zero. By the criterion we fixed in advance,
seeing the same texts again matters beyond the sharpening it causes: $B$ flags the
drilled models without explaining their fragility.

\paragraph{The weights.} The second measurement is how far the adapter moved the
weights it modifies: the Frobenius norm of its change, relative to theirs. It
separates the models, but only as the number of passes already does. Across all
\nv{rb.wc.judged.n} trained 9B models with a stored adapter and a known
stress-test loss, every model drilled for about eight passes moved its
weights by \nv{rb.wc.drilled.min}\% to \nv{rb.wc.drilled.max}\% and lost
\nv{rb.wc.drilled.lossmin} to \nv{rb.wc.drilled.lossmax} points, and every other
model (once-trained, fresh-solution, sharpened, or drilled for one or three passes)
moved them by at most \nv{rb.wc.other.max}\% and lost at most
\nv{rb.wc.other.lossmax} (Figure~\ref{fig:weights}a). Among the drilled models the
change does not predict the size of the loss. The measure adds one thing: the sharpened
model sits near the drilled models on $B$ but with the once-trained ones on the
weights, at \nv{rb.wc.O-sharp-u140}\%. The comparison is post hoc, except for one
prediction: before training a model drilled for five passes, we registered that its
weights would move by about \nv{rb.p5.pred}\%, inside the gap. They moved
\nv{rb.wc.D-u91}\%, and the model lost \nv{rb.p5.loss} points (interval
\nv{rb.p5.loss.ci}), between the two groups (Appendix~\ref{app:released}).

%% file: sections/suppressed.tex
\section{What the later stage takes away}
\label{sec:lost}

\begin{figure}[t]
\begin{minipage}[c]{0.40\linewidth}
\figfile{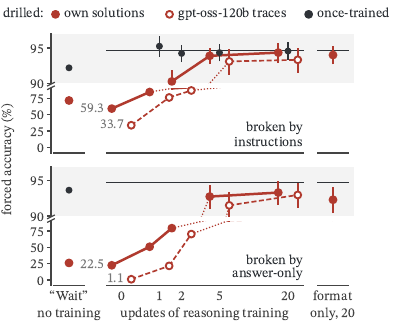}{\linewidth}
\end{minipage}\hfill
\begin{minipage}[c]{0.57\linewidth}
\caption{A few updates of reasoning training restore the broken drilled models; the
once-trained model, trained the same way, barely moves. Forced accuracy after $k$
updates of reasoning training (filled: own solutions; hollow: gpt-oss-120b's
traces), against the once-trained model after the same stress test (line) and given
the same training (dots). Top: after the instruction stress test; bottom: after the
answer-only stress test. The grey strip magnifies 90--97\%; bars: paired 95\%
intervals of $R_k$, the share of the gap closed. ``Wait'': nudged to continue, no
training; format only: twenty updates on the format-only set.}
\label{fig:relearn}
\end{minipage}
\end{figure}

A model can lose accuracy in two ways: it can stop choosing to reason, or it can
reason worse. The instruction stress test does the first to every 9B model we
checked, the base included: answering freely, none of them opens its reasoning block
any more. Scored freely, the drilled and once-trained models are \nv{mfree.it.gapDO} points
apart; with the block opened they are \nv{m.it.gapDO} apart. So the lost decision
to reason hides most of the difference (Appendix~\ref{app:decision}). Is the
gap that remains under forcing a loss of ability?

Forcing the block open does not make a model reason to the end: the broken drilled
models often stop after a few lines or loop (Appendix~\ref{app:mathfail}). If
stopping early were the whole problem, pushing the model on should recover its
accuracy. We tried two test-time probes. Appending ``Wait'' whenever a response
ended early or without a boxed answer, as in the budget forcing of
\citet{muennighoff2025s1}, won back \nv{rf.share.it}\% of the drilled model's extra
loss after the instruction stress test and \nv{rf.share.b}\% after the answer-only
stress test (Table~\ref{tab:reforce}). Forbidding it to end its turn before closing
the reasoning block won back \nv{rb.em.it.pct}\% of its loss after the instruction
stress test and nothing (\nv{rb.em.b.pct}\%) after the answer-only stress test, where it boxed answer
after answer until the token limit (scored after the fact by its first boxed answer,
a looser rule, it won back about half; Appendix~\ref{app:masking}). So, scored as registered, neither probe
wins back half of the loss, which our registered criterion calls lost ability. The next experiment overturns that reading.

If the ability were lost, it would have to be learned again. We continued training
each broken drilled model at the reasoning rate, on the base model's own correct
solutions to problems no model had seen. We measured $R_k$: after $k$ updates, the
fraction of the gap between the broken drilled model and the once-trained model
(after the same stress test) that has closed. $R_5$ of at least
\nv{reg.RL.suppressed} would mean suppressed rather than lost (registered). After the
instruction stress test, five updates closed nearly the whole gap for the drilled
model, $R_5=\nv{rl.D.it.R5}$. The same model after the answer-only stress test, and
the model drilled on gpt-oss-120b's traces after either stress test, recovered about
as far. The once-trained model, trained the same way, barely moved, so the training
restores rather than adds (Figure~\ref{fig:relearn}).

Training on mathematics could still be teaching mathematics. So we also trained the
own-solution drilled model, after each stress test, on a format-only set: the base
model's forced responses to prompts from the broad chat mix, which show the format
of reasoning with almost no mathematics (Appendix~\ref{app:relearn}). Twenty updates
closed the gap just as well: $R_{20}$ of \nv{rl.D.it.habR} after the instruction
stress test and \nv{rl.D.b.habR} after the answer-only stress test. By the bar we had
registered, a share above one half, we drop the claim that the drilled model lost more
than the habit of reasoning.

So the later stage suppresses the drilled model's reasoning; it does not erase it. A
few updates of training on how to reason bring it back, and test-time probes do
not. Fine-tuning in controlled settings likewise wraps capabilities rather than
removing them \citep{jain2024mechanistically}.

%% file: sections/remedies.tex
\section{Preventing the break, and where the claim stops}
\label{sec:remedies}

\paragraph{Replay.} Mixing some earlier data into later training is a common defence against forgetting
\citep{rolnick2019experience,scialom2022continual}. We replaced \nv{d.RP.replay}\% of
each batch with the model's own training texts, which for the drilled model are its
repeated solutions. In the
gentler stage the break disappeared: the drilled model's extra loss was
\nv{rp.ex.D} points, with an interval that includes zero, against \nv{r2.ex} without
replay, and the same held with gpt-oss-120b's traces. In the stress tests replay
helped without preventing the break. After the instruction stress test the drilled
model kept \nv{rb.rh.it.with}\% instead of \nv{rb.rh.it.without}\%; after the
answer-only stress test, \nv{rb.rh.b.with}\% instead of \nv{rb.rh.b.without}\%.
Replay thus cut the two losses by \nv{rb.rh.it.cut}\% and \nv{rb.rh.b.cut}\%. We had
predicted that replay would at least halve the loss in both stress tests, so the
prediction failed, narrowly, in the answer-only stress test
(Appendix~\ref{app:replay}). Because replay helps, our claim concerns later training
without it.

\paragraph{Fresh solutions.} The control that identified the cause is also a
remedy. A new correct solution at every visit prevented the break in every stage
that continued the 9B adapter (Section~\ref{sec:math}). It needs up to \nv{d.P.kept} correct solutions per problem
instead of one, but no more problems. On the 9B model, where there was little
accuracy to gain, it cost none. Elsewhere it did not always help: on the synthetic
skill below it cost some learning, and on the instruct model below it lost
\nv{rb.is.forced.it-stress.ex.P} points more than the once-trained one in the
instruction stress test (Appendix~\ref{app:released}).

\paragraph{A skill the base model lacks.} Everything so far reshapes a skill the base
model already has, the setting s1 and LIMO argue for
\citep{muennighoff2025s1,ye2025limo}. Does repetition do the same to a new one? We
built a task: multiplying two five-digit numbers in base 7 within \nv{d.SK.cap}
tokens, where the base model solves \nv{sk.base}\% while a solver's worked solutions
all fit (Appendix~\ref{app:skill}). Trained on the solver's solutions as in
mathematics, the drilled model learned much less: at handoff it solved
\nv{sk.D.h}\% of held-out problems, against \nv{sk.O.h}\% for the once-trained model,
with the fresh-solution model in between. In the gentler stage the drilled model then
lost no more points than the once-trained one, though somewhat more as a share of
what it had learned, and the fresh-solution model lost the most (each from its own
handoff, a post hoc reading; Table~\ref{tab:sk}). The instruction stress test wiped out the skill in all three. So on a procedure new to the base model,
repetition's main cost was in learning, and what was learned was fragile whatever the
recipe. This is one synthetic task.

\paragraph{An instruct model as the student.} We wanted a setting where reasoning
training adds accuracy, closer to s1 and LIMO: Qwen2.5-7B-Instruct on gpt-oss-120b's
traces (Appendix~\ref{app:released}). Instead the traces lowered its accuracy, and three of our
four registered predictions for it failed. But the one-epoch stage took even the untrained student from \nv{rb.is.student.forced}\% to \nv{rb.is.student.epoch}\% in forced
accuracy, and about as far in free accuracy, so this setting cannot test repetition.
Whether a gain at handoff survives later training remains open.

%% file: sections/related.tex
\section{Related work}
\label{sec:related}

\paragraph{Repetition and durability.} \citet{kopiczko2026repetition} measure the
advantage of repetition right after training; we ask whether it lasts. Facts seen in
identical text are forgotten faster than paraphrased ones \citep{chang2024factual},
edits trained with paraphrases survive later fine-tuning better
\citep{wen2025retention}, and repeated pretraining data harm a model out of
proportion to their share \citep{hernandez2022repeated,chudnovsky2026internal}. More
epochs of reasoning fine-tuning can win after training and lose after a later RL
stage on the same skill \citep{kang2026quagmires}. Our later stages do not train the
skill, and the fresh-solution model plays the role of the paraphrases. Because the
drilled model ends a later stage below the untrained base, that stage suppresses
reasoning the base already had, not just a gain.

\paragraph{What makes a skill last.} Forgetting of earlier capabilities tracks the
forward KL from the base model on the new task \citep{shenfeld2026razor} and shrinks
when training data are resampled from the current model each epoch
\citep{chen2026retaining} or are its own rewrites of the targets
\citep{yang2024sdft}. These studies follow old skills while a new one is learned; we
follow the new one through the next stage. How a skill is acquired shapes its
retention in ways invisible right after training
\citep{feng2026early,springer2025overtrained}; our sharpened model shows that a large
forward divergence from the base model need not make a skill fragile.

\paragraph{Recoverable losses and sharpening.} Capabilities that seem lost after
fine-tuning are often still there
\citep{kotha2024implicit,jain2024mechanistically,zheng2025spurious,qi2025shallow},
and knowledge that unlearning appears to remove returns after brief fine-tuning on
loosely related data \citep{hu2025jogging}, as our drilled models' mathematics
returns after format-only training. Training on responses without reasoning traces
can suppress explicit reasoning while, in several settings, accuracy given valid
reasoning stays high \citep{twist2026collapse}. RL mostly reweights reasoning the
base model already produces \citep{yue2025rl,venhoff2026base}; longer fine-tuning
raises perplexity on reference solutions while accuracy holds \citep{ruan2025overmemorization} and can lower
pass@$k$ \citep{dang2025weight,nguyen2026coverage}. Our results may depend on base models that
already verify and backtrack \citep{gandhi2025cognitive}.

%% file: sections/discussion.tex
\section{Discussion and limitations}
\label{sec:discussion}

For people who train reasoning models, the lesson is to test a model after its
next stage of training, with its reasoning forced as well as free: on mathematics,
accuracy at handoff never told the drilled model apart. On a skill the base model
has, a new solution at every visit is better than a repeated one. An adapter that saw
repeated texts should not be trained further: merge it and start a new one, or mix a
little of the reasoning data back in. The first protected it in the one-epoch stage
but not in the instruction stress test. Why the drilled weights stay fragile under a new adapter is open
\citep[compare][]{springer2025overtrained}.

\paragraph{Limitations.}
\label{sec:limits}
\emph{Training method.} The reasoning training uses LoRA, while s1 and LIMO fully
fine-tune instruct models, so the continued-adapter case that breaks in the one-epoch
stage is not theirs; reasoning training with full fine-tuning is untested, and LoRA
learns less and forgets less than full fine-tuning \citep{biderman2024lora}.
\emph{Ceiling.} The base model is near its ceiling on our test problems, so reasoning
training adds no forced accuracy (on AIME it lowered it, to \nv{aime.h.D140}\% and
\nv{aime.h.O140}\% against the base's \nv{aime.h.base}\%); it changes the model's
default from answering directly to reasoning first. Our attempt at a setting where
reasoning training adds accuracy (Section~\ref{sec:remedies}) did not add any, and the synthetic skill, where the picture changes, is a single task
defined by a token budget. \emph{Breadth.} Our tests are mathematics with integer
answers and one arithmetic task, with no code or science; Nemotron, already
post-trained, does not reproduce s1's order of stages; the 9B model moved only
slightly toward gpt-oss-120b's traces (Appendix~\ref{app:scale}); and our one-epoch
stage is one dataset at one learning rate. At 35B and on Nemotron the stress tests
also cost the base and the once-trained model accuracy (Table~\ref{tab:full}).
\emph{Evidence.} The weight-change measure is post hoc and shows a separation, not a
cause. Across drilled sets of the 9B design the direction held, but the size of the break under the
instruction stress test varied almost fivefold.

%% file: sections/statements.tex
\subsection*{Ethics statement}
The study uses public datasets under their licences: MATH-500 (MIT),
NuminaMath-TIR through the T\"ulu~3 SFT mixture (Apache 2.0; the mixture is
ODC-BY), AIME 2025 and 2026 through MathArena (CC BY-NC-SA 4.0), No Robots
(CC BY-NC 4.0), and, for the broad chat mix, other sources of the T\"ulu~3 SFT
mixture, each under its own licence. Third-party data are released as
identifiers only, with code that rebuilds them from the public sources; the
arithmetic tasks are synthetic. The study involves no human subjects.

\subsection*{Reproducibility statement}
Every experiment was registered before its data, except where marked: the records
give the design, the predictions and their thresholds, and Appendix~\ref{app:ledger}
lists every prediction with its outcome, those added in revision included. Analyses not registered in advance are labelled post
hoc in the text or listed as such in Appendix~\ref{app:ledger}. The repository at
\url{https://github.com/ely2ba/reasoning-durability} holds these records, the code, the
frozen outputs and the number registry. Every number in the paper is written by code into a registry that
names its source: the frozen outputs for results, and the records for design
values, registered thresholds and the few results that only a record states. An
independent audit recomputed the main results from the raw run files. The arms and their
training are in Section~\ref{sec:setup} and Appendix~\ref{app:arms}, the math data
pipeline in Appendix~\ref{app:mathdata}, the other math models and the Nemotron chat
template in Appendix~\ref{app:main}, the released-checkpoint and instruct-student
experiments in Appendix~\ref{app:released}, the arithmetic task and its scoring in
Appendices~\ref{app:tces} and~\ref{app:strict}, and the prompt formats and
sampling settings in the repository. The runs the paper reports used about
\$\nv{comp.total} of Tinker compute at list prices (the teacher's traces at assumed
rates); \texttt{records/compute.md} in the repository has the breakdown. The
experiments added in revision used about \$\nv{rb.comp.tinker} of Tinker compute at
list prices and about \$\nv{rb.comp.modal} of rented GPU time; the designs, budgets
and spend are in \texttt{records/revision-2.md}.

%% file: sections/app-a-training.tex
\section{Training and evaluation in detail}
\label{app:training}
\label{app:arms}

This appendix says how every model in the paper was trained and how it was
scored, so that any result in the paper can be read against it. It supports the
design of Section~\ref{sec:setup}: on mathematics, every model gets the same
number of updates on texts from the same source, and only the texts differ.
Table~\ref{tab:arms} also gives each model's short code, which the released
records and the row labels of some tables use.

\subsection{The models and their codes}

The text calls each model by what was done to it, and Table~\ref{tab:arms} gives
the code that the records use for it. A suffix such as -u140 or @20 gives the
number of updates, so D-u140 is the drilled model after \nv{d.drill.updates}
updates. The math models train on the base model's own correct forced solutions,
drawn at temperature 1 and top-p 1 and kept if correct and finished before the
token cap (Appendix~\ref{app:mathdata}). On the arithmetic task the clone instead learns from \nv{d.S.samples} samples of the
teacher, \nv{d.S.perprompt} per prompt, kept whether right or wrong. Two models
from an earlier replay study \citep{sheikh2026duraseed}, trained on an RL run's
own traces and on solver derivations, are described in Appendix~\ref{app:tces}.

\begin{table}[htbp]
\centering
\caption{Every model trained for the paper, with its code and its name in the
text. Passes: how many times the average training text was seen. All models use
learning rate \nv{d.acq.lr} (the reasoning rate), Adam without gradient clipping,
and a token loss normalized by the set's mean length, in batches of
\nv{d.stage.batch}; the anchored and diluted models add a second stream of
\nv{d.anchor.extra} examples to each batch. The 35B and Nemotron runs repeat the
base, drilled and once-trained models, each on its own model's solutions. On the
arithmetic task, M0, the starting model, is the base model with a format adapter;
R-P and R-S, the trace-replay and solver models of an earlier study, are described
in Appendix~\ref{app:tces}.}
\label{tab:arms}
\footnotesize
\setlength{\tabcolsep}{2.5pt}
\begin{tabularx}{\linewidth}{@{}l l >{\raggedright\arraybackslash}p{0.95in} >{\raggedright\arraybackslash}X r r@{}}
\toprule
Task & Code & Name in the text & Trained on & Updates & Passes \\
\midrule
Math & D & drilled & \nv{d.drill.n} of the base model's correct solutions & \nv{d.upd.low}--\nv{d.drill.updates} & \nv{d.pass.low}--\nv{d.pass.high} \\
& O & once-trained & \nv{d.O.n} of the base model's correct solutions & \nv{d.upd.mid}, \nv{d.drill.updates} & \nv{d.O.pass.mid}, \nv{d.O.pass.high} \\
& P & fresh-solution & D's \nv{d.drill.n} problems, a fresh correct solution at each visit & \nv{d.drill.updates} & \nv{d.P.passes} \\
& D2 & second drilled set & \nv{d.drill.n} other solutions from O's set & \nv{d.drill.updates} & \nv{d.pass.high} \\
& D-r128 & rank-\nv{d.rank.alt} drilled & D's solutions, rank-\nv{d.rank.alt} adapters & \nv{d.drill.updates} & \nv{d.pass.high} \\
& D-T & drilled on traces & gpt-oss-120b traces for \nv{mt.n.D} problems & \nv{d.upd.low}, \nv{d.drill.updates} & \nv{mt.pass.D20}, \nv{mt.pass.D} \\
& O-T & once-trained on traces & gpt-oss-120b traces for \nv{mt.n.O} problems & \nv{d.drill.updates} & \nv{d.O.pass.high} \\
& P-T & fresh-trace & D-T's problems, a fresh trace at each visit & \nv{d.drill.updates} & \nv{d.P.passes} \\
& O-sharp & sharpened & O's problems, the base model's solutions at temperature \nv{d.SH.temp} & \nv{d.drill.updates} & \nv{d.O.pass.high} \\
& D, O, P & the same three, two further runs & as D, O and P, with a new subset, orders and adapter seed & \nv{d.drill.updates} & \nv{d.pass.high}, \nv{d.O.pass.high}, \nv{d.P.passes} \\
\addlinespace
Base-7 skill & D-S & drilled & \nv{d.drill.n} problems, one worked solution each & \nv{d.drill.updates} & \nv{d.pass.high} \\
& O-S & once-trained & \nv{d.O.n} problems, one worked solution each & \nv{d.drill.updates} & \nv{d.O.pass.high} \\
& P-S & fresh-solution & D-S's problems, a different solution at each visit & \nv{d.drill.updates} & \nv{d.P.passes} \\
\addlinespace
Arithmetic & T & teacher & on-policy RL & \nv{d.T.updates} & -- \\
& S & clone & \nv{d.S.samples} samples of T, unfiltered & \nv{d.S.updates} & \nv{d.S.passes} \\
& U & unfiltered drilled & \nv{d.drill.n} samples of T, unfiltered & \nv{d.upd.low}--\nv{d.drill.updates} & \nv{d.pass.low}--\nv{d.pass.high} \\
& F & correct-only drilled & \nv{d.drill.n} correct samples of T & \nv{d.upd.low}--\nv{d.drill.updates} & \nv{d.pass.low}--\nv{d.pass.high} \\
& FA1, FA025 & anchored & F's samples plus \nv{d.anchor.extra} samples of M0 per batch & \nv{d.drill.updates} & \nv{d.pass.high} \\
& FN & diluted & F's samples plus \nv{d.anchor.extra} No Robots examples per batch & \nv{d.drill.updates} & \nv{d.pass.high} \\
& U2 & second unfiltered drilled & \nv{d.drill.n} other samples of T & \nv{d.drill.updates} & \nv{d.pass.high} \\
\bottomrule
\end{tabularx}
\end{table}

\subsection{The math problems and the scorer}
\label{app:mathdata}

The training problems come from T\"ulu~3's decontaminated NuminaMath-TIR subset
\citep{lambert2024tulu3,numina2024tir}, which has \nv{m.src.rows} rows and
\nv{d.math.distinct} distinct problems. A problem's reference answer is the last
boxed answer of its reference solution. Table~\ref{tab:screens} lists the screens
that cut this source down to the training pool. Decontamination against the test
sets used exact matching, 13-gram overlap with and without punctuation, and a
paraphrase search reviewed by hand. Its \nv{m.contam} hits include an AIME 2026
problem word for word and two MATH-500 paraphrases. After their removal, a second
pass of exact and 13-gram matching found none. In a \nv{m.ref.rows}-row sample,
the reference answers agree with the reference solutions' own code output in
\nv{m.ref.agree}\% of cases.

A problem is also skipped if the base model solves it within \nv{d.math.easy}
tokens after an empty reasoning block. Even with an empty block the model usually
writes out its working (in calibration, \nv{m.calib.direct}\% of such short samples were still
writing at the cap), so this screen removes only problems solved almost at once.
Each remaining problem then receives one forced sample at temperature 1 and top-p 1 with a cap of \nv{d.budget.long} tokens,
kept if it is correct and ends before the cap. The drilled model uses the first
\nv{d.drill.n} kept examples, which form a random subset of the once-trained
model's \nv{d.O.n}.

The scorer takes the last boxed answer and parses it as an integer. Before
parsing, it handles nested and escaped braces, strips units, degrees, currency,
percent signs and thousands separators, and treats an unclosed box as no answer.
It passes \nv{m.scorer.units} unit examples, recovers all \nv{d.math.evalall}
MATH-500 reference answers, and read the final box correctly in all
\nv{m.scorer.hand} base solutions that we checked by hand. Evaluation leaves out
MATH-500 levels 1--2, because the base model solved them in full in calibration
(Table~\ref{tab:calib}), and one AIME 2026 problem that is in the training source.

\begin{table}[htbp]
\centering
\caption{Screens of the math pool, in order of removal.}
\label{tab:screens}
\small
\begin{tabular}{@{}p{2.2in} r@{}}
\toprule
Removed & Problems \\
\midrule
Answer not a single integer & \nv{m.scr.nonint} \\
Solution boxes several different answers & \nv{m.scr.multi} \\
Duplicate copies disagree & \nv{m.scr.dup} \\
Figures & \nv{m.scr.fig} \\
Prompt over \nv{m.scr.prompt} tokens & \nv{m.scr.long} \\
Contamination hits & \nv{m.contam} \\
Guessable or not single-answer & \nv{m.scr.guess} \\
\midrule
Kept & \nv{d.math.pool} \\
\bottomrule
\end{tabular}
\end{table}

\begin{table}[htbp]
\centering
\caption{Calibration of the base model, one draw per problem and mode. Direct: an
empty reasoning block with a \nv{d.math.easy}-token cap.}
\label{tab:calib}
\small
\setlength{\tabcolsep}{3pt}
\begin{tabular}{@{}l r r r r@{}}
\toprule
Source & Forced & Free & Direct & Median tokens \\
\midrule
MATH-500 levels 1--2 & \nv{cal.l12.forced} & \nv{cal.l12.free} & \nv{cal.l12.direct} & \nv{cal.l12.med} \\
MATH-500 level 3 & \nv{cal.l3.forced} & \nv{cal.l3.free} & \nv{cal.l3.direct} & \nv{cal.l3.med} \\
MATH-500 level 4 & \nv{cal.l4.forced} & \nv{cal.l4.free} & \nv{cal.l4.direct} & \nv{cal.l4.med} \\
MATH-500 level 5 & \nv{cal.l5.forced} & \nv{cal.l5.free} & \nv{cal.l5.direct} & \nv{cal.l5.med} \\
OlympiadBench (integer) & \nv{cal.oly.forced} & \nv{cal.oly.free} & \nv{cal.oly.direct} & \nv{cal.oly.med} \\
NuminaMath-TIR (integer) & \nv{cal.num.forced} & \nv{cal.num.free} & \nv{cal.num.direct} & \nv{cal.num.med} \\
\bottomrule
\end{tabular}
\end{table}

\subsection{How the other math models were built}
\label{app:main}

\paragraph{Fresh solutions.} The fresh-solution model visits the drilled model's
\nv{d.drill.n} problems in the drilled model's cycling order for
\nv{d.drill.updates} updates, but at each visit it trains on a different correct
forced sample of the base model (top-p 1, cap \nv{d.budget.long}). The drilled
model's own sample of a problem comes first. Sampling for a problem stops at
\nv{d.P.kept} kept samples or \nv{d.P.draws} draws. Of the \nv{d.drill.n}
problems, \nv{mm.P.distinct} yielded \nv{d.P.kept} distinct correct samples. The
other two yielded fewer, so they repeat \nv{mm.P.repeat} of the \nv{mm.slots}
training slots.

\paragraph{The second drilled set and rank \nv{d.rank.alt}.} The second drilled
set takes \nv{d.drill.n} of the once-trained model's solutions that the drilled
model does not use, drawn by hash, and cycles them for
\nv{d.drill.updates} updates. It is compared with the once-trained model of the
first run. The rank-\nv{d.rank.alt} drilled model uses the drilled model's data
and order with a wider adapter. The two ranks fit the drilled texts about equally
well: the negative log-likelihood of the drilled texts falls from the base model's
\nv{mm.R.nll.base} to \nv{mm.R.nll.r32} at rank \nv{d.rank} and to
\nv{mm.R.nll.r128} at rank \nv{d.rank.alt}, so the two drops are within
\nv{mm.R.fitdiff}\% of each other (registered margin: \nv{d.R.fit}\%).

\paragraph{Teacher traces.} The traces were written by gpt-oss-120b at high
reasoning effort, temperature 1 and at most \nv{d.budget.long} tokens. A trace was
kept if it was a clean reply whose last boxed integer matched the answer. A
problem entered the sets of the drilled and once-trained models on traces only if
the teacher solved it within its first four draws. The drilled model on traces
uses \nv{mt.n.D} problems and the once-trained model on traces \nv{mt.n.O}. The
fresh-trace model gets about \nv{mt.traces.P} kept traces for each of the drilled
model's problems.

\paragraph{Nemotron.} Nemotron~3 Nano is prompted in its own chat template, and
on \nv{mnem.tmpl.n} prompts our renderer produces exactly the template's tokens.
Forced reasoning prefills Nemotron's own opening of the reasoning block, a direct
answer prefills an empty reasoning block, and every turn ends on its end-of-turn
token. The later stages use the same examples as the Qwen runs, rendered as
answer-only turns. A newly initialized LoRA adapter reproduces the model exactly
over \nv{mnem.pre.pos} positions.

\paragraph{The remaining models.} The sharpened model is described with its
results in Appendix~\ref{app:sharpen}, the two further runs in
Appendix~\ref{app:runs}, the models of the synthetic skill in
Appendix~\ref{app:skill}, and those of the arithmetic task in
Appendix~\ref{app:tces}.

\subsection{The later stages}
\label{app:stages}

Section~\ref{sec:setup} defines the main later stages: the one-epoch stage, one pass over the No Robots instructions at the reasoning rate and our realistic case, and the two stress tests, twenty updates at
\nv{d.stage.lr} on the instructions (the instruction stress test) or on puzzles
with bare answers (the answer-only stress test). Each later stage starts with a
fresh optimizer; the base model starts it from a new adapter and each trained
model continues its own.

\paragraph{The one-epoch stage.} The one-epoch stage is a single pass over
\nv{rs.rows} No Robots training rows: \nv{rs.updates} updates of
\nv{d.stage.batch} at \nv{d.acq.lr}, evaluated after the last. The rows are all but one of those that pass the selection rules of the instruction stress test, in a fixed hash order.

\paragraph{Longer stages.} Each longer stage runs \nv{d.R2.updates} updates of
\nv{d.stage.batch} and sees each example once. On the instructions it uses
\nv{r2.examples} distinct No Robots examples. The instruction stress test's \nv{d.norobots} examples come first and in the same order, so at \nv{d.stage.lr} its first
\nv{d.stage.updates} updates replay the instruction stress test. At
\nv{d.stage.lrlow}, the reasoning rate, \nv{d.R2.updates} updates give the same
total step, learning rate times updates, as \nv{d.stage.updates} updates at
\nv{d.stage.lr}, a first-order match under Adam. \nv{d.R2.updates} updates on the
instructions at the reasoning rate are what the paper calls the gentler stage.

\paragraph{The broad chat mix.} The broad chat mix has \nv{d.R5.rows} distinct
single-turn rows of the T\"ulu~3 SFT mixture \citep{lambert2024tulu3}, rendered
like the instructions. They come from its non-math sources, leaving out No Robots (which supplies the instructions) and the hard-coded identity prompts. By source: Aya \nv{r5.mix.aya}, FLAN \nv{r5.mix.flan}, Evol-CodeAlpaca \nv{r5.mix.codealpaca}, WildJailbreak
\nv{r5.mix.wildjailbreak}, WildGuardMix \nv{r5.mix.wildguard}, WildChat
\nv{r5.mix.wildchat}, PersonaHub code \nv{r5.mix.phcode} and instruction following
\nv{r5.mix.phif}, CoCoNot \nv{r5.mix.coconot}, SciRIFF \nv{r5.mix.sciriff}, OASST
\nv{r5.mix.oasst} and TableGPT \nv{r5.mix.tablegpt}. The sources are non-math,
but the mix is not free of mathematics. A broad pattern for mathematics (a boxed
answer, a fraction, inline mathematics, ``solve'' or ``calculate'') matches
\nv{r5.mathlike.n} rows (\nv{r5.mathlike}\%), mostly code tasks that calculate
something and grade-school word problems. Worked math word problems make up about
\nv{r5.wordprob}\% of the mix.

\subsection{How the models are scored}
\label{app:eval}

\paragraph{Forced and free answers.} In free generation the model completes the
prompt as it chooses. In forced generation we prefill the response with \think,
which opens a reasoning block, so the model always reasons before it answers. The
math evaluations draw four forced and two free samples for each of the
\nv{d.math.eval} test problems, scored as in Appendix~\ref{app:mathdata}.

\paragraph{Loss, extra loss and intervals.} A model's loss in a later stage is
its drop in forced accuracy from its own handoff, so each model is measured against itself. Its extra loss over a comparison model is how much more it lost; a
positive extra loss means that the first model lost more. Intervals are 95\%
paired bootstrap intervals over test problems; they do not cover training runs.

\paragraph{Token budgets.} A sample counts as correct within a budget of $N$
tokens only if it finished, correct, inside $N$ tokens. So the first $N$ tokens
of a long sample are distributed exactly as a sample capped at $N$. Math is
sampled to \nv{d.budget.long} tokens (\nv{d.budget.aime} for AIME). The arithmetic task was
first sampled within \nv{d.budget.short} tokens and later to \nv{d.budget.long}
(Appendix~\ref{app:budgets}).

\paragraph{Mathematics within smaller budgets} (Table~\ref{tab:budget-math}).
The drilled model's loss is not a matter of running out of tokens. If it were,
the drilled model would solve many more problems as the budget grows, and many of
its responses would end at the cap. After the answer-only stress test, it solves
\nv{mbud.4096.b.D140}\% within \nv{d.budget.short} tokens and
\nv{mbud.16384.b.D140}\% within \nv{d.budget.long}, and only \nv{mcap.b.D140}\%
of its responses reach the cap. The once-trained model, for comparison, stays
within about one point of its handoff accuracy at every budget.

\begin{table}[htbp]
\centering
\caption{Math forced accuracy (\%) of the 9B models within each token budget,
MATH-500 levels 3--5, with the shares of responses at the cap and without a boxed
answer, and the mean response length. Rows: handoff, or after the instruction or
answer-only stress test. O-u140 and D-u140: the once-trained
and drilled models after \nv{d.drill.updates} updates.}
\label{tab:budget-math}
\small
\setlength{\tabcolsep}{5pt}
\begin{tabular}{@{}l r r r r r r r@{}}
\toprule
Model and stage & \nv{d.budget.xs} & \nv{d.budget.short} & \nv{d.budget.mid} & \nv{d.budget.long} & Capped & Unboxed & Mean tokens \\
\midrule
Base, handoff & \nv{mbud.2048.h.base} & \nv{mbud.4096.h.base} & \nv{mbud.8192.h.base} & \nv{mbud.16384.h.base} & \nv{mcap.h.base} & \nv{munbox.h.base} & \nv{mtok.h.base} \\
Base, instruction & \nv{mbud.2048.it.base} & \nv{mbud.4096.it.base} & \nv{mbud.8192.it.base} & \nv{mbud.16384.it.base} & \nv{mcap.it.base} & \nv{munbox.it.base} & \nv{mtok.it.base} \\
Base, answer-only & \nv{mbud.2048.b.base} & \nv{mbud.4096.b.base} & \nv{mbud.8192.b.base} & \nv{mbud.16384.b.base} & \nv{mcap.b.base} & \nv{munbox.b.base} & \nv{mtok.b.base} \\
O-u140, handoff & \nv{mbud.2048.h.O140} & \nv{mbud.4096.h.O140} & \nv{mbud.8192.h.O140} & \nv{mbud.16384.h.O140} & \nv{mcap.h.O140} & \nv{munbox.h.O140} & \nv{mtok.h.O140} \\
O-u140, instruction & \nv{mbud.2048.it.O140} & \nv{mbud.4096.it.O140} & \nv{mbud.8192.it.O140} & \nv{mbud.16384.it.O140} & \nv{mcap.it.O140} & \nv{munbox.it.O140} & \nv{mtok.it.O140} \\
O-u140, answer-only & \nv{mbud.2048.b.O140} & \nv{mbud.4096.b.O140} & \nv{mbud.8192.b.O140} & \nv{mbud.16384.b.O140} & \nv{mcap.b.O140} & \nv{munbox.b.O140} & \nv{mtok.b.O140} \\
D-u140, handoff & \nv{mbud.2048.h.D140} & \nv{mbud.4096.h.D140} & \nv{mbud.8192.h.D140} & \nv{mbud.16384.h.D140} & \nv{mcap.h.D140} & \nv{munbox.h.D140} & \nv{mtok.h.D140} \\
D-u140, instruction & \nv{mbud.2048.it.D140} & \nv{mbud.4096.it.D140} & \nv{mbud.8192.it.D140} & \nv{mbud.16384.it.D140} & \nv{mcap.it.D140} & \nv{munbox.it.D140} & \nv{mtok.it.D140} \\
D-u140, answer-only & \nv{mbud.2048.b.D140} & \nv{mbud.4096.b.D140} & \nv{mbud.8192.b.D140} & \nv{mbud.16384.b.D140} & \nv{mcap.b.D140} & \nv{munbox.b.D140} & \nv{mtok.b.D140} \\
\bottomrule
\end{tabular}
\end{table}

%% file: sections/app-b-results.tex
\section{Every model and every run}
\label{app:results}
\label{app:scale}
\label{app:revision}

This appendix gives the full results behind Section~\ref{sec:math}: every model at
handoff and after each later stage, every extra loss with its interval, and the
repetitions of the 9B design on other runs, models and data. In every test in
Figure~\ref{fig:replication}b, the drilled model's extra loss over its comparison
model in the stress tests has an interval above zero. Its size, though, varies
widely, which is why the main text claims the direction of the effect and not its
size. The row labels use the codes of Table~\ref{tab:arms}.

\subsection{All models at handoff and after the stress tests}

Table~\ref{tab:full} extends Table~\ref{tab:main} to every model, and adds free
accuracy and the divergence $B$ at handoff: how much less likely the base model's
own solutions become under the trained model ($B$ is introduced in
Section~\ref{sec:signature}, defined in Appendix~\ref{app:topp} and discussed in
Appendix~\ref{app:divergence}). Table~\ref{tab:excess} gives every extra loss in
the paper with its interval.

\begin{table}[htbp]
\centering
\caption{Every model at handoff and after twenty updates of each stress test.
Forced accuracy (\%) within \nv{d.budget.long} tokens; free accuracy at handoff.
$B$ is given as a percentage of the starting model's own negative log-likelihood
per token (9B math: \nv{mb.nll} nats; 35B: \nv{m35.nll}; Nemotron:
\nv{mnem.nll}; arithmetic, scored to \nv{d.budget.short} tokens: \nv{n1.nll}).
Math: MATH-500 levels 3--5, four forced draws. Teacher traces: the same design on
gpt-oss-120b's traces, where $B$ is descriptive only. Arithmetic: \nv{d.items}
targeted problems, two draws. Its $B$ is scored on samples drawn at top-p
\nv{d.topp}, which gives every checkpoint a shared offset, so small negative values
occur (Appendix~\ref{app:topp}). Beyond Table~\ref{tab:main}, the table adds the
second drilled set (D2), the rank-\nv{d.rank.alt} drilled model (D-r128), the 35B
and Nemotron runs, Nemotron's starting model, the arithmetic task with its second
unfiltered drilled model (U2), and the exploratory anchored and diluted models
FA025, FA1 and FN (Appendix~\ref{app:anchor}). The last column is the share of
forced responses (\%) that reach the \nv{d.budget.long}-token cap after the
answer-only stress test. Codes: D, O and P are the drilled, once-trained and
fresh-solution models; -T marks gpt-oss-120b's traces; -u20, -u60 and -u140 give
the updates. P-u140 saw each problem \nv{d.pass.high} times, with a fresh solution
each time. On arithmetic, M0 is the starting model, T the RL teacher and S its
clone; U and F are drilled on all of the teacher's samples and on its correct ones
only.}
\label{tab:full}
\footnotesize
\setlength{\tabcolsep}{3pt}
\begin{tabular}{@{}l l r r r r r r r@{}}
\toprule
& & & \multicolumn{3}{c}{At handoff} & \multicolumn{2}{c}{Forced after stress test} & At cap \\
\cmidrule(lr){4-6}\cmidrule(lr){7-8}
Task & Model & Passes & Free & Forced & $B$ (\%) & Instruction & Answer-only & answer-only \\
\midrule
Math, 9B & Base & -- & \nv{mfree.h.base} & \nv{m.h.base} & \nv{m.bfrac.base} & \nv{m.it.base} & \nv{m.b.base} & \nv{mcap.b.base} \\
& D-u20 & \nv{d.pass.low} & \nv{mfree.h.D20} & \nv{m.h.D20} & \nv{mb.frac.D20} & \nv{m.it.D20} & \nv{m.b.D20} & \nv{mcap.b.D20} \\
& D-u60 & \nv{d.pass.mid} & \nv{mfree.h.D60} & \nv{m.h.D60} & \nv{mb.frac.D60} & \nv{m.it.D60} & \nv{m.b.D60} & \nv{mcap.b.D60} \\
& D-u140 & \nv{d.pass.high} & \nv{mfree.h.D140} & \nv{m.h.D140} & \nv{mb.frac.D140} & \nv{m.it.D140} & \nv{m.b.D140} & \nv{mcap.b.D140} \\
& O-u140 & \nv{d.O.pass.high} & \nv{mfree.h.O140} & \nv{m.h.O140} & \nv{mb.frac.O140} & \nv{m.it.O140} & \nv{m.b.O140} & \nv{mcap.b.O140} \\
& P-u140 (fresh) & \nv{d.P.passes} & \nv{mm.P.free} & \nv{mm.P.h} & \nv{mm.P.bfrac} & \nv{mm.P.it} & \nv{mm.P.b} & \nv{mm.P.cap.b} \\
& D2-u140 & \nv{d.pass.high} & \nv{mm.D2.free} & \nv{mm.D2.h} & \nv{mm.D2.bfrac} & \nv{mm.D2.it} & \nv{mm.D2.b} & \nv{mm.D2.cap.b} \\
& D-r128-u140 & \nv{d.pass.high} & \nv{mm.R.free} & \nv{mm.R.h} & \nv{mm.R.bfrac} & \nv{mm.R.it} & \nv{mm.R.b} & \nv{mm.R.cap.b} \\
\addlinespace
Teacher & D-T-u20 & \nv{mt.pass.D20} & \nv{mt.D20.free} & \nv{mt.D20.h} & \nv{mt.bfrac.D20} & \nv{mt.D20.it} & \nv{mt.D20.b} & \nv{mt.cap.b.D20} \\
traces & D-T-u140 & \nv{mt.pass.D} & \nv{mt.D.free} & \nv{mt.D.h} & \nv{mt.bfrac.D} & \nv{mt.D.it} & \nv{mt.D.b} & \nv{mt.cap.b.D} \\
& O-T-u140 & \nv{d.O.pass.high} & \nv{mt.O.free} & \nv{mt.O.h} & \nv{mt.bfrac.O} & \nv{mt.O.it} & \nv{mt.O.b} & \nv{mt.cap.b.O} \\
& P-T-u140 & \nv{d.P.passes} & \nv{mt.P.free} & \nv{mt.P.h} & \nv{mt.bfrac.P} & \nv{mt.P.it} & \nv{mt.P.b} & \nv{mt.cap.b.P} \\
\addlinespace
Math, 35B & Base & -- & \nv{m35.free.h.base} & \nv{m35.h.base} & \nv{m.bfrac.base} & \nv{m35.it.base} & \nv{m35.b.base} & \nv{m35.cap.b.base} \\
& D-u140 & \nv{d.pass.high} & \nv{m35.free.h.D140} & \nv{m35.h.D140} & \nv{m35.bfrac.D140} & \nv{m35.it.D140} & \nv{m35.b.D140} & \nv{m35.cap.b.D140} \\
& O-u140 & \nv{d.O.pass.high} & \nv{m35.free.h.O140} & \nv{m35.h.O140} & \nv{m35.bfrac.O140} & \nv{m35.it.O140} & \nv{m35.b.O140} & \nv{m35.cap.b.O140} \\
\addlinespace
Nemotron & Start & -- & \nv{mnem.free.h.base} & \nv{mnem.h.base} & \nv{m.bfrac.base} & \nv{mnem.it.base} & \nv{mnem.b.base} & \nv{mnem.cap.b.base} \\
& D-u140 & \nv{d.pass.high} & \nv{mnem.free.h.D140} & \nv{mnem.h.D140} & \nv{mnem.bfrac.D140} & \nv{mnem.it.D140} & \nv{mnem.b.D140} & \nv{mnem.cap.b.D140} \\
& O-u140 & \nv{d.O.pass.high} & \nv{mnem.free.h.O140} & \nv{mnem.h.O140} & \nv{mnem.bfrac.O140} & \nv{mnem.it.O140} & \nv{mnem.b.O140} & \nv{mnem.cap.b.O140} \\
\addlinespace
Arithmetic & M0 & -- & -- & \nv{tces16.M0.u0} & \nv{n1.bfrac.M0} & -- & \nv{tces16.M0.u20} & \nv{tces16cap.M0.u20} \\
& T-u30 (RL) & -- & -- & \nv{tces16.T.u0} & \nv{n1.bfrac.T} & -- & \nv{tces16.T.u20} & \nv{tces16cap.T.u20} \\
& S-u230 (clone) & \nv{d.S.passes} & -- & \nv{tces16.S.u0} & \nv{n1.bfrac.S} & -- & \nv{tces16.S.u20} & \nv{tces16cap.S.u20} \\
& U-u140 & \nv{d.pass.high} & -- & \nv{tces16.U140.u0} & \nv{n1.bfrac.U140} & -- & \nv{tces16.U140.u20} & \nv{tces16cap.U140.u20} \\
& F-u140 & \nv{d.pass.high} & -- & \nv{tces16.F140.u0} & \nv{n1.bfrac.F140} & -- & \nv{tces16.F140.u20} & \nv{tces16cap.F140.u20} \\
& U2-u140 & \nv{d.pass.high} & -- & \nv{anc.h16.U2} & \nv{n1.bfrac.U2} & -- & \nv{anc.a16.U2} & \nv{anc.cap.U2} \\
& FA025-u140 & \nv{d.pass.high} & -- & \nv{anc.h16.FA025} & \nv{n1.bfrac.FA025} & -- & \nv{anc.a16.FA025} & \nv{anc.cap.FA025} \\
& FA1-u140 & \nv{d.pass.high} & -- & \nv{anc.h16.FA1} & \nv{n1.bfrac.FA1} & -- & \nv{anc.a16.FA1} & \nv{anc.cap.FA1} \\
& FN-u140 & \nv{d.pass.high} & -- & \nv{anc.h16.FN} & \nv{n1.bfrac.FN} & -- & \nv{anc.a16.FN} & \nv{anc.cap.FN} \\
\bottomrule
\end{tabular}
\end{table}

\begin{table}[htbp]
\centering
\caption{Extra loss of forced accuracy, in points with 95\% paired intervals:
MATH-500 for the 9B model unless noted, and the targeted problems of the
arithmetic task. A positive extra loss means that the first model lost more than
the second between handoff and the end of the later stage. The stress tests run
\nv{d.stage.updates} updates at \nv{d.stage.lr}; the one-epoch stage runs
\nv{rs.updates} updates at \nv{d.stage.lrlow}. Rows on the instructions at \nv{d.stage.lrlow} for \nv{d.R2.updates} updates are the gentler stage (Appendix~\ref{app:stages}).}
\label{tab:excess}
\small
\begin{tabular}{@{}l l r r@{}}
\toprule
Comparison & Later stage & Extra loss & Interval \\
\midrule
Drilled vs.\ once-trained & one-epoch stage & \nv{rs.ex.D} & \nv{rs.ex.D.ci} \\
Fresh-solution vs.\ once-trained & one-epoch stage & \nv{rs.ex.P} & \nv{rs.ex.P.ci} \\
\addlinespace
Drilled vs.\ once-trained & instruction stress test & \nv{mex.O.it} & \nv{mexci.O.it} \\
 & answer-only stress test & \nv{mex.O.b} & \nv{mexci.O.b} \\
Drilled vs.\ fresh-solution & instruction stress test & \nv{mm.DP.it} & \nv{mm.DP.it.ci} \\
 & answer-only stress test & \nv{mm.DP.b} & \nv{mm.DP.b.ci} \\
Fresh-solution vs.\ once-trained & instruction stress test & \nv{mm.PO.it} & \nv{mm.PO.it.ci} \\
 & answer-only stress test & \nv{mm.PO.b} & \nv{mm.PO.b.ci} \\
Sharpened vs.\ once-trained & instruction stress test & \nv{sh.ex.it} & \nv{sh.ex.it.ci} \\
 & answer-only stress test & \nv{sh.ex.b} & \nv{sh.ex.b.ci} \\
Teacher traces: drilled vs.\ once & instruction stress test & \nv{mt.ex.O.it} & \nv{mt.ex.O.it.ci} \\
 & answer-only stress test & \nv{mt.ex.O.b} & \nv{mt.ex.O.b.ci} \\
Teacher traces: fresh vs.\ once & instruction stress test & \nv{mt.PO.it} & \nv{mt.PO.it.ci} \\
 & answer-only stress test & \nv{mt.PO.b} & \nv{mt.PO.b.ci} \\
\addlinespace
Drilled vs.\ once-trained & instructions, \nv{d.stage.lrlow}, \nv{d.R2.updates} updates & \nv{r2.ex} & \nv{r2.ex.ci} \\
 & chat mix, \nv{d.stage.lrlow}, \nv{d.R2.updates} updates & \nv{r5.ex.D} & \nv{r5.ex.D.ci} \\
 & instructions, \nv{d.stage.lr}, \nv{d.R2.updates} updates & \nv{r3.ex60} & \nv{r3.ex60.ci} \\
Teacher traces: drilled vs.\ once & instructions, \nv{d.stage.lrlow}, \nv{d.R2.updates} updates & \nv{r4.ex60} & \nv{r4.ex60.ci} \\
 & chat mix, \nv{d.stage.lrlow}, \nv{d.R2.updates} updates & \nv{r5.ex.DT} & \nv{r5.ex.DT.ci} \\
\addlinespace
Drilled vs.\ once-trained, replay & instructions, \nv{d.stage.lrlow}, \nv{d.R2.updates} updates & \nv{rp.ex.D} & \nv{rp.ex.D.ci} \\
Teacher traces, replay & instructions, \nv{d.stage.lrlow}, \nv{d.R2.updates} updates & \nv{rp.ex.DT} & \nv{rp.ex.DT.ci} \\
\addlinespace
Second run: drilled vs.\ once & instruction stress test & \nv{sd.s2.ex.it} & \nv{sd.s2.ex.it.ci} \\
Third run: drilled vs.\ once & instruction stress test & \nv{sd.s3.ex.it} & \nv{sd.s3.ex.it.ci} \\
Second drilled set vs.\ once & instruction stress test & \nv{mm.D2O.it} & \nv{mm.D2O.it.ci} \\
 & answer-only stress test & \nv{mm.D2O.b} & \nv{mm.D2O.b.ci} \\
Rank \nv{d.rank.alt}: drilled vs.\ once & instruction stress test & \nv{mm.RO.it} & \nv{mm.RO.it.ci} \\
 & answer-only stress test & \nv{mm.RO.b} & \nv{mm.RO.b.ci} \\
35B-A3B: drilled vs.\ once & instruction stress test & \nv{m35.ex.it} & \nv{m35.ex.it.ci} \\
 & answer-only stress test & \nv{m35.ex.b} & \nv{m35.ex.b.ci} \\
Nemotron: drilled vs.\ once & instruction stress test & \nv{mnem.ex.it} & \nv{mnem.ex.it.ci} \\
 & answer-only stress test & \nv{mnem.ex.b} & \nv{mnem.ex.b.ci} \\
Arithmetic: correct-only drilled vs.\ clone & answer-only stress test & \nv{tces16ex.F140} & \nv{tces16ex.F140.ci} \\
Arithmetic: unfiltered drilled vs.\ clone & answer-only stress test & \nv{tces16ex.U140} & \nv{tces16ex.U140.ci} \\
\bottomrule
\end{tabular}
\end{table}

\subsection{The one-epoch stage}
\label{app:oneepoch}

In the one-epoch stage, our realistic case, only the drilled model's loss has an interval above zero. The stage is a single pass over everyday instructions at the reasoning
rate (Appendix~\ref{app:stages}), the kind of stage a released model might go
through next. The base model starts it from a new adapter, and each trained model
continues its own. The base, once-trained and fresh-solution models end within a
point of their handoff accuracy (Table~\ref{tab:rs}). The drilled model's extra
loss over the once-trained model is \nv{rs.ex.D} points, interval \nv{rs.ex.D.ci}, and the fresh-solution model's is \nv{rs.ex.P}, interval \nv{rs.ex.P.ci}. Both registered predictions held: the first extra loss is at least \nv{reg.RS.D} points with its interval above zero, and the second is within \nv{reg.RS.P} points of zero.

\begin{table}[htbp]
\centering
\caption{The one-epoch stage at \nv{d.acq.lr}: forced accuracy (\%) of the 9B
models at handoff and after the stage, and each model's loss, in points with its
95\% interval.}
\label{tab:rs}
\small
\begin{tabular}{@{}l r r r r@{}}
\toprule
Model & Handoff & After & Loss & Interval \\
\midrule
Base, new adapter & \nv{rs.base.h} & \nv{rs.base.a} & \nv{rs.loss.base} & \nv{rs.loss.base.ci} \\
Drilled & \nv{rs.D.h} & \nv{rs.D.a} & \nv{rs.loss.D} & \nv{rs.loss.D.ci} \\
Once-trained & \nv{rs.O.h} & \nv{rs.O.a} & \nv{rs.loss.O} & \nv{rs.loss.O.ci} \\
Fresh-solution & \nv{rs.P.h} & \nv{rs.P.a} & \nv{rs.loss.P} & \nv{rs.loss.P.ci} \\
\bottomrule
\end{tabular}
\end{table}

\subsection{Two further runs of the 9B design}
\label{app:runs}

Our intervals cover test problems, not training runs, so one run cannot show how
much the break depends on which problems were drilled, in what order and from
which initialization. We therefore ran the 9B design twice more. Each run draws a
new subset of \nv{d.drill.n} of the once-trained model's problems, disjoint from
those of the drilled model and of the second drilled set; the two new subsets
share \nv{d.SD.shared} problems. Each run also uses a new order for the drilled
model, a new order for the once-trained model's \nv{d.O.n} texts and a new adapter
initialization. The three models are trained as in the first run, and the
fresh-solution model's solutions for the new problems are sampled under the same
rule. Each model is evaluated at handoff, after the instruction stress test and
after the gentler stage (\nv{d.R2.updates} updates at \nv{d.stage.lrlow} on the
instructions).

For each run and stage we had registered three predictions. The drilled model's
extra loss over the once-trained and over the fresh-solution model would have its
interval above zero; secondarily, it would be at least 5 points; and the fresh-solution model's
extra loss over the once-trained model would stay within 3 points of zero. All
three held in both runs and both stages, with the fresh-solution model within
\nv{sd.PO.absmax} points of the once-trained one (Table~\ref{tab:sd}).  Pooled over four drilled sets (the drilled
model and the second drilled set of the first run, and the drilled models of the
two new runs), the drilled model's extra loss in the instruction stress test
averages \nv{sd.pool.it.mean} points (range
\nv{sd.pool.it.min}--\nv{sd.pool.it.max}, standard deviation \nv{sd.pool.it.sd}). In the gentler stage, over the three
runs, it averages \nv{sd.pool.lr.mean} (\nv{sd.pool.lr.min}--\nv{sd.pool.lr.max}).
The direction of the effect is stable across runs; its size in the instruction
stress test is not.

\begin{table}[htbp]
\centering
\caption{The two further runs of the 9B design: forced accuracy (\%) at handoff
and after each stage, and the drilled model's extra loss over the once-trained
model, in points with its 95\% interval. Gentler stage: \nv{d.R2.updates}
updates at \nv{d.stage.lrlow}.}
\label{tab:sd}
\small
\setlength{\tabcolsep}{4pt}
\begin{tabular}{@{}l l r r r@{}}
\toprule
Run & Model & Handoff & Instruction stress test & Gentler stage \\
\midrule
Second & Drilled & \nv{sd.s2.D.h} & \nv{sd.s2.D.it} & \nv{sd.s2.D.lr} \\
& Once-trained & \nv{sd.s2.O.h} & \nv{sd.s2.O.it} & \nv{sd.s2.O.lr} \\
& Fresh-solution & \nv{sd.s2.P.h} & \nv{sd.s2.P.it} & \nv{sd.s2.P.lr} \\
& Drilled's extra loss & & \nv{sd.s2.ex.it} \nv{sd.s2.ex.it.ci} & \nv{sd.s2.ex.lr} \nv{sd.s2.ex.lr.ci} \\
\addlinespace
Third & Drilled & \nv{sd.s3.D.h} & \nv{sd.s3.D.it} & \nv{sd.s3.D.lr} \\
& Once-trained & \nv{sd.s3.O.h} & \nv{sd.s3.O.it} & \nv{sd.s3.O.lr} \\
& Fresh-solution & \nv{sd.s3.P.h} & \nv{sd.s3.P.it} & \nv{sd.s3.P.lr} \\
& Drilled's extra loss & & \nv{sd.s3.ex.it} \nv{sd.s3.ex.it.ci} & \nv{sd.s3.ex.lr} \nv{sd.s3.ex.lr.ci} \\
\bottomrule
\end{tabular}
\end{table}

\subsection{A larger model and another family}

To check that the break does not belong to one model, we repeated the base, drilled and once-trained models on Qwen3.5-35B-A3B-Base and on Nemotron~3 Nano, whose
starting model is already post-trained. On both, the drilled model again lost
more than the once-trained model in both stress tests (Table~\ref{tab:scale};
intervals in Table~\ref{tab:excess}). At 35B, though, the instruction stress test
costs every model more than on the 9B, not only the drilled one. The once-trained
model goes from \nv{m35.h.O140}\% to \nv{m35.it.O140}\%, and the untrained base,
which starts from a new adapter, from \nv{m35.h.base}\% to \nv{m35.it.base}\%.

\begin{table}[htbp]
\centering
\caption{A second scale and a second family: Qwen3.5-35B-A3B-Base and
Nemotron~3 Nano. Forced accuracy (\%) within \nv{d.budget.long} tokens (AIME:
\nv{d.budget.aime}), four draws, at handoff and after twenty updates of the
instruction stress test or of the answer-only stress test. Nemotron's starting
model is already post-trained. D-u20, D-u60 and D-u140: the drilled model after
\nv{d.pass.low}, \nv{d.pass.mid} and \nv{d.pass.high} passes; O-u140: the
once-trained model. Last row: the extra loss of the drilled model (D-u140) over
the once-trained model, in points.}
\label{tab:scale}
\footnotesize
\setlength{\tabcolsep}{3.5pt}
\begin{tabular}{@{}l rrr rrr@{}}
\toprule
& \multicolumn{3}{c}{35B-A3B} & \multicolumn{3}{c}{Nemotron} \\
\cmidrule(lr){2-4}\cmidrule(lr){5-7}
State & Handoff & Instruction & Answer-only & Handoff & Instruction & Answer-only \\
\midrule
\multicolumn{7}{@{}l}{\emph{MATH-500, levels 3--5}} \\
Base & \nv{m35.h.base} & \nv{m35.it.base} & \nv{m35.b.base} & \nv{mnem.h.base} & \nv{mnem.it.base} & \nv{mnem.b.base} \\
D-u20 & \nv{m35.h.D20} & \nv{m35.it.D20} & \nv{m35.b.D20} & \nv{mnem.h.D20} & \nv{mnem.it.D20} & \nv{mnem.b.D20} \\
D-u60 & \nv{m35.h.D60} & \nv{m35.it.D60} & \nv{m35.b.D60} & \nv{mnem.h.D60} & \nv{mnem.it.D60} & \nv{mnem.b.D60} \\
D-u140 & \nv{m35.h.D140} & \nv{m35.it.D140} & \nv{m35.b.D140} & \nv{mnem.h.D140} & \nv{mnem.it.D140} & \nv{mnem.b.D140} \\
O-u140 & \nv{m35.h.O140} & \nv{m35.it.O140} & \nv{m35.b.O140} & \nv{mnem.h.O140} & \nv{mnem.it.O140} & \nv{mnem.b.O140} \\
\multicolumn{7}{@{}l}{\emph{AIME 2025 and 2026}} \\
Base & \nv{m35.aime.h.base} & \nv{m35.aime.it.base} & -- & \nv{mnem.aime.h.base} & \nv{mnem.aime.it.base} & -- \\
D-u140 & \nv{m35.aime.h.D140} & \nv{m35.aime.it.D140} & -- & \nv{mnem.aime.h.D140} & \nv{mnem.aime.it.D140} & -- \\
O-u140 & \nv{m35.aime.h.O140} & \nv{m35.aime.it.O140} & -- & \nv{mnem.aime.h.O140} & \nv{mnem.aime.it.O140} & -- \\
\midrule
Extra loss & & \nv{m35.ex.it} & \nv{m35.ex.b} & & \nv{mnem.ex.it} & \nv{mnem.ex.b} \\
\bottomrule
\end{tabular}
\end{table}

\subsection{Where the break sets in}
\label{app:passes}

The drilled models were saved after \nv{d.pass.low}, \nv{d.pass.mid} and
\nv{d.pass.high} passes, so the tables show where the break sets in. On the 9B
model it comes between \nv{d.pass.mid} and \nv{d.pass.high} passes. At
\nv{d.pass.mid} passes the drilled model loses \nv{mloss.it.D60} points (interval
\nv{mlossci.it.D60}) in the instruction stress test and at most \nv{mloss.b.D60}
(interval \nv{mlossci.b.D60}) in the answer-only stress test, although its $B$
has already risen to \nv{mb.D60.3}, \nv{mb.frac.D60}\% of the base model's
negative log-likelihood.

On the other two models the break comes earlier. At \nv{d.pass.mid} passes the
35B's drilled model loses \nv{m35.loss.b.D60} points (interval
\nv{m35.loss.b.D60.ci}) in the answer-only stress test, at a $B$ of
\nv{m35.bfrac.D60}\% of its base's. Nemotron's loses \nv{mnem.loss.it.D60}
(interval \nv{mnem.loss.it.D60.ci}) in the instruction stress test, at a $B$ of
only \nv{mnem.bfrac.D60}\%, a stage in which its once-trained model also falls
from \nv{mnem.h.O140}\% to \nv{mnem.it.O140}\%, and its starting model from \nv{mnem.h.base}\% to
\nv{mnem.it.base}\%. On the arithmetic task the checkpoints at \nv{d.pass.mid}
passes, with a $B$ of \nv{n1x.bfrac.U60}--\nv{n1x.bfrac.F60}\% of the starting
model's, lost clearly more than the 9B math model at the one budget where both
tasks were read, \nv{d.budget.short} tokens (Appendix~\ref{app:budgets}). The number of passes at which the break appears thus depends on the model and on the task.

\subsection{The second drilled set, rank \nv{d.rank.alt} and teacher traces}

The second drilled set and the rank-\nv{d.rank.alt} drilled model
(Appendix~\ref{app:main}) both lose more than the once-trained model in both
stress tests, with intervals above zero (Tables~\ref{tab:full}
and~\ref{tab:excess}).

\paragraph{Teacher traces.} The drilled model on gpt-oss-120b's traces breaks in
both stress tests (Section~\ref{sec:math}). That result also tests a rival
reading that we had registered and called consolidation: repeating a stronger
model's traces would make the reasoning more durable, with an extra loss under
\nv{reg.T2.thr} points in each stress test. It failed. The drilled model's extra
loss on traces is \nv{mt.ex.O.it} points in the instruction stress test and
\nv{mt.ex.O.b} in the answer-only stress test. We had also set two conditions for
trusting a null result here. The drilling had to reach \nv{mt.drill.req} of the
fall in training loss seen on the model's own solutions; it reached only
\nv{mt.drill.ratio}. The once-trained model's held-out negative log-likelihood on
the traces had to fall \nv{mt.held.req}\%; it fell only \nv{mt.held.drop}\%. Neither
condition was met, but the result is not null, so this changes no outcome. It does
show that the student moved little toward the teacher.

\subsection{AIME}
\label{app:aime}

AIME 2025 and 2026 are a harder check (\nv{d.math.aime} problems, four draws,
within \nv{d.budget.aime} tokens), and there the instruction stress test again
separates the drilled model from the others. The 9B base goes from
\nv{aime.h.base}\% to \nv{aime.it.base}\%, the once-trained model from
\nv{aime.h.O140}\% to \nv{aime.it.O140}\% and the fresh-solution model from
\nv{mm.P.aime.h}\% to \nv{mm.P.aime.it}\%. The drilled model falls from
\nv{aime.h.D140}\% to \nv{aime.it.D140}\%. On gpt-oss-120b's traces, the
once-trained model goes from \nv{mt.aime.O.h}\% to \nv{mt.aime.O.it}\%, while the
drilled model falls from \nv{mt.aime.D.h}\% to \nv{mt.aime.D.it}\%. With
\nv{d.math.aime} problems the interval is about $\pm$\nv{aime.ci} points, so AIME
is descriptive. The contrast repeats at 35B and on Nemotron, where the untrained
35B base also collapses (Table~\ref{tab:scale}).

\subsection{How the later stage's rate and length change the break}
\label{app:stage}

Section~\ref{sec:dose} reports that, on instruction data, the size of the break
depends mostly on the learning rate of the later stage. This subsection gives the
trajectories behind that statement; the longer stages and the broad chat mix are
defined in Appendix~\ref{app:stages}. At the reasoning rate, the length of the
stage changes the extra loss little: the drilled model's extra loss was
\nv{r2.ex20} points after twenty updates on the instructions and \nv{r2.ex} after
sixty.

\begin{figure}[htbp]
\centering
\figfile{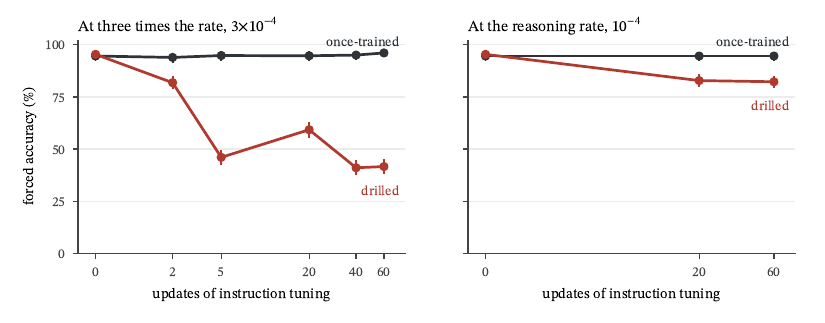}{\linewidth}
\caption{Forced accuracy (\%, 95\% intervals) of the drilled and once-trained 9B
models through \nv{d.R2.updates} updates on the No Robots instructions, at
\nv{d.stage.lr} (the rate of the stress tests) and at the reasoning rate,
\nv{d.stage.lrlow} (the gentler stage), on \nv{r2.examples} distinct examples
whose first \nv{d.norobots} are those of the instruction stress test. At
\nv{d.stage.lr}, the states after 2 and 5 updates come from a replay of the
stress test whose training losses match it within \nv{d.R1.match}\%, and those
after 40 and 60 from a run that replays its first 20 updates exactly, so the
points trace one trajectory. Lines join the measured points.}
\label{fig:trajectory}
\end{figure}

\paragraph{Accuracy} (Figure~\ref{fig:trajectory}). At the rate of the stress
tests, the once-trained model holds
(\nv{m.it.O140}--\nv{r3.f.O.u60}\%), while the drilled model's forced accuracy is
\nv{m.it.D140}\% after \nv{d.stage.updates} updates, \nv{r3.f.D.u40}\% after 40
and \nv{r3.f.D.u60}\% after \nv{d.R2.updates}. At the reasoning rate,
\nv{d.stage.lrlow}, the once-trained model is at \nv{r2.f.O.u20}\% and \nv{r2.f.O.u60}\% after
\nv{d.stage.updates} and \nv{d.R2.updates} updates, and the drilled model at
\nv{r2.f.D.u20}\% and \nv{r2.f.D.u60}\%. With the teacher's traces at
\nv{d.stage.lrlow}, the once-trained model is at \nv{r4.f.OT.u20}\% and
\nv{r4.f.OT.u60}\%, and the drilled model at \nv{r4.f.DT.u20}\% and
\nv{r4.f.DT.u60}\%. After \nv{d.R2.updates} updates on the broad chat mix, the
two once-trained models, on their own solutions and on traces, are at
\nv{r5.f.O}\% and \nv{r5.f.OT}\%, and the two drilled models at \nv{r5.f.D}\%
and \nv{r5.f.DT}\%. Two registered predictions held: both broad-mix extra losses have intervals above zero, and the two once-trained models lose at most \nv{r5.req} points.
The registered secondary prediction that each extra loss reaches \nv{r5.req.sec} points held for the drilled model on its own
solutions (\nv{r5.ex.D}) and failed for the one on traces (\nv{r5.ex.DT}).

\paragraph{Training loss.} On the broad mix, the drilled models start with a
higher training loss than the once-trained ones and end level with them. Over the
first five updates the mean loss per example is \nv{r5.loss.D.u1-5} and \nv{r5.loss.DT.u1-5} for the drilled models on their own
solutions and on traces, against \nv{r5.loss.O.u1-5} and \nv{r5.loss.OT.u1-5}
for the matching once-trained models. Over the last five updates it is
\nv{r5.loss.D.u56-60} and \nv{r5.loss.DT.u56-60} for the drilled models, against
\nv{r5.loss.O.u56-60} and \nv{r5.loss.OT.u56-60} for the once-trained ones. On
the instructions at \nv{d.stage.lrlow}, the drilled model on traces goes from
\nv{r4.loss.DT.u1-5} to \nv{r4.loss.DT.u56-60}, and the once-trained model on
traces from \nv{r4.loss.OT.u1-5} to \nv{r4.loss.OT.u56-60}.

%% file: sections/app-c-released.tex
\section{Released checkpoints, weight change and an instruct student}
\label{app:released}

This appendix reports the experiments added in revision, each registered
before its data except where marked. Does the break need
the continued adapter, or the constant learning rate of the paper's later stages
(Section~\ref{sec:released})? In the one-epoch stage it needs the continued adapter.
What does drilling do to the weights (Section~\ref{sec:signature})? It enlarges the adapter's $U$ factors, so each later update of a continued drilled adapter moves the weights further. Does the design carry over to an instruct model trained on a stronger
model's traces (Section~\ref{sec:remedies})? That setting could not tell us. The Tinker SDK the paper pinned (0.25.0) no longer opens sessions, so our Tinker runs
use 0.39.1 with tinker-cookbook 0.7.0; we checked that the prompts these build match
the paper's stored prompts token for token (200 of 200).

Table~\ref{tab:scope} shows where the evidence for the break lies after these
experiments. Almost all of it concerns a later stage that continues the reasoning
adapter.

\begin{table}[htbp]
\centering
\caption{What the evidence covers. Entries are the drilled model's extra loss over
the once-trained model in forced accuracy (points), on the 9B design unless named.
Middle column: the one-epoch stage and the gentler stage, at the reasoning rate (a
tenth of it for full fine-tuning); right column: the two stress tests. The reasoning training is LoRA
throughout.}
\label{tab:scope}
\footnotesize
\begin{tabularx}{\linewidth}{@{}>{\raggedright\arraybackslash}p{0.2\linewidth} >{\raggedright\arraybackslash}X >{\raggedright\arraybackslash}X@{}}
\toprule
Later stage starts from & One-epoch and gentler stages & Stress tests \\
\midrule
The reasoning adapter, continued (Sections~\ref{sec:math}, \ref{sec:signature}) &
breaks: \nv{rs.ex.D} in the one-epoch stage, \nv{rb.cs.standard.ex} under the standard
schedule; \nv{r2.ex} in the gentler stage, and \nv{sd.s2.ex.lr} and \nv{sd.s3.ex.lr} in
runs 2 and 3 &
breaks: \nv{mex.O.it} and \nv{mex.O.b}; also on a second drilled set, at rank
\nv{d.rank.alt}, with a stronger model's traces, in runs 2 and 3, at 35B and on
Nemotron; five passes: \nv{rb.p5.ex} (loss \nv{rb.p5.loss}, a partial break; readings registered, no prediction) \\
\addlinespace
Merged weights, new adapter (Section~\ref{sec:released}) &
no harm: \nv{rb.rc.epoch.ex.D} \nv{rb.rc.epoch.ex.D.ci}, one drilled set &
instruction stress test: \nv{rb.rc.it-stress.ex.D} \nv{rb.rc.it-stress.ex.D.ci}, one
run; answer-only stress test: every model loses most of its accuracy, not informative
about repetition \\
\addlinespace
Merged weights, full fine-tuning (Section~\ref{sec:released}) &
little harm: \nv{rb.rc.epoch-full.ex.D} \nv{rb.rc.epoch-full.ex.D.ci} &
instruction stress test, one amended run: every model loses much;
\nv{rb.rc.it-stress-full.ex.D}, but the fresh-solution model lost
\nv{rb.rc.it-stress-full.lessP} less than the once-trained one, so not informative about
repetition \\
\addlinespace
Reasoning trained by full fine-tuning, or with a gain at handoff &
untested (our instruct-student attempt gave no gain) & untested \\
\bottomrule
\end{tabularx}
\end{table}

\subsection{Merging the adapters reproduces the handoff accuracy}

The merged checkpoints behave like the Tinker models they came from. Each 9B handoff
state was saved on Tinker as sampler weights and downloaded as an adapter for Hugging Face PEFT, the
library we trained with off Tinker (rank \nv{d.rank}, $\alpha=\nv{d.rank}$, on the
attention and linear-attention projections, the MLP and the output head). In this
appendix an adapter changes a weight matrix by the product $UV$ of two factors, where
$U$ starts at zero. Tinker adapts the query, key and value projections of a
linear-attention layer separately, but the Hugging Face checkpoint stores them as the
rows of one matrix, in that order, so we added each adapter's $UV$ to the
corresponding rows. The merged weights were saved in bfloat16 in the base
checkpoint's own layout.

Two checks compare the merged models with Tinker. First, the merged models scored the
first 64 of the base model's forced solutions token by token. For the untrained base
the mean absolute difference from Tinker's log-probabilities of the same texts is
\nv{rb.rc.check.M0} nats per token, which measures the two software setups alone; for
the drilled, once-trained and fresh-solution models it is \nv{rb.rc.check.D-u140},
\nv{rb.rc.check.O-u140} and \nv{rb.rc.check.P-u140}. Second, sampled with vLLM 0.34.0
from the paper's exact prompt tokens (temperature 1, top-p 0.95, four draws, 16,384
tokens), the merged models solve \nv{rb.rc.h.D}\%, \nv{rb.rc.h.O}\% and
\nv{rb.rc.h.P}\% of the test problems at handoff, against \nv{rb.rc.tinker.D}\%,
\nv{rb.rc.tinker.O}\% and \nv{rb.rc.tinker.P}\% on Tinker, and the base
\nv{rb.rc.h.base}\% against \nv{rb.rc.tinker.base}\%. We call PEFT and this sampler
the new software. Every loss and extra loss in it is measured from the handoff
accuracy in that software.

\subsection{Later stages from merged weights: a new adapter protects in the one-epoch stage}

The new adapter is a rank-\nv{d.rank} LoRA ($\alpha=\nv{d.rank}$) on every linear
layer of the text model, trained in the new software on one H100 in bfloat16 mixed
precision. Unlike Tinker's adapter, it covers two small projections of the
linear-attention layers and does not cover the output head. Batches, data and order
are the paper's: \nv{d.stage.batch} examples per update, each example's loss averaged
over its target tokens. The standard schedule is AdamW ($\beta=(0.9, 0.999)$,
$\epsilon=10^{-8}$, no weight decay) with a linear warm-up over the first 10\% of
updates, cosine decay to zero and gradient clipping at 1.0; the constant schedule is
the paper's (Adam with $\beta=(0.9,0.95)$, $\epsilon=10^{-12}$, no clipping). After
the stage the new adapter is merged and the model sampled as above. Full fine-tuning
trains every weight of the text model in float32 with 8-bit AdamW on one H200, at a
peak rate of \nv{rb.full.lr} on the standard schedule, the usual tenfold reduction
from the LoRA rate.

In the one-epoch stage a new adapter protects the drilled model, against our
registered prediction. On the standard schedule at a peak rate of \nv{d.acq.lr}, we
had predicted a drilled extra loss of at least 5 points with its interval above
zero; it was \nv{rb.rc.epoch.ex.D} \nv{rb.rc.epoch.ex.D.ci} (registered; failed). The
registered secondary stage, at a peak rate of $2\times10^{-4}$, was not run.

That stage changed both the adapter and the schedule. To separate them we added
stages after that result, as exploratory runs whose readings were fixed in words
before their data (Table~\ref{tab:released}). The schedule is not what protects. On
Tinker, continuing its own adapter through the one-epoch stage on the standard
schedule (clipping at 32 on Tinker's summed loss, which is 1.0 per example), the
once-trained model lost \nv{rb.cs.once-standard.loss}
\nv{rb.cs.once-standard.loss.ci} points (registered: at most 3; held) and the drilled
model \nv{rb.cs.standard.loss} \nv{rb.cs.standard.loss.ci}, against
\nv{rb.cs.constant.loss} \nv{rb.cs.constant.loss.ci} at the constant rate. The drilled model's extra loss under the standard schedule is therefore \nv{rb.cs.standard.ex}
\nv{rb.cs.standard.ex.ci}. A new adapter protects under either schedule: at the
constant rate the drilled model lost \nv{rb.rc.epoch-const.loss.D}
\nv{rb.rc.epoch-const.loss.D.ci}. Full fine-tuning for one epoch, a registered secondary stage with no threshold, also did little harm: the once-trained model lost \nv{rb.rc.epoch-full.loss.O}
\nv{rb.rc.epoch-full.loss.O.ci} and the drilled model \nv{rb.rc.epoch-full.loss.D}
\nv{rb.rc.epoch-full.loss.D.ci}, an extra loss of \nv{rb.rc.epoch-full.ex.D}
\nv{rb.rc.epoch-full.ex.D.ci}.

With a new adapter, the instruction stress test still costs the drilled model more than the others; the answer-only stress test costs every model most of its accuracy (both exploratory, one run). In the instruction stress test with
a new adapter, the once-trained model lost \nv{rb.rc.it-stress.loss.O}
\nv{rb.rc.it-stress.loss.O.ci}, the fresh-solution model \nv{rb.rc.it-stress.loss.P}
\nv{rb.rc.it-stress.loss.P.ci} and the drilled model \nv{rb.rc.it-stress.loss.D}
\nv{rb.rc.it-stress.loss.D.ci}. In the answer-only stress test with a new adapter
every model lost most of its accuracy (drilled \nv{rb.rc.b-stress.loss.D},
once-trained \nv{rb.rc.b-stress.loss.O}, fresh-solution \nv{rb.rc.b-stress.loss.P}
points). No untrained base went through that stage in the new software. On Tinker
the base, starting a new adapter, kept \nv{m.b.base}\% after the answer-only stress
test, so the stage may be harsher here than there. We do not know why the two
software setups differ in harshness, so we compare results across them only where
both ran the same stage.

The instruction stress test with full fine-tuning needed one amendment. We registered
it with the paper's constant-rate optimizer at $3\times10^{-5}$ (a tenth of the LoRA
stress rate) and the prediction that the drilled model's extra loss would have its
interval above zero. With that optimizer (Adam $\beta=(0.9,0.95)$,
$\epsilon=10^{-12}$, no clipping, 8-bit states) both models diverged within four
updates, their training loss rising above 9 nats per token from about 1.7. We
stopped the run before any evaluation and, recording the change first, ran it once
more on the standard schedule (two updates of warm-up to $3\times10^{-5}$, cosine
decay, clipping at 1.0). The stage then costs every model accuracy, the drilled model
most: the once-trained model fell to \nv{rb.rc.it-stress-full.O}\% (a loss of
\nv{rb.rc.it-stress-full.loss.O} \nv{rb.rc.it-stress-full.loss.O.ci}), the
fresh-solution model to \nv{rb.rc.it-stress-full.P}\% and the drilled model to
\nv{rb.rc.it-stress-full.D}\% (\nv{rb.rc.it-stress-full.loss.D}
\nv{rb.rc.it-stress-full.loss.D.ci}). The drilled model's extra loss is
\nv{rb.rc.it-stress-full.ex.D} \nv{rb.rc.it-stress-full.ex.D.ci} (registered; held,
in this amended run). We had also registered, before its run, that the fresh-solution
model would lose within 5 points of the once-trained one. It lost
\nv{rb.rc.it-stress-full.loss.P} \nv{rb.rc.it-stress-full.loss.P.ci},
\nv{rb.rc.it-stress-full.ex.P} points relative to the once-trained model (registered:
within 5; failed, in the direction of more robustness). With the two models without
repetition this far apart, the stage cannot isolate repetition.

Two controls, registered before their data, ask whether the new software, or leaving out the output head, rather than the new adapter itself protects the drilled model in the one-epoch stage at the constant
rate. If the software protected, continuing the drilled model's own adapter in it
should be harmless. It is not. Because the checkpoint fuses the linear-attention query, key and value matrices,
those deltas were merged and frozen; every other module, the output head included,
continued from Tinker's factors. The drilled model fell
from \nv{rb.rc.h.D}\% to \nv{rb.rc.epoch-cont-peft.D}\%, a loss of
\nv{rb.rc.epoch-cont-peft.loss.D} \nv{rb.rc.epoch-cont-peft.loss.D.ci} (registered:
at least 5; held). If leaving out the output head protected, a new adapter that
covers the head should not. It still does: the drilled model ended at
\nv{rb.rc.epoch-const-head.D}\% (a loss of \nv{rb.rc.epoch-const-head.loss.D}
\nv{rb.rc.epoch-const-head.loss.D.ci}; registered: at most 3; held). The continued
adapter lost less here than on Tinker (\nv{rb.cs.constant.loss}), with part of it
frozen, and neither control has a once-trained arm. Within those limits, neither the
software nor leaving out the output head explains the protection. In the one-epoch
stage it comes from the new adapter.

\subsection{Drilled adapters move the weights furthest}

Adapters drilled for about eight passes moved the base weights furthest. For an adapter that adds $\Delta W$ to a set of weight matrices $W$, the
weight change is $\|\Delta W\|_F/\|W\|_F$ over all of them, in percent. This measure
is post hoc: we first saw it while merging the three adapters above, then computed it
for every 9B handoff state still stored on Tinker. Table~\ref{tab:weights} lists them
with each model's loss in the instruction stress test, continuing its adapter. One
further state, the once-trained model after 60 updates, has no stress-test
measurement and is not shown (\nv{rb.wc.O-u60}\%). The five-pass drilled model was
registered in advance: we predicted its weight change at about \nv{rb.p5.pred}\%,
interpolated over the passes of the paper's dose series, and fixed that a stress-test
loss of at most 3 points, more than 10 points, or anything between, would be read as
no break, a break and a partial break. Its weight change fell between the two
groups, as predicted, and its loss was a partial break.

\begin{table}[htbp]
\centering
\caption{Weight change at handoff ($\|\Delta W\|_F/\|W\|_F$, \%) and points lost in
the instruction stress test, continuing the adapter, for every stored 9B handoff
state. $\|U\|_F$: the norm of the adapter's $U$ factors, output head left out (post
hoc). Texts: own, the base model's own solutions; teacher, gpt-oss-120b's traces.
Runs 2 and 3: the two further runs with new drilled problems, orders and
initializations.}
\label{tab:weights}
\small
\begin{tabular}{@{}l l r r r@{}}
\toprule
Model & Texts & Weight change & $\|U\|_F$ & Points lost \\
\midrule
Drilled, \nv{d.pass.high} passes & own & \nv{rb.wc.D-u140} & \nv{rb.fn.D-u140} & \nv{rb.sl.D-u140} \\
Drilled, \nv{d.pass.high} passes, second set & own & \nv{rb.wc.D2-u140} & \nv{rb.fn.D2-u140} & \nv{rb.sl.D2-u140} \\
Drilled, \nv{d.pass.high} passes, rank \nv{d.rank.alt} & own & \nv{rb.wc.D-r128-u140} & -- & \nv{rb.sl.D-r128-u140} \\
Drilled, \nv{d.pass.high} passes, run 2 & own & \nv{rb.wc.s2-D-u140} & \nv{rb.fn.s2-D-u140} & \nv{rb.sl.s2-D-u140} \\
Drilled, \nv{d.pass.high} passes, run 3 & own & \nv{rb.wc.s3-D-u140} & \nv{rb.fn.s3-D-u140} & \nv{rb.sl.s3-D-u140} \\
Drilled, \nv{mt.pass.D} passes & teacher & \nv{rb.wc.D-T-u140} & \nv{rb.fn.D-T-u140} & \nv{rb.sl.D-T-u140} \\
\addlinespace
Drilled, \nv{rb.p5.passes} passes (registered) & own & \nv{rb.wc.D-u91} & \nv{rb.fn.D-u91} & \nv{rb.p5.loss} \\
\addlinespace
Drilled, \nv{d.pass.mid} passes & own & \nv{rb.wc.D-u60} & \nv{rb.fn.D-u60} & \nv{rb.sl.D-u60} \\
Drilled, \nv{d.pass.low} passes & own & \nv{rb.wc.D-u20} & \nv{rb.fn.D-u20} & \nv{rb.sl.D-u20} \\
Drilled, \nv{d.pass.low} passes & teacher & \nv{rb.wc.D-T-u20} & \nv{rb.fn.D-T-u20} & \nv{rb.sl.D-T-u20} \\
Once-trained & own & \nv{rb.wc.O-u140} & \nv{rb.fn.O-u140} & \nv{rb.sl.O-u140} \\
Once-trained, runs 2 and 3 & own & \nv{rb.wc.s2-O-u140}, \nv{rb.wc.s3-O-u140} & \nv{rb.fn.s2-O-u140}, \nv{rb.fn.s3-O-u140} & \nv{rb.sl.s2-O-u140}, \nv{rb.sl.s3-O-u140} \\
Once-trained & teacher & \nv{rb.wc.O-T-u140} & \nv{rb.fn.O-T-u140} & \nv{rb.sl.O-T-u140} \\
Fresh-solution & own & \nv{rb.wc.P-u140} & \nv{rb.fn.P-u140} & \nv{rb.sl.P-u140} \\
Fresh-solution, runs 2 and 3 & own & \nv{rb.wc.s2-P-u140}, \nv{rb.wc.s3-P-u140} & \nv{rb.fn.s2-P-u140}, \nv{rb.fn.s3-P-u140} & \nv{rb.sl.s2-P-u140}, \nv{rb.sl.s3-P-u140} \\
Fresh-solution & teacher & \nv{rb.wc.P-T-u140} & \nv{rb.fn.P-T-u140} & \nv{rb.sl.P-T-u140} \\
Sharpened & own & \nv{rb.wc.O-sharp-u140} & \nv{rb.fn.O-sharp-u140} & \nv{rb.sl.O-sharp-u140} \\
\bottomrule
\end{tabular}
\end{table}

The size of the $U$ factors separates the models the same way. The table gives their
Frobenius norm with the output head left out (post hoc; the rank-\nv{d.rank.alt}
adapter's is not comparable). Every adapter drilled for about eight passes has a norm
between \nv{rb.fn.drilled.min} and \nv{rb.fn.drilled.max}, every other rank-32
adapter except the five-pass one at most \nv{rb.fn.other.max}, and the five-pass one
\nv{rb.fn.D-u91}, in the gap like its loss. But within the drilled group the norm
varies by a few percent while the loss varies fivefold. The size of the factors, like
the weight change, separates repetition from its absence but does not set the size of
the break.

Table~\ref{tab:measures} lists every measure of the weights used in the paper, with
how it treats the output head and whether it was registered.

\begin{table}[htbp]
\centering
\caption{Measures of the weights: the drilled model's value over the once-trained
model's. Pooled: $\|\Delta W\|_F/\|W\|_F$ over all adapted matrices together; mean:
the mean over matrices of each matrix's ratio. Later updates are counted from
handoff; stress updates are updates of the instruction stress test, gentler updates
those of the gentler stage. A new adapter does not cover the output head.}
\label{tab:measures}
\footnotesize
\setlength{\tabcolsep}{3pt}
\begin{tabular}{@{}p{0.34\linewidth} l l r l@{}}
\toprule
What & Measure & Head & Drilled / once & Status \\
\midrule
Change at handoff & pooled & included & \nv{rb.wc.ratio} & post hoc \\
$U$ factors at handoff & Frobenius norm & included & \nv{rb.am.ratioB} & registered ($\geq$1.5) \\
$U$ factors at handoff & Frobenius norm & left out & \nv{rb.fn.ratio} & post hoc \\
Continued adapter, 1 and 5 stress updates & pooled & included & \nv{rb.am.ratio1}, \nv{rb.am.ratio5} & registered at 5 ($\geq$2) \\
Continued adapter, 60 gentler updates & pooled & included & \nv{rb.am.ratio60} & measured, no prediction \\
Continued adapter, 5 stress updates & pooled & left out & \nv{rb.am.ratio5.nohead} & post hoc \\
Continued adapter, 5 stress updates & pooled & alone & \nv{rb.am.ratio5.head} & post hoc \\
Continued adapter, 1 and 5 stress, 60 gentler updates & mean & left out & \nv{rb.am.mean.ratio1}, \nv{rb.am.mean.ratio5}, \nv{rb.am.mean.ratio60} & post hoc \\
New adapter, end of each stage & mean & not covered & \nv{rb.rc.move.it-stress.ratio}--\nv{rb.rc.move.epoch.ratio} & post hoc \\
\bottomrule
\end{tabular}
\end{table}

\paragraph{Each later update moves the drilled weights further.} This tests the
account of Section~\ref{sec:signature}: if the drilled adapter's factors are larger,
each later update of that adapter moves the weights further. We registered that the
drilled adapter's $U$ factors would have at least 1.5 times the Frobenius norm of the
once-trained adapter's, and that after five updates of the instruction stress test
the drilled model's weights would have moved at least twice as far since handoff as
the once-trained model's, with the fresh-solution model within a factor 1.25 of the
once-trained one. We measured the states of the paper's own stress test after 1, 2
and 5 updates and of the gentler stage after 20 and 60. Both predictions held.

With the output head included, the norms of the $U$ factors are \nv{rb.am.D.normB}
(drilled), \nv{rb.am.O.normB} (once-trained) and \nv{rb.am.P.normB}
(fresh-solution), and those of the $V$ factors \nv{rb.am.D.normA}, \nv{rb.am.O.normA}
and \nv{rb.am.P.normA}. After 1, 2 and 5 stress-test updates the weights had moved
since handoff by \nv{rb.am.D.it.u1}\%, \nv{rb.am.D.it.u2}\% and
\nv{rb.am.D.it.u5}\% (drilled), \nv{rb.am.O.it.u1}\%, \nv{rb.am.O.it.u2}\% and
\nv{rb.am.O.it.u5}\% (once-trained) and \nv{rb.am.P.it.u1}\%, \nv{rb.am.P.it.u2}\% and
\nv{rb.am.P.it.u5}\% (fresh-solution). In the gentler stage, after 20 and 60 updates,
the drilled weights had moved \nv{rb.am.D.gentle.u20}\% and \nv{rb.am.D.gentle.u60}\%,
against \nv{rb.am.O.gentle.u20}\% and \nv{rb.am.O.gentle.u60}\% for the once-trained
ones. The output head carries the largest part of the drilled adapter's change.
Leaving it out, the drilled and once-trained models moved \nv{rb.am.D.it.u5.nohead}\%
and \nv{rb.am.O.it.u5.nohead}\% after five stress-test updates, a ratio of
\nv{rb.am.ratio5.nohead}, below the twofold we had registered for all the weights;
the head alone moved \nv{rb.am.ratio5.head} times as far for the drilled model.

A new adapter, by contrast, moves every model alike. On the measure the merge
records, the mean over matrices without the output head (which a new adapter does not
cover), it changed the weights by \nv{rb.rc.move.epoch.D}\% for the drilled and
\nv{rb.rc.move.epoch.O}\% for the once-trained model in the one-epoch stage, and by
\nv{rb.rc.move.it-stress.D}\% and \nv{rb.rc.move.it-stress.O}\% in the instruction
stress test: a ratio of \nv{rb.rc.move.it-stress.ratio} to
\nv{rb.rc.move.epoch.ratio}. Measured the same way after the fact, the continued
adapters' later updates moved the drilled weights \nv{rb.am.mean.ratio1}, \nv{rb.am.mean.ratio5} and
\nv{rb.am.mean.ratio60} times as far as the once-trained ones, after one and five
stress-test updates and sixty updates of the gentler stage, and the fresh-solution
weights \nv{rb.am.mean.ratioP5} times as far after five.

We also asked, after the fact, in which direction the later updates of a continued
adapter move the weights. Without the output head, the cosine between the later
stage's change and the change the reasoning training had made is between $-0.02$ and
$-0.06$ for every model and stage measured. The updates are nearly orthogonal to what
training wrote, for the drilled model as for the others. So the later stage does not
simply undo the training.

\subsection{An instruct student: the one-epoch stage harms every model}

This part repeats the design on an instruct model trained on a stronger model's
traces, closer to s1 and LIMO \citep{muennighoff2025s1,ye2025limo}. It cannot test repetition: the traces gave no gain at
handoff, and the one-epoch stage removed half or more of every model's accuracy,
trained or not (a reading made after the results).

The student is Qwen2.5-7B-Instruct in its own chat template with its default system
prompt. Forced reasoning prefills \texttt{<think>} and a newline in the assistant
turn, and a turn ends at its end-of-turn token. The training texts are the paper's
teacher arms, unchanged in content (Appendix~\ref{app:main}), rebuilt from the stored
traces as an assistant turn: the trace inside the reasoning block, then the final
answer. The drilled model cycles one trace for each of \nv{rb.is.D.n} problems for
\nv{d.drill.updates} updates; the once-trained model sees \nv{rb.is.O.n} traces once;
the fresh-solution model visits the drilled problems with a new trace each time.
Training follows the paper's recipe (rank-\nv{d.rank} LoRA, constant \nv{d.acq.lr},
Adam $\beta=(0.9,0.95)$), in the new software, and the adapters were then merged.

The traces lowered the student's accuracy. We had registered that the setting
answers the question of a gain at handoff only if the once-trained model's forced
accuracy exceeded the untrained student's free accuracy by at least 5 points
(registered; failed). The student solved \nv{rb.is.student.free}\% free (\nv{rb.is.student.forced}\% forced)
before training and, forced, \nv{rb.is.forced.h.D}\% (drilled),
\nv{rb.is.forced.h.O}\% (once-trained) and \nv{rb.is.forced.h.P}\% (fresh-solution)
after it. With a new adapter, the one-epoch stage left the drilled and once-trained
models at \nv{rb.is.forced.epoch.D}\% and \nv{rb.is.forced.epoch.O}\% (extra loss
\nv{rb.is.forced.epoch.ex.D} \nv{rb.is.forced.epoch.ex.D.ci}), and the instruction
stress test left the three at \nv{rb.is.forced.it-stress.D}\%,
\nv{rb.is.forced.it-stress.O}\% and \nv{rb.is.forced.it-stress.P}\% (drilled extra
loss \nv{rb.is.forced.it-stress.ex.D} \nv{rb.is.forced.it-stress.ex.D.ci}).

The paper's protocol, though, continues the reasoning adapter. So we trained the
drilled and once-trained students again with their adapters saved (same data, order
and settings; at handoff \nv{rb.ic.h.D}\% and \nv{rb.ic.h.O}\%) and registered that,
continuing their own adapters through the instruction stress test, the drilled one
would lose more, with the interval of the extra loss above zero. It held, near the
floor: the drilled student fell to \nv{rb.ic.it-stress.D}\% and the once-trained one
to \nv{rb.ic.it-stress.O}\%, an extra loss of \nv{rb.ic.it-stress.ex.D}
\nv{rb.ic.it-stress.ex.D.ci}. Registered predictions 1 to 3 of this part failed: no
gain at handoff; drilled extra losses of \nv{rb.is.forced.epoch.ex.D}
\nv{rb.is.forced.epoch.ex.D.ci} and \nv{rb.is.forced.it-stress.ex.D}
\nv{rb.is.forced.it-stress.ex.D.ci} with a new adapter, where we had predicted one with its interval above zero; and a fresh-solution student that lost \nv{rb.is.forced.it-stress.ex.P}
\nv{rb.is.forced.it-stress.ex.P.ci} points more than the once-trained one in the
stress test, not within 3.

The untrained student shows why. Put through the one-epoch stage with no prediction,
it fell from \nv{rb.is.student.forced}\% to \nv{rb.is.student.epoch}\% forced (a loss
of \nv{rb.is.student.epoch.loss} \nv{rb.is.student.epoch.loss.ci}), below where the
trained students ended. Our reading, made after these results, is that in this
setting the stage removes half or more of every model's forced accuracy, trained or
not, so the instruct student cannot test repetition. But the reasoning block is not
native to this model's chat format, so the forced opening itself might have been what
failed. We therefore measured free accuracy after the stages, with the readings fixed
beforehand and no prediction. It fell as far. After the one-epoch stage the untrained student's free accuracy went
from \nv{rb.is.student.free}\% to \nv{rb.is.student.epochfree}\% (a loss of
\nv{rb.is.student.epochfree.loss} \nv{rb.is.student.epochfree.loss.ci}), the drilled
student's from \nv{rb.is.free.h.D}\% to \nv{rb.is.free.epoch.D}\% and the
once-trained one's from \nv{rb.is.free.h.O}\% to \nv{rb.is.free.epoch.O}\% (drilled
extra loss \nv{rb.is.free.epoch.ex.D} \nv{rb.is.free.epoch.ex.D.ci}). After the
instruction stress test the drilled, once-trained and fresh-solution students were at
\nv{rb.is.free.it-stress.D}\%, \nv{rb.is.free.it-stress.O}\% and
\nv{rb.is.free.it-stress.P}\% (drilled extra loss \nv{rb.is.free.it-stress.ex.D}
\nv{rb.is.free.it-stress.ex.D.ci}). The stage itself harms this chat model, whether or
not its reasoning is forced, and the drilled student loses no more than the others.

%% file: sections/app-d-failure.tex
\section{How the drilled models fail, and what brings the reasoning back}
\label{app:failure}

This appendix asks how the broken drilled models fail and what brings their
accuracy back, in support of Section~\ref{sec:lost}: the later stage suppresses the
drilled model's reasoning rather than erasing it. With its reasoning block opened,
the broken drilled model stops after a few lines or goes round in loops.
Test-time probes win back less than half of the loss under the registered scoring;
a few updates of training win back nearly all of it, even on data with almost no
mathematics.

\subsection{What the failing responses look like}
\label{app:mathfail}

\paragraph{The 9B model} (Table~\ref{tab:mathfail}). The broken drilled model does
not search without end; it ends its turn early. After the answer-only stress test,
\nv{mf.b.D.short}\% of its forced responses stop within \nv{mf.short.tokens} tokens,
and only \nv{mf.b.D.cap}\% reach the token limit. \nv{mf.b.D.ends}\% end the turn
from inside the reasoning block, and only \nv{mf.closes.n} of the \nv{mf.n}
responses close the block at all. \nv{mf.b.D.unbox}\% leave no boxed answer, and
none uses the stress test's \texttt{<answer>} format.

Missing boxes do not explain the loss. Crediting a correct final integer without a
box adds \nv{mf.credit.it} points to the drilled model's accuracy after the
instruction stress test and \nv{mf.credit.b} after the answer-only stress test.
With the once-trained model scored the same way, the drilled model's extra loss over it is still \nv{mf.lenient.it} and
\nv{mf.lenient.b} points, against \nv{mex.O.it} and \nv{mex.O.b} under the usual
scoring.

\begin{table}[t]
\centering
\caption{Forced math responses of the 9B models after twenty updates of each stress
test (\nv{mf.n} per model). Correct: forced accuracy (\%); crediting integers: the
same, counting a correct final integer without a box. ``Ends its turn'': the response
stops by opening a new user turn.}
\label{tab:mathfail}
\footnotesize
\setlength{\tabcolsep}{3pt}
\begin{tabular}{@{}l r r r r r r r r@{}}
\toprule
& & Crediting & & Median & Under \nv{mf.short.tokens} & Ends & Closes & \\
Model and stress test & Correct & integers & Unboxed & tokens & tokens & its turn & block & At cap \\
\midrule
Drilled, answer-only & \nv{mf.b.D.acc} & \nv{mf.b.D.int} & \nv{mf.b.D.unbox} & \nv{mf.b.D.med} & \nv{mf.b.D.short} & \nv{mf.b.D.ends} & \nv{mf.b.D.closes} & \nv{mf.b.D.cap} \\
Drilled, instruction & \nv{mf.it.D.acc} & \nv{mf.it.D.int} & \nv{mf.it.D.unbox} & \nv{mf.it.D.med} & \nv{mf.it.D.short} & \nv{mf.it.D.ends} & \nv{mf.it.D.closes} & \nv{mf.it.D.cap} \\
Once-trained, answer-only & \nv{mf.b.O.acc} & \nv{mf.b.O.int} & \nv{mf.b.O.unbox} & \nv{mf.b.O.med} & \nv{mf.b.O.short} & \nv{mf.b.O.ends} & \nv{mf.b.O.closes} & \nv{mf.b.O.cap} \\
Base, answer-only & \nv{mf.b.base.acc} & \nv{mf.b.base.int} & \nv{mf.b.base.unbox} & \nv{mf.b.base.med} & \nv{mf.b.base.short} & \nv{mf.b.base.ends} & \nv{mf.b.base.closes} & \nv{mf.b.base.cap} \\
\bottomrule
\end{tabular}
\end{table}

\paragraph{AIME.} On AIME the drilled model fails both ways after the instruction
stress test. \nv{aime.it.D140.cap}\% of its responses reach
the \nv{d.budget.aime}-token cap, \nv{aime.it.D140.loop}\% of those in a loop, and
\nv{aime.it.D140.unboxed}\% leave no boxed answer. The long cap matters on AIME:
within \nv{d.budget.long} tokens the base model solves \nv{aime.base.h16}\% at
handoff, against \nv{aime.h.base}\% within \nv{d.budget.aime}.

\paragraph{The other drilled models.} The other drilled models fail through the
same mix of early stops and loops, in varying proportions. After the answer-only
stress test, the second drilled set and the rank-\nv{d.rank.alt} drilled model end
their forced turns even sooner than the first run's drilled model, with medians of
\nv{mm.D2.med.b} and \nv{mm.R.med.b} tokens. The drilled model on gpt-oss-120b's
traces also ends its turns early: its forced responses have a median of
\nv{mt.med.it.D} tokens after the instruction stress test, with \nv{mt.unbox.it.D}\%
unboxed, and of \nv{mt.med.b.D} after the answer-only stress test. The 35B drilled
model mixes the two failures after the answer-only stress test. Its median response
has \nv{m35.med.b.D140} tokens and \nv{m35.short.b.D140}\% of its responses end
within \nv{d.reforce.tokens} tokens, while \nv{m35.cap.b.D140}\% reach the cap,
\nv{m35.loop.b.D140}\% of those in a loop, and \nv{m35.unbox.b.D140}\% leave no boxed
answer. Nemotron's drilled model gives short, often unboxed answers after the
instruction stress test: \nv{mnem.unbox.it.D140}\% unboxed, a mean of
\nv{mnem.tok.it.D140} tokens and \nv{mnem.cap.it.D140}\% at the cap.

\paragraph{After the longer stages.} After
\nv{d.R2.updates} updates at \nv{d.stage.lrlow}, the reasoning rate, on the broad
chat mix and, for the models trained on traces, also on the instructions (the
gentler stage). The drilled models write shorter responses than the once-trained
ones (median \nv{r45.med.D.min}--\nv{r45.med.D.max} tokens against
\nv{r45.med.O.min}--\nv{r45.med.O.max}). They close the reasoning block in only
\nv{r45.close.D.min}--\nv{r45.close.D.max}\% of responses, against about
\nv{r45.close.O.min}\%, so the answer usually stays inside the block and the turn
ends there. Their answers are unboxed in
\nv{r45.unbox.D.min}--\nv{r45.unbox.D.max}\% of responses, against
\nv{r45.unbox.O.min}--\nv{r45.unbox.O.max}\%. On the broad chat mix most of the gap
is wrong answers, not missing ones; for the traces on the instructions, about half
of it is missing answers. In every state \nv{r45.cap.min}--\nv{r45.cap.max}\% of
responses reach the cap. These counts are descriptive.

\paragraph{The arithmetic task.} There the drilled models fail differently: they
search until the token cap without finding an answer (Appendix~\ref{app:search}).

\subsection{Nudging with ``Wait''}
\label{app:reforce}

If the drilled model only stopped too soon, pushing it on should bring its accuracy
back. It does so only in part: appending ``Wait'' under the budget rule removes \nv{rf.share.it}\% of the drilled model's extra loss after the instruction stress test and \nv{rf.share.b}\% after the answer-only stress test (Table~\ref{tab:reforce}). The share reported for this probe
in Section~\ref{sec:lost} is a share of that extra loss.

The probe reuses the forced samples of the base, drilled and once-trained 9B models
at handoff and after each stress test (\nv{mf.n} per model and stage); no new first
pass is drawn. A sample that ended its turn before the \nv{d.budget.long}-token cap
is continued as follows: its end of turn is removed, ``\textbackslash
n\textbackslash nWait'' is appended, and sampling resumes from the whole sequence
with the evaluation's settings and the remaining budget, at most
\nv{d.reforce.times} times and never past \nv{d.budget.long} tokens in all. The last
boxed integer of the final text is scored. Under the budget rule, the primary one,
a sample is continued while it has fewer than \nv{d.reforce.tokens} tokens or no
boxed answer; under the answer rule, only while it has no boxed answer. Neither rule
looks at correctness, and both apply to every model.

We had registered that if the drilled model's extra loss, with every model scored
under the same rule, stayed at least half of its first-pass value, with its interval
above zero, we would read the competence as lost. It stayed above half after both
stress tests and under both rules (Table~\ref{tab:reforce}). Relearning
(Appendix~\ref{app:relearn}) later overturned
that reading: a few updates restore the competence, so the paper calls it
suppressed.

The budget rule continued \nv{rfb.cont.D.it}\% of the drilled model's responses
after the instruction stress test and \nv{rfb.cont.D.b}\% after the answer-only
stress test. After the answer-only stress test the continued responses ran long
without finding more answers (median \nv{rfb.med.D.b} tokens, \nv{rfb.cap.D.b}\% at
the cap). The answer rule, which continues only unboxed responses, recovers more
there: \nv{rfa.D.b}\%, against \nv{rfb.D.b}\% under the budget rule. The cue has a
cost of its own: under the budget rule it lowers the accuracy of every unbroken model
by \nv{rf.cost.min}--\nv{rf.cost.max} points, except the base model
after the answer-only stress test, which loses \nv{rf.cost.baseb}
(Table~\ref{tab:reforce-all}).

\begin{table}[htbp]
\centering
\caption{Nudging the drilled 9B model with ``Wait''. Top: the drilled model's extra
loss over the once-trained model, in points with 95\% paired intervals, with every
model scored under the same rule; the first pass is the registered result. Bottom:
the drilled model's forced accuracy (\%). Budget rule (primary): a response that ends
before \nv{d.reforce.tokens} tokens or without a boxed answer is continued with
``Wait''. Answer rule: only unboxed responses are continued.}
\label{tab:reforce}
\small
\setlength{\tabcolsep}{4pt}
\begin{tabular}{@{}l r r@{}}
\toprule
& Instruction stress test & Answer-only stress test \\
\midrule
\multicolumn{3}{@{}l}{\emph{Extra loss over the once-trained model}} \\
First pass & \nv{mex.O.it} \nv{mexci.O.it} & \nv{mex.O.b} \nv{mexci.O.b} \\
Budget rule & \nv{rfb.ex.it} \nv{rfb.ex.it.ci} & \nv{rfb.ex.b} \nv{rfb.ex.b.ci} \\
Answer rule & \nv{rfa.ex.it} \nv{rfa.ex.it.ci} & \nv{rfa.ex.b} \nv{rfa.ex.b.ci} \\
\multicolumn{3}{@{}l}{\emph{Drilled model, forced accuracy}} \\
First pass & \nv{m.it.D140} & \nv{m.b.D140} \\
Budget rule & \nv{rfb.D.it} & \nv{rfb.D.b} \\
Answer rule & \nv{rfa.D.it} & \nv{rfa.D.b} \\
\bottomrule
\end{tabular}
\end{table}

\begin{table}[htbp]
\centering
\caption{Forced accuracy (\%) of the 9B models on MATH-500 levels 3--5 when nudged
with ``Wait'', at handoff and after each stress test: first pass, budget rule and
answer rule.}
\label{tab:reforce-all}
\small
\begin{tabular}{@{}l r r r@{}}
\toprule
State & First pass & Budget rule & Answer rule \\
\midrule
Base & \nv{m.h.base} & \nv{rfb.base} & \nv{rfa.base} \\
Base, instruction & \nv{m.it.base} & \nv{rfb.base.it} & \nv{rfa.base.it} \\
Base, answer-only & \nv{m.b.base} & \nv{rfb.base.b} & \nv{rfa.base.b} \\
Drilled & \nv{m.h.D140} & \nv{rfb.D} & \nv{rfa.D} \\
Drilled, instruction & \nv{m.it.D140} & \nv{rfb.D.it} & \nv{rfa.D.it} \\
Drilled, answer-only & \nv{m.b.D140} & \nv{rfb.D.b} & \nv{rfa.D.b} \\
Once-trained & \nv{m.h.O140} & \nv{rfb.O} & \nv{rfa.O} \\
Once-trained, instruction & \nv{m.it.O140} & \nv{rfb.O.it} & \nv{rfa.O.it} \\
Once-trained, answer-only & \nv{m.b.O140} & \nv{rfb.O.b} & \nv{rfa.O.b} \\
\bottomrule
\end{tabular}
\end{table}

\subsection{Forbidding the end of the turn}
\label{app:masking}

If the drilled model's loss were only a habit of stopping early, forbidding it to
stop should give the reasoning back without any training. It does not: under the
registered scoring, the mask wins back less than half of the model's loss after
either stress test (added in revision; registered before its data).
Unlike that of ``Wait'', this share is a share of the model's own loss, the gap
between its accuracy without the mask and its handoff accuracy, not of its extra
loss over the once-trained model.

Of the drilled model's \nv{d.math.eval}~$\times$~4 forced answers after the
answer-only stress test, almost
all end by opening a new user turn (the ``User:'' of the role-colon chat format), and
a few with the end-of-text token. We ran the two stress tests again from the drilled
model, merged the results into the base weights and sampled them with vLLM
(Appendix~\ref{app:released}). The mask gives the tokens ``User'', `` User'', a tab
followed by ``User'', and end-of-text a logit bias of $-100$ until the model has
written the token that closes its reasoning block; after it, sampling is free and
stops as usual. Prompts, draws and token limit are those of every forced evaluation,
and each model is also sampled without the mask on the same software. We registered
that recovering at least 0.8 of the gap in both stress tests would mean the loss is
a stopping habit that a test-time probe removes, and under 0.5 in both that it is
not; we predicted under 0.5.

Without the mask, on this software, the two broken models solve
\nv{rb.em.it.plain}\% and \nv{rb.em.b.plain}\% (the paper's runs gave
\nv{m.it.D140}\% and \nv{m.b.D140}\%), against \nv{rb.rc.h.D}\% at handoff. After
the instruction stress test the mask raised accuracy to \nv{rb.em.it.masked}\%, a
recovered share of \nv{rb.em.it.share} \nv{rb.em.it.share.ci}, and the model closed
its reasoning block in \nv{rb.em.it.closes}\% of answers, against
\nv{rb.em.it.closes.plain}\% without the mask. After the answer-only stress test the
mask lowered accuracy to \nv{rb.em.b.masked}\%, a recovered share of
\nv{rb.em.b.share} \nv{rb.em.b.share.ci}. Unable to end its turn, the model wrote
until the token limit in \nv{rb.em.b.capped}\% of samples and closed its reasoning
block in \nv{rb.em.b.closes}\%, boxing \nv{rb.em.b.boxes} answers in the median
sample. The prediction held (registered: under 0.5): the mask does not restore
the reasoning.

The registered scoring reads the last boxed answer, so it penalises a model that
keeps writing after a right answer. We therefore added two scorings after seeing the
samples. The first judges only the masked samples by their first boxed answer, and
measures the share against the unmasked accuracy under the registered scoring. The
masked model is then right in \nv{rb.em.it.first}\% of samples after the instruction
stress test, a share of \nv{rb.em.it.share.first}, and in \nv{rb.em.b.first}\% after
the answer-only stress test, a share of \nv{rb.em.b.share.first}
\nv{rb.em.b.share.first.ci} (added after seeing the samples). The second scores every
sample by its first boxed answer and gives shares of \nv{rb.em.it.share.firstall}
\nv{rb.em.it.share.firstall.ci} and \nv{rb.em.b.share.firstall}
\nv{rb.em.b.share.firstall.ci} (added after seeing the samples). But this second rule
also lowers the handoff accuracy from \nv{rb.rc.h.D}\% to
\nv{rb.em.it.handoff.first}\%, because models that end their turn box intermediate
answers before the final one. That is why the first scoring does not apply the
first-answer rule to samples that end their turn.

By the registered scoring, then, forbidding the end of the turn wins back less than
half of the loss, and by the scorings added after seeing the samples between about a
third and three-quarters. Five updates of reasoning training win back nearly all of
it (Appendix~\ref{app:relearn}).

\subsection{Later training takes the decision to reason first}
\label{app:decision}

\paragraph{The measure.} A model can lose accuracy after a later stage because it
stops choosing to reason or because it reasons worse. We measure the decision to
reason as the share of free responses that open the reasoning block (free and forced
answers are defined in Appendix~\ref{app:eval}). Nemotron's later stages train on
replies with an empty reasoning block, so for Nemotron opening the block no longer
marks reasoning, and we read its decision from
free accuracy. On the arithmetic task we read it from the share of openings that
skip the search (Appendix~\ref{app:tcesdecision}).

\begin{figure}[htbp]
\centering
\figfile{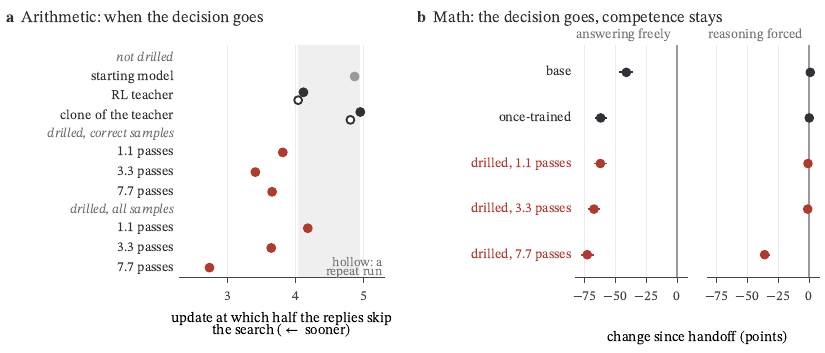}{\linewidth}
\caption{Later training takes the decision first. (a)~The arithmetic task: the
update of the answer-only stress test at which half of each checkpoint's openings
answer at once, without a search; hollow: a second run of the first six updates;
the band spans the checkpoints that were not drilled. (b)~Mathematics on the 9B:
the change in free and forced accuracy between handoff and the end of the
instruction stress test, with 95\% paired intervals. Every model stops choosing to
reason, and forcing gives back the accuracy of every model but the one drilled for
\nv{d.pass.high} passes.}
\label{fig:clocks}
\end{figure}

\paragraph{The decision goes first.} After the instruction stress test no model of
the first 9B run opens a reasoning block when it answers freely, and free accuracy
falls to \nv{mfree.it.min}--\nv{mfree.it.max}\%. Yet with the block opened for it,
every model that was not drilled keeps its competence: the base gains
\nv{mgain.it.base} points and the once-trained model \nv{mgain.it.O140}
(Figure~\ref{fig:clocks}b). At 35B no model opens the block either, though there the
base and the once-trained model also lose forced accuracy. On the arithmetic task
the teacher, the clone and the starting model skip the search on half of their
openings within about five updates of the answer-only stress test, and drilling
moves this point earlier. With the block opened, those three still solve
\nv{tces16.M0.u20}--\nv{tces16.S.u20}\% (Appendix~\ref{app:tcesdecision},
Figure~\ref{fig:clocks}a).

The decision does not always go. After the answer-only stress test the 9B
once-trained model keeps \nv{mfree.b.O140}\% free accuracy, often reasoning in plain
text. The registered prediction that free accuracy falls more than forced accuracy
fails in two cases: the 35B base after the instruction stress test, and Nemotron's
starting model after the answer-only stress test (registered; failed).

\subsection{A few updates of training restore the reasoning, even with almost no mathematics}
\label{app:relearn}

\paragraph{Relearning} (Table~\ref{tab:rl}). The starting points are the four
drilled models broken by the stress tests: the drilled model on its own solutions and
the one on gpt-oss-120b's traces, each after
the instruction stress test and after the answer-only stress test. The once-trained
model after the instruction stress test, trained the same way, is a ceiling control.
The math data are the base model's own correct forced solutions to \nv{d.RL.n}
problems of the training pool that no model had trained on and that the test set
does not contain. They were drawn in pool order after the problems screened for the
first run, with the same screen and keep rule (Appendix~\ref{app:mathdata}). Each
model continues its own adapter at \nv{d.acq.lr} in batches of \nv{d.stage.batch}
with a fresh optimizer, sees each example once, and is evaluated after 1, 2, 5 and
20 updates.

The share recovered is $R_k = (a_k - a_0)/(a_{\mathrm{ref}} - a_0)$, where $a_k$ is
the drilled model's accuracy after $k$ updates and $a_{\mathrm{ref}}$ the matching
once-trained model's after the same stress test (\nv{rl.ref}\%). That is, $R_k$ is
the fraction of the gap to the once-trained model that $k$ updates close. $R_5$ of the
drilled
model after the instruction stress test is at least \nv{reg.RL1.thr} (registered;
held), and the ceiling control moves by at most
\nv{rl.ceil.max} points, within the registered \nv{reg.RL2.thr} (registered; held).
We had fixed in advance that an $R_5$ of at least \nv{reg.RL.suppressed} would mean
that the drilled model's competence was suppressed and restored, and that is the
reading the paper takes.

\paragraph{The format-only control.} Training on mathematics could still be
teaching mathematics. So the two drilled models broken on their own solutions were
also trained on data that show the format of reasoning with almost no mathematics,
in the same way but for 20 updates and evaluated after the last. The data are the
base model's forced responses to the first prompts of the broad chat mix
(Appendix~\ref{app:stages}), kept until \nv{d.RL.n} had a closed reasoning block
followed by an answer. About \nv{r5.wordprob}\% of that mix are worked math word
problems. We had registered that a share of at least \nv{reg.RL.habit} from these
data would make us drop the claim that the loss is more than a lost habit. Both
models passed it (Table~\ref{tab:rl}), so the paper drops that claim.

\begin{table}[htbp]
\centering
\caption{Relearning: forced accuracy (\%) after $k$ updates of reasoning training
from each broken model, the share $R_5$ recovered after five updates with its
95\% interval, and forced accuracy after twenty updates of the format-only data,
with its share $R_{20}$. IT: the instruction stress test; AO: the answer-only
stress test. The once-trained model is the ceiling control.}
\label{tab:rl}
\small
\setlength{\tabcolsep}{4pt}
\begin{tabular}{@{}l r r r r r l l@{}}
\toprule
Origin & $k=0$ & 1 & 2 & 5 & 20 & $R_5$ & Format only, $k=20$ \\
\midrule
Drilled, IT & \nv{rl.D.it.u0} & \nv{rl.D.it.u1} & \nv{rl.D.it.u2} & \nv{rl.D.it.u5} & \nv{rl.D.it.u20} & \nv{rl.D.it.R5} \nv{rl.D.it.R5.ci} & \nv{rl.D.it.hab} (\nv{rl.D.it.habR}) \\
Drilled, AO & \nv{rl.D.b.u0} & \nv{rl.D.b.u1} & \nv{rl.D.b.u2} & \nv{rl.D.b.u5} & \nv{rl.D.b.u20} & \nv{rl.D.b.R5} \nv{rl.D.b.R5.ci} & \nv{rl.D.b.hab} (\nv{rl.D.b.habR}) \\
Drilled on traces, IT & \nv{rl.DT.it.u0} & \nv{rl.DT.it.u1} & \nv{rl.DT.it.u2} & \nv{rl.DT.it.u5} & \nv{rl.DT.it.u20} & \nv{rl.DT.it.R5} \nv{rl.DT.it.R5.ci} & -- \\
Drilled on traces, AO & \nv{rl.DT.b.u0} & \nv{rl.DT.b.u1} & \nv{rl.DT.b.u2} & \nv{rl.DT.b.u5} & \nv{rl.DT.b.u20} & \nv{rl.DT.b.R5} \nv{rl.DT.b.R5.ci} & -- \\
\addlinespace
Once-trained, IT & \nv{rl.O.it.u0} & \nv{rl.O.it.u1} & \nv{rl.O.it.u2} & \nv{rl.O.it.u5} & \nv{rl.O.it.u20} & -- & -- \\
\bottomrule
\end{tabular}
\end{table}

%% file: sections/app-e-divergence.tex
\section{The divergence \texorpdfstring{$B$}{B}}
\label{app:divergence}

$B$ measures how much less likely a trained model makes the base model's own
reasoning, per token, than the base model does (Section~\ref{sec:signature}); on
the arithmetic task the reference is the starting model, the base model with a
format adapter. This appendix asks how $B$ is scored, where in a response it builds
up, whether it predicts the break, and what the later stage does to the drilled
model's probabilities. It supports Section~\ref{sec:signature}: $B$ flags every
drilled model trained without added data at handoff, but its level does not decide
the outcome, and a model sharpened without repetition, three-quarters as much as the
drilled model, was not fragile.

\subsection{How \texorpdfstring{$B$}{B} is scored}
\label{app:topp}

For a trained model $\theta$ we take solutions $y$ that the base model wrote with its
reasoning forced, and average, over the positions $P$ after the prefix, how much less
likely each token is under $\theta$ than under the base model:
\[
B(\theta) = \frac{1}{|P|}\sum_{t\in P}\Big[\log p_{\mathrm{base}}(y_t\mid y_{<t}) - \log p_\theta(y_t\mid y_{<t})\Big].
\]
That is, $B$ is zero when training has left the base model's reasoning as likely as
it was, and positive when training has made it less likely. On mathematics the
samples are drawn at temperature 1 and cut at \nv{d.math.cut} tokens, so $B$
estimates the forward divergence $\mathrm{KL}(p_{\mathrm{base}}\,\|\,p_\theta)$. On
the arithmetic task every checkpoint is scored on one common set of the starting
model's samples.

\paragraph{The top-p offset.} The arithmetic task's common set was drawn at top-p
\nv{d.topp}, which shifts $B$. On such samples $B$ equals the divergence from the
starting model's truncated distribution minus an
offset between \nv{b.offset.lo} and \nv{b.offset.hi} nats per token. The offset is
shared by every checkpoint, so it keeps their order and the gaps between them, but
it can push $B$ below zero.

\paragraph{Full samples and the first \nv{d.tces.cut} tokens}
(Table~\ref{tab:bfull}). The paper's $B$ on the arithmetic task is scored over each
common-set sample's full length, at most \nv{d.budget.short} tokens. The first
scoring used only the first \nv{d.tces.cut} tokens, which hold
\nv{tces.frac.tokens}\% of the common set's tokens. The choice changes the level of
$B$, not the order: every state with positive $B$ has
\nv{n1x.shrink.min}--\nv{n1x.shrink.max}\% less of it over the full samples than
over the first \nv{d.tces.cut} tokens, and the order of the states in the dose
comparison is unchanged.

\paragraph{The noise floor at 35B.} At 35B the measurement itself is noisy. A new
adapter, which should reproduce the base exactly, scored the 35B base's texts with
a mean absolute difference of \nv{m35.pre.mad} nats per token, above the registered
bound of
\nv{m35.pre.thr}. The reason is that this mixture-of-experts model is not
deterministic: two independently created base samplers differ by \nv{m35.pre.two}
on average. On the run's common set, a second base sampler gives
$B=$~\nv{m35.noise.B} \nv{m35.noise.ci} and a mean absolute per-token difference of
\nv{m35.noise.mad}. Every state's $B$, the once-trained
model's \nv{m35.B.O140} included, therefore lies above the floor.

\begin{table}[htbp]
\centering
\caption{$B$ (nats per token) of each arithmetic checkpoint, scored over the full
samples and over the first \nv{d.tces.cut} tokens, with $A$, the share of
positions (\%) at which the checkpoint's most likely token differs from the
starting model's, and the share of the starting model's probability kept on the
alternatives it sampled, both over the full samples. T-u30: the RL teacher;
S-u230: its clone; U and F: drilled on all of the teacher's samples and on its
correct ones, after 20, 60 or 140 updates; U2: a second unfiltered drilled set;
R-P and R-S: the trace-replay and solver-trained models of an earlier study;
FA025, FA1 and FN: the anchored and diluted models.}
\label{tab:bfull}
\footnotesize
\setlength{\tabcolsep}{3.5pt}
\begin{tabular}{@{}l r r r r r@{}}
\toprule
State & Passes & $B$, full & $B$, \nv{d.tces.cut} & $A$ (\%) & Kept \\
\midrule
T-u30 & -- & \nv{n1.b.T} & \nv{b.T} & \nv{n1.a.T} & \nv{n1.alt.T} \\
S-u230 & \nv{d.S.passes} & \nv{n1.b.S} & \nv{b.S} & \nv{n1.a.S} & \nv{n1.alt.S} \\
U-u20 & \nv{d.pass.low} & \nv{n1x.b.U20} & \nv{b.U20} & \nv{n1x.a.U20} & \nv{n1x.alt.U20} \\
F-u20 & \nv{d.pass.low} & \nv{n1x.b.F20} & \nv{b.F20} & \nv{n1x.a.F20} & \nv{n1x.alt.F20} \\
U-u60 & \nv{d.pass.mid} & \nv{n1x.b.U60} & \nv{b.U60} & \nv{n1x.a.U60} & \nv{n1x.alt.U60} \\
F-u60 & \nv{d.pass.mid} & \nv{n1x.b.F60} & \nv{b.F60} & \nv{n1x.a.F60} & \nv{n1x.alt.F60} \\
U-u140 & \nv{d.pass.high} & \nv{n1.b.U140} & \nv{b.U140} & \nv{n1.a.U140} & \nv{n1.alt.U140} \\
F-u140 & \nv{d.pass.high} & \nv{n1.b.F140} & \nv{b.F140} & \nv{n1.a.F140} & \nv{n1.alt.F140} \\
U2-u140 & \nv{d.pass.high} & \nv{n1.b.U2} & \nv{anc.b.U2} & \nv{n1.a.U2} & \nv{n1.alt.U2} \\
R-P-u140 & \nv{d.pass.high} & \nv{n1x.b.RP} & \nv{b.RP} & \nv{n1x.a.RP} & \nv{n1x.alt.RP} \\
R-S-u200 & \nv{d.RS.passes} & \nv{n1x.b.RS} & \nv{b.RS} & \nv{n1x.a.RS} & \nv{n1x.alt.RS} \\
FA025-u140 & \nv{d.pass.high} & \nv{n1.b.FA025} & \nv{anc.b.FA025} & \nv{n1.a.FA025} & \nv{n1.alt.FA025} \\
FA1-u140 & \nv{d.pass.high} & \nv{n1.b.FA1} & \nv{anc.b.FA1} & \nv{n1.a.FA1} & \nv{n1.alt.FA1} \\
FN-u140 & \nv{d.pass.high} & \nv{n1.b.FN} & \nv{anc.b.FN} & \nv{n1.a.FN} & \nv{n1.alt.FN} \\
\bottomrule
\end{tabular}
\end{table}

\subsection{\texorpdfstring{$B$}{B} in every model}

\paragraph{The larger models.} At 35B and on Nemotron the once-trained models leave
the base model's reasoning almost untouched ($B$ of \nv{m35.B.O140} and
\nv{mnem.B.O140}), while the drilled models' $B$ is \nv{m35.B.D140} and
\nv{mnem.B.D140}. Most of it is over-sharpening:
\nv{m35.kept}\% and \nv{mnem.kept}\% of it comes from positions where the drilled
model keeps the base model's top token.

\begin{figure}[t]
\centering
\figfile{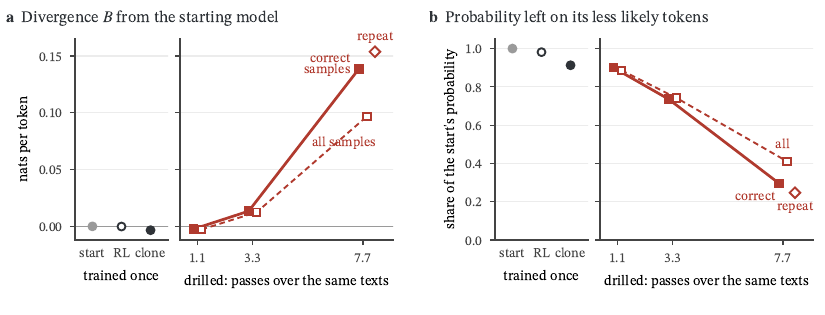}{\linewidth}
\caption{The arithmetic task at handoff, on the starting model's (M0's) samples
drawn at top-p \nv{d.topp} and scored to \nv{d.budget.short} tokens; the top-p
offset (Appendix~\ref{app:topp}) puts the lasting checkpoints slightly below zero
in (a). Left of each panel: M0, the RL teacher and its clone, each trained about
once on its texts. Right: the drilled checkpoints against passes, on the teacher's
correct samples (F, filled) and on all its samples (U, open, set slightly apart),
and a repeat of U (diamond). (a)~$B$; the intervals of its differences from the
clone are narrower than the markers. (b)~Where M0 sampled a less likely token,
the share of its probability that the checkpoint keeps.}
\label{fig:search}
\end{figure}

\paragraph{The arithmetic task} (Figure~\ref{fig:search}). For the teacher and the
clone $B$ is \nv{n1.b.T} and \nv{n1.b.S}, slightly negative because of the top-p
offset. Scored over the full samples, the unfiltered, correct-only and second
unfiltered drilled models have $B$ of \nv{n1.b.U140}, \nv{n1.b.F140} and \nv{n1.b.U2}
nats per token; for scale, the
starting model's own negative log-likelihood of these samples is \nv{n1.nll}. $B$
rises with passes, from \nv{n1x.b.U20} to \nv{n1x.b.U60} and \nv{n1.b.U140} for the
unfiltered drilled model and from \nv{n1x.b.F20} to \nv{n1x.b.F60} and
\nv{n1.b.F140} for the correct-only one. About two-thirds of the drilled
checkpoints' $B$ (\nv{n1.kept.F140}--\nv{n1.kept.U2}\%) comes from positions where
they keep the starting model's most likely token. They change the top choice at
\nv{n1.a.U140}--\nv{n1.a.U2}\% of positions, against \nv{n1.a.T}--\nv{n1.a.S}\% for
the teacher and the clone. Every checkpoint drilled to \nv{d.pass.high} passes
without added data (not anchored or diluted) exceeds the $B$ of every checkpoint
that lasts (the starting model, the teacher and the clone) by at least \nv{n1x.b.gap}
nats per token, and the checkpoints at
\nv{d.pass.mid} passes fall between.

\paragraph{The alternatives at handoff.} Where the starting model sampled one of its
less likely tokens, the teacher and the clone keep
\nv{n1.alt.S}--\nv{n1.alt.T} of the probability the starting model gave it, while
the drilled checkpoints keep only \nv{n1.alt.U2}--\nv{n1.alt.U140}
(Figure~\ref{fig:search}b). Drilling takes most of the probability away from the
alternatives.

\subsection{Where in a response \texorpdfstring{$B$}{B} builds up}

\begin{table}[t]
\centering
\caption{$B$ (nats per token) by position on the arithmetic starting model's
forced texts, first \nv{d.tces.cut} tokens. Positions 1--3 are the forced prefix.
T-u30: the RL teacher; S-u230: its clone; U and F: drilled on all of the teacher's
samples and on its correct ones, after 20, 60 or 140 updates; R-P-u140 and
R-S-u200: the trace-replay and solver-trained models of an earlier study.}
\label{tab:profile}
\small
\setlength{\tabcolsep}{5pt}
\begin{tabular}{@{}l r r r r r r r r@{}}
\toprule
State & 4--7 & 8--15 & 16--31 & 32--63 & 64--127 & 128--255 & 256--511 & 512--1,024 \\
\midrule
T-u30 & \nv{bprof.T.4-7} & \nv{bprof.T.8-15} & \nv{bprof.T.16-31} & \nv{bprof.T.32-63} & \nv{bprof.T.64-127} & \nv{bprof.T.128-255} & \nv{bprof.T.256-511} & \nv{bprof.T.512-1024} \\
S-u230 & \nv{bprof.S.4-7} & \nv{bprof.S.8-15} & \nv{bprof.S.16-31} & \nv{bprof.S.32-63} & \nv{bprof.S.64-127} & \nv{bprof.S.128-255} & \nv{bprof.S.256-511} & \nv{bprof.S.512-1024} \\
U-u20 & \nv{bprof.U20.4-7} & \nv{bprof.U20.8-15} & \nv{bprof.U20.16-31} & \nv{bprof.U20.32-63} & \nv{bprof.U20.64-127} & \nv{bprof.U20.128-255} & \nv{bprof.U20.256-511} & \nv{bprof.U20.512-1024} \\
F-u20 & \nv{bprof.F20.4-7} & \nv{bprof.F20.8-15} & \nv{bprof.F20.16-31} & \nv{bprof.F20.32-63} & \nv{bprof.F20.64-127} & \nv{bprof.F20.128-255} & \nv{bprof.F20.256-511} & \nv{bprof.F20.512-1024} \\
U-u60 & \nv{bprof.U60.4-7} & \nv{bprof.U60.8-15} & \nv{bprof.U60.16-31} & \nv{bprof.U60.32-63} & \nv{bprof.U60.64-127} & \nv{bprof.U60.128-255} & \nv{bprof.U60.256-511} & \nv{bprof.U60.512-1024} \\
F-u60 & \nv{bprof.F60.4-7} & \nv{bprof.F60.8-15} & \nv{bprof.F60.16-31} & \nv{bprof.F60.32-63} & \nv{bprof.F60.64-127} & \nv{bprof.F60.128-255} & \nv{bprof.F60.256-511} & \nv{bprof.F60.512-1024} \\
U-u140 & \nv{bprof.U140.4-7} & \nv{bprof.U140.8-15} & \nv{bprof.U140.16-31} & \nv{bprof.U140.32-63} & \nv{bprof.U140.64-127} & \nv{bprof.U140.128-255} & \nv{bprof.U140.256-511} & \nv{bprof.U140.512-1024} \\
F-u140 & \nv{bprof.F140.4-7} & \nv{bprof.F140.8-15} & \nv{bprof.F140.16-31} & \nv{bprof.F140.32-63} & \nv{bprof.F140.64-127} & \nv{bprof.F140.128-255} & \nv{bprof.F140.256-511} & \nv{bprof.F140.512-1024} \\
R-P-u140 & \nv{bprof.RP.4-7} & \nv{bprof.RP.8-15} & \nv{bprof.RP.16-31} & \nv{bprof.RP.32-63} & \nv{bprof.RP.64-127} & \nv{bprof.RP.128-255} & \nv{bprof.RP.256-511} & \nv{bprof.RP.512-1024} \\
R-S-u200 & \nv{bprof.RS.4-7} & \nv{bprof.RS.8-15} & \nv{bprof.RS.16-31} & \nv{bprof.RS.32-63} & \nv{bprof.RS.64-127} & \nv{bprof.RS.128-255} & \nv{bprof.RS.256-511} & \nv{bprof.RS.512-1024} \\
\bottomrule
\end{tabular}
\end{table}

On the arithmetic task the drilled checkpoints depart from the starting model mainly in the search, and only slightly in the opening (Table~\ref{tab:profile}). From position 8 on, the
checkpoints that last (here the teacher, the clone and the two
checkpoints drilled \nv{d.pass.low} passes) stay within about
\nv{bprof.lasting.max} nats per token of the starting model. The drilled
checkpoints differ only slightly over positions 4--31, where the text mostly
restates the problem, and depart most over positions 128--255. The solver-trained
model differs from the first positions on, consistent with a
different kind of text. The profile is an observation; no experiment varies it.

Over the same first \nv{d.tces.cut} tokens, and over all of the starting model's
branch points, the drilled checkpoints raise the probability of the starting model's
top choice from \nv{br.m0top} to \nv{br.U140}--\nv{br.F140}, while
the checkpoints that last raise it only to \nv{br.last.to}.

\subsection{\texorpdfstring{$B$}{B} as a predictor}

\begin{figure}[htbp]
\centering
\figfile{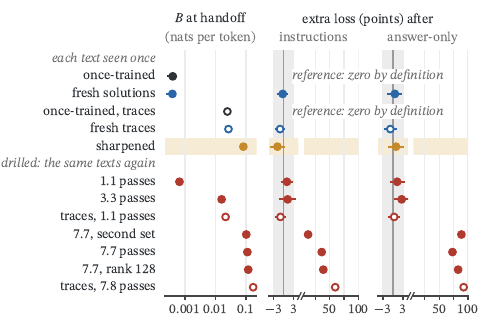}{0.55\linewidth}
\caption{$B$ separates the models drilled for about eight passes but does not track the size of their
break. One row per 9B model; hollow: trained on gpt-oss-120b's traces. Left: $B$ at
handoff (log scale). Middle and right: the extra loss after the instruction and the
answer-only stress tests, in points with 95\% intervals, over the once-trained model
trained on the same kind of texts (zero by definition); grey: the margin of
$\pm$\nv{reg.SH.within} points fixed in advance. The sharpened model (gold) moved its
outputs three-quarters as far as the drilled model and stayed within the margin.}
\label{fig:bflag}
\end{figure}

If $B$ measured fragility, a larger $B$ should mean a larger loss. Its rank correlations come from separating the dose groups; within a group it does not rank the loss.

\paragraph{The arithmetic task.} Across the starting model and the nine
search-trained states with forced data after the answer-only stress test, the rank
correlation between $B$ and the forced points lost by update 20, within
\nv{d.budget.short} tokens, is \nv{rho.B} (\nv{rho.B.RS} with the solver-trained
model). The order of these states by $B$ is the same over the full samples and over
the first \nv{d.tces.cut} tokens, so the registered values stand. But the states
form three dose groups, and the group label alone gives \nv{rho.group} (the number
of passes \nv{rho.passes}). Within groups $B$ does not rank the loss
(\nv{rho.within.low} in the low group without the starting model,
\nv{rho.within.high} in the high group). Handoff accuracy does not rank it either
(forced \nv{rho.fa}, free \nv{rho.ua}). The anchored models show that a low $B$ does
not guarantee that a drilled model lasts (Appendix~\ref{app:anchor}).

\paragraph{Mathematics.} On the 9B models (Figure~\ref{fig:bflag}), $B$ separates
those drilled for about eight passes from the others but does not track the size of
their extra loss, and the sharpened model has a large $B$ without the break. Across
the five states of each run, $B$ ranks the forced loss at Spearman \nv{m.rho.it} and
\nv{m.rho.b} on the 9B,
\nv{m35.rho.it} and \nv{m35.rho.b} at 35B, and \nv{mnem.rho.it} and \nv{mnem.rho.b}
on Nemotron, after the instruction and the answer-only stress tests respectively.
With five points each, all are descriptive.

\subsection{The loss jump}
\label{app:profile}

If $B$ cannot say how much a model will lose, the first update of the later stage
might. One update into the arithmetic task's answer-only stress test, the loss
on the next batch rises for the checkpoints drilled \nv{d.pass.high} passes, from
\nv{lj.U140.u1} to \nv{lj.U140.u2} for the unfiltered one and from \nv{lj.F140.u1} to
\nv{lj.F140.u2} for the correct-only one, and barely moves for the trace-replay model
(\nv{lj.RP.u1} to \nv{lj.RP.u2}), while every other checkpoint's falls
(Table~\ref{tab:lossjump}). Across eleven starting points the
update-2 loss correlates with the change in forced accuracy by update 20 at
Pearson \nv{pearson.loss2}, and across the starting model and nine search-trained
states its rank correlation with the points lost is \nv{rho.loss2}. The update-1
loss, taken before any update on the new task, gives \nv{rho.loss1}, so the signal
needs one update and is not a handoff measure.

The jump marks the start of the break, not the break itself. The correct-only drilled
model's forced accuracy is \nv{tces4.F140.u0}\% at handoff, \nv{tces4.cliff.F140}\%
after two updates and \nv{tces4.F140.u20}\% after twenty, within \nv{d.budget.short}
tokens. In two other stages the jump shrinks or vanishes, though the drilled
competence still erodes. On the same answer-only data at \nv{d.stage.lrlow} it is
smaller (\nv{lj.gentle.U140} and \nv{lj.gentle.F140}, against
\nv{lj.gentle.once.min}--\nv{lj.gentle.once.max} for the checkpoints drilled
\nv{d.pass.low} passes). In the instruction stress test there is none
(\nv{lj.it.U140} and \nv{lj.it.F140} at update 2, against
\nv{lj.it.others.min}--\nv{lj.it.others.max} for the starting model and the
checkpoints drilled \nv{d.pass.low} passes).

\begin{table}[htbp]
\centering
\caption{Loss on the next batch of the arithmetic task's answer-only stress test
(summed over \nv{d.stage.batch} examples); the batches are identical across starting
points, and the update-1 loss is taken before any update. M0: the starting model;
T-u30: the RL teacher; S-u230: its clone; U and F: drilled on all of the teacher's
samples and on its correct ones, after 20, 60 or 140 updates; R-P and R-S: the
trace-replay and solver-trained models of an earlier study.}
\label{tab:lossjump}
\small
\begin{tabular}{@{}l r r r@{}}
\toprule
Origin & Update 1 & Update 2 & Update 3 \\
\midrule
M0 & \nv{lj.M0.u1} & \nv{lj.M0.u2} & \nv{lj.M0.u3} \\
T-u30 & \nv{lj.T.u1} & \nv{lj.T.u2} & \nv{lj.T.u3} \\
S-u230 & \nv{lj.S.u1} & \nv{lj.S.u2} & \nv{lj.S.u3} \\
U-u20 & \nv{lj.U20.u1} & \nv{lj.U20.u2} & \nv{lj.U20.u3} \\
F-u20 & \nv{lj.F20.u1} & \nv{lj.F20.u2} & \nv{lj.F20.u3} \\
U-u60 & \nv{lj.U60.u1} & \nv{lj.U60.u2} & \nv{lj.U60.u3} \\
F-u60 & \nv{lj.F60.u1} & \nv{lj.F60.u2} & \nv{lj.F60.u3} \\
U-u140 & \nv{lj.U140.u1} & \nv{lj.U140.u2} & \nv{lj.U140.u3} \\
F-u140 & \nv{lj.F140.u1} & \nv{lj.F140.u2} & \nv{lj.F140.u3} \\
R-P-u140 & \nv{lj.RP.u1} & \nv{lj.RP.u2} & \nv{lj.RP.u3} \\
R-S-u200 & \nv{lj.RS.u1} & \nv{lj.RS.u2} & \nv{lj.RS.u3} \\
\bottomrule
\end{tabular}
\end{table}

\subsection{What the later stage does to the drilled model}
\label{app:process}

The instruction stress test (at \nv{d.stage.lr}) drives the drilled 9B model off its
own solutions, barely moves the other models off theirs, and makes the base model's
reasoning slightly more likely under the drilled model. We measured how far it moves each 9B
model on two sets of texts: the solutions the model itself wrote at handoff, and the
base model's solutions.

\begin{figure}[htbp]
\centering
\figfile{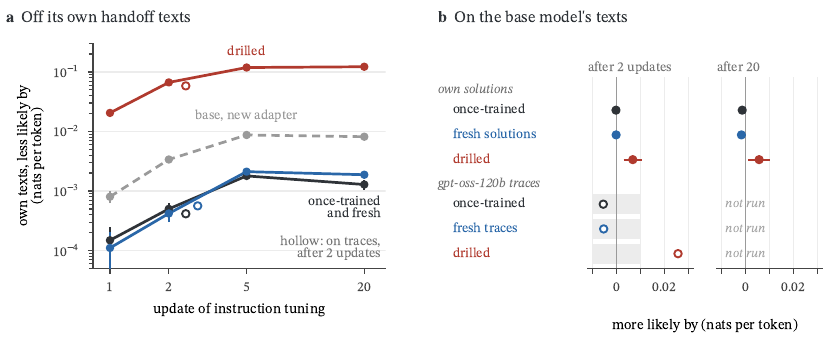}{\linewidth}
\caption{How far the instruction stress test, at \nv{d.stage.lr}, moves each 9B
model. (a)~How much less likely each model's own handoff texts become (the KL from
its handoff, nats per token, log scale, 95\% intervals) after 1, 2, 5 and 20
updates; the base, which starts from a new adapter, dashed; hollow: the models
trained on gpt-oss-120b's traces, after 2 updates. (b)~How much more likely the base
model's texts become under each model after 2 and 20 updates; the shaded band is the
one registered for the models trained on traces, and the drilled one falls outside
it.}
\label{fig:movement}
\end{figure}

\paragraph{The model's own solutions.} Every model fits the instructions about
equally well from the second update on, with training losses at updates 2--5 within
\nv{r1.lossspread.max}\% of one another. (On the broad chat mix, by contrast, the
drilled models start with a higher training loss than the once-trained ones and
catch up only over the last five updates; Appendix~\ref{app:stage}.) Yet the stress
test drives only the drilled model off its own solutions
(Figure~\ref{fig:movement}a). After two updates, the once-trained and
fresh-solution models make their own handoff texts \nv{r1.d.O.u2} and
\nv{r1.d.P.u2} nats per token less likely, and the drilled model its own
\nv{r1.d.D.u2}.

From update 5 on, the drilled model gives its own solutions a lower likelihood than
the untrained base does. After 20 updates their negative log-likelihood is
\nv{r1.own.D20} nats per token under the drilled model and \nv{r1.own.base} under the
base. Part of this drop is expected, because any update that spreads probability
lowers a peaked model's likelihood of its own samples. If the stress test simply
undid the drilling, however, those solutions would become no less likely than they
are under the base.

\paragraph{The base model's solutions.} We had also predicted that the stress test
would move the drilled model further than the others on the base model's texts, and
that prediction failed (Figure~\ref{fig:movement}b). The base model's reasoning
instead becomes slightly more likely under the drilled model, by about
\nv{r3b.D.approx} of the \nv{mb.D140.3} nats per token it had lost. Its accuracy
still falls, and it does not come back within \nv{d.R2.updates} updates
(Appendix~\ref{app:stage}, Figure~\ref{fig:trajectory}).

\paragraph{An out-of-sample test.} We then registered this pattern as a
prediction for the models trained on gpt-oss-120b's traces. Its first part held:
after two updates the drilled model's own solutions lose \nv{mt.d.D.own} nats per
token, \nv{mt.d.ratio} times the once-trained model's \nv{mt.d.O.own}. Its second
part failed. We had predicted that the drilled model's likelihood of the base
model's reasoning would change by less than \nv{mt.p5.band} nats per token after
two updates. It rose by \nv{mt.d.D.toward}, while under the once-trained and
fresh-trace models it fell by \nv{mt.d.O.base} and
\nv{mt.d.P.base}. So the large drop on the model's own solutions replicated, and so
did the movement toward the base, which we had not predicted in either setting. In
both settings the drilled model moved toward the base's reasoning without regaining
the base's competence.

\paragraph{A possible explanation.} One post hoc explanation, which we have not
tested, is that drilling narrows the model's reasoning onto a few sharpened paths
and prunes the base model's alternatives (the drilled 9B model keeps
\nv{mb.alt.D140} of the probability the base gave them), and that the later stage
then flattens those paths without giving back what was pruned. Two later
experiments qualify it. A model sharpened three-quarters as much as the drilled
model, without repetition, was not fragile (Appendix~\ref{app:sharpen}). And a few
updates of reasoning training restore the drilled model (Appendix~\ref{app:relearn}),
so whatever the later stage takes is cheap to give back.

\subsection{Sharpening without repetition}
\label{app:sharpen}

A model sharpened without repetition was not fragile. Most of the drilled model's $B$
comes from over-sharpening, making the base model's likely choices even more likely
(Section~\ref{sec:signature}). If that caused the break, a model sharpened without
repetition should break too, so we trained one (Table~\ref{tab:sh}). For each of
the once-trained model's \nv{d.O.n} problems, in its order, the sharpened model takes
the base model's first correct and
uncapped forced sample at temperature \nv{d.SH.temp} (top-p 1), drawing up to
\nv{d.SH.draws} times. \nv{sh.texts} problems had one; the slots of the rest are
filled by cycling from the start, as for the teacher's traces. It is trained exactly
as the once-trained model.

Before any evaluation its $B$ had to reach \nv{reg.SH.gate} nats per token, or the
model would have been sampled and trained again at temperature 0.3; it reached
\nv{sh.B} \nv{sh.B.ci}, against \nv{sh.B.D} for the drilled model. Its extra loss
over the once-trained model is \nv{sh.ex.it} \nv{sh.ex.it.ci} after the instruction
stress test and \nv{sh.ex.b} \nv{sh.ex.b.ci} after the answer-only stress test. So
the registered reading ``sharpening alone does not break it'' (within
\nv{reg.SH.within} points after both stress tests) applies. The registered prediction
that its answer-only extra loss would stay under \nv{reg.SH.half}, half of the
drilled model's, held.

\begin{table}[htbp]
\centering
\caption{Sharpening without repetition: $B$ at handoff (nats per token) and forced
accuracy (\%) at handoff and after the two stress tests, for the sharpened 9B model
and, for comparison, the once-trained and drilled models.}
\label{tab:sh}
\small
\begin{tabular}{@{}l r r r r@{}}
\toprule
Model & $B$ & Handoff & Instruction stress test & Answer-only stress test \\
\midrule
Sharpened & \nv{sh.B} & \nv{sh.h} & \nv{sh.it} & \nv{sh.b} \\
Once-trained & \nv{sh.B.O} & \nv{m.h.O140} & \nv{m.it.O140} & \nv{m.b.O140} \\
Drilled & \nv{sh.B.D} & \nv{m.h.D140} & \nv{m.it.D140} & \nv{m.b.D140} \\
\bottomrule
\end{tabular}
\end{table}

%% file: sections/app-f-replay.tex
\section{Replay}
\label{app:replay}

A common defence against forgetting is to mix some of the earlier training data into
later training, which is called replay. This appendix asks whether a small dose of
replay protects the drilled model. In the gentler stage it does: replaying the
model's own training texts as \nv{d.RP.replay}\% of each batch removed the break,
both for the model's own solutions and for gpt-oss-120b's traces. In the two stress
tests the same dose reduced the break without removing it. These results support the
replay result of Section~\ref{sec:remedies}, and with it the limit of our claim to
later training that does not replay the reasoning data.

The replay stage is the gentler stage, \nv{d.R2.updates} updates at
\nv{d.stage.lrlow} on the instructions (Appendix~\ref{app:stages}), with one change.
Each batch of \nv{d.stage.batch} holds 30 of its instruction examples, in order, and
2 of the model's own training examples, taken in its training order and cycled, with
their original loss weights. For the drilled models these are their repeated texts.

We had registered that an extra loss of at most \nv{reg.RP.prevent} points, with an
upper bound under \nv{reg.RP.upper}, would mean that replay prevents the break. Both
drilled models meet it. The drilled model's extra loss over the once-trained model
fell from \nv{r2.ex} without replay to \nv{rp.ex.D} \nv{rp.ex.D.ci}, a reduction of
\nv{rp.cut.D}\% (Table~\ref{tab:rp}); we had predicted at least
\nv{reg.RP.mitigate}\% (registered; held). With gpt-oss-120b's traces it fell from
\nv{r4.ex60} to \nv{rp.ex.DT} \nv{rp.ex.DT.ci}, indistinguishable from zero.

\begin{table}[htbp]
\centering
\caption{Replay: forced accuracy (\%) of the 9B models at handoff and after the
gentler stage (\nv{d.R2.updates} updates of instructions at \nv{d.stage.lrlow}), with
\nv{d.RP.replay}\% of each batch replayed from the model's own training texts.}
\label{tab:rp}
\small
\begin{tabular}{@{}l l r r@{}}
\toprule
Training texts & Model & Handoff & After \\
\midrule
Own solutions & Drilled & \nv{rp.D.h} & \nv{rp.D.a} \\
& Once-trained & \nv{rp.O.h} & \nv{rp.O.a} \\
gpt-oss-120b traces & Drilled & \nv{rp.DT.h} & \nv{rp.DT.a} \\
& Once-trained & \nv{rp.OT.h} & \nv{rp.OT.a} \\
\bottomrule
\end{tabular}
\end{table}

\paragraph{Replay in the stress tests.} In the stress tests replay helps but does not
prevent the break. This test was added in revision and registered before
its data. The stages are the paper's two stress tests, twenty updates of
\nv{d.stage.batch} at \nv{d.stage.lr} continuing the drilled adapter, with each batch
holding 30 of the stage's examples in order and the drilled model's next 2 training
texts in its training order, as above. These runs used a newer version of Tinker's
software than the paper's (Appendix~\ref{app:released}). We measure the drilled
model's loss from handoff, because the once-trained model lost nothing in either
stress test without replay and was not rerun.

We registered three readings for that loss. At most 3 points means replay prevents
the break; at most half of the loss without replay means it mitigates the break; more
than three-quarters of it means replay fails. We made one prediction: replay would
mitigate the break in both stress tests. We did not predict prevention, because the
step is three times that of the gentler stage. After the instruction stress test the
drilled model kept \nv{rb.rh.it.with}\%: a loss of \nv{rb.rh.it.loss} points
\nv{rb.rh.it.loss.ci}, against \nv{rb.rh.it.loss.without} without replay, a reduction
of \nv{rb.rh.it.cut}\%. So replay mitigated the break there, as predicted, and left a
loss just above the 3 points we had set for prevention. After the answer-only stress
test it kept \nv{rb.rh.b.with}\%: a loss of \nv{rb.rh.b.loss} \nv{rb.rh.b.loss.ci},
against \nv{rb.rh.b.loss.without} without replay, a reduction of \nv{rb.rh.b.cut}\%.
That is just short of the halving we had predicted. The prediction therefore failed,
because of the answer-only stress test alone.

%% file: sections/app-g-skill.tex
\section{A skill the base model lacks}
\label{app:skill}

Everything else in the paper reshapes a skill the base model already has. This
appendix asks what repetition does when training must teach a procedure that the
base model cannot carry out within a fixed token budget: multiplying two five-digit
base-7 numbers within \nv{d.SK.cap} tokens. Here repetition's main cost was in
learning. The drilled model learned less by handoff than the once-trained model, and
measured from its own handoff it lost no more in the gentler stage (a reading made
after the runs). This supports the boundary drawn in Section~\ref{sec:remedies}: the
large fragility we found belongs to training that reshapes a skill the base model
already has.

\subsection{Choosing the task and its token budget}

We registered two tasks, to be tried in order: multiplying two base-7 numbers of five
and then six digits, and sorting 12 and then 16 words under an alphabet given in the
prompt. The registered rule took the first setting on which the base model's forced
accuracy is at most \nv{reg.SK.cal}\%. Each task has an exact checker and a solver
that writes several distinct worked solutions per problem, by different methods and
in different step orders. For base-7 multiplication these come from long
multiplication with either number on top, conversion through base 10 and back, and
collecting digit products by place.

A first check showed that the base model solves these tasks when given room. On \nv{d.SK.smoke} problems per setting, four draws
each, most of its responses within \nv{d.budget.short} tokens were cut off. Within
\nv{d.budget.mid} tokens it solved \nv{sk.smoke8.b5}\% of the five-digit products,
\nv{sk.smoke8.b6}\% of the six-digit ones, and \nv{sk.smoke8.w12}\% and
\nv{sk.smoke8.w16}\% of the two sorts. Before any sample at a shorter budget, we
redefined the skill as solving within \nv{d.SK.cap} tokens and kept everything else as
registered. That budget was set by the worked solutions (mean \nv{sk.text.mean}
tokens for five digits, longest \nv{sk.text.max}), not by any result of the base
model. At that budget the base model solves \nv{sk.cal}\% of \nv{d.SK.items}
five-digit calibration problems, so the five-digit setting was chosen.

\subsection{Training and results: the cost is in learning}

The three models are built as in mathematics. The drilled model trains on
\nv{d.drill.n} problems with one worked solution each for \nv{d.drill.updates}
updates, the once-trained model on \nv{d.O.n} problems once, and the fresh-solution
model on the drilled model's problems with a different worked solution at each visit.
The solutions of each problem are shuffled, so every model sees a mix of methods. A
registered gate required the once-trained model to beat the base model by at least
\nv{reg.SK.gate} points before the other models were trained. It beat it by
\nv{sk.gate} \nv{sk.gate.ci}, with \nv{sk.gate.cap}\% of responses at the budget
(registered; held). All three models were then tested after the gentler stage and
after the instruction stress test (Table~\ref{tab:sk}).

The registered outcomes compare where the drilled and once-trained models end after
each stage. A break means the drilled model ends at least \nv{reg.SK.break} points
below the once-trained model with its interval above zero; consolidation means it
ends within \nv{reg.SK.within}. The fresh-solution model is a costless remedy if it is
within \nv{reg.SK.within} points of the better of the two, both at handoff and after
the stage. As worded, the gentler stage gives a break: the drilled model ends
\nv{sk.OD.lr} points below the once-trained model \nv{sk.OD.lr.ci}. But the drilled
model also started far lower. Its extra loss from handoff, which we computed only
after the runs, is \nv{sk.ex.lr} \nv{sk.ex.lr.ci}, so we read the gap as one of
learning and report both readings. As a share of each model's handoff accuracy (also
post hoc), the stage cost the drilled model \nv{sk.share.D.lr}\%, the once-trained
model \nv{sk.share.O.lr}\% and the fresh-solution model \nv{sk.share.P.lr}\%.

The other registered outcomes fail in the gentler stage. Consolidation fails, and so
does the costless remedy: the fresh-solution model minus the better of the other two
is \nv{sk.fix.h} points at handoff and \nv{sk.fix.lr} after. The fresh-solution
model's lead over the drilled one at handoff, \nv{sk.PD.h} \nv{sk.PD.h.ci}, was not
registered. After the instruction stress test every model is below half a percent, so
consolidation holds there only trivially (\nv{sk.OD.it} points, \nv{sk.OD.it.ci}). The
stress test removes the skill from all three models.

\begin{table}[htbp]
\centering
\caption{The synthetic skill: forced accuracy (\%) on \nv{d.SK.items} held-out
five-digit base-7 products within \nv{d.SK.cap} tokens, four draws each, at handoff
and after each later stage, and each model's loss in the gentler stage
(\nv{d.R2.updates} updates at \nv{d.stage.lrlow}) with its 95\% interval.}
\label{tab:sk}
\small
\begin{tabular}{@{}l r r r r@{}}
\toprule
Model & Handoff & Gentler stage & Instruction stress test & Loss, gentler stage \\
\midrule
Base & \nv{sk.base} & -- & -- & -- \\
Drilled & \nv{sk.D.h} & \nv{sk.D.lr} & \nv{sk.D.it} & \nv{sk.loss.D.lr} \nv{sk.loss.D.lr.ci} \\
Once-trained & \nv{sk.O.h} & \nv{sk.O.lr} & \nv{sk.O.it} & \nv{sk.loss.O.lr} \nv{sk.loss.O.lr.ci} \\
Fresh-solution & \nv{sk.P.h} & \nv{sk.P.lr} & \nv{sk.P.it} & \nv{sk.loss.P.lr} \nv{sk.loss.P.lr.ci} \\
\bottomrule
\end{tabular}
\end{table}

%% file: sections/app-h-arithmetic.tex
\section{The arithmetic task}
\label{app:tces}

This appendix gives the evidence behind one statement of Section~\ref{sec:math}:
on an arithmetic search task, models drilled on a teacher's samples break in the
answer-only stress test, while the teacher, its starting model and a clone of the
teacher lose at most \nv{tces16.lastingloss.max} points. This line is exploratory:
we registered its predictions (Table~\ref{tab:ledger-arith}), but the paper's
confirmatory tests are on mathematics. One measure recurs below: $B$, how much
less likely the starting model's own forced samples become under a trained model
(the divergence of Section~\ref{sec:signature}, with the starting model as the
reference; Appendix~\ref{app:divergence} discusses it for these models). Besides
the task, its models and the results, the appendix gives the scoring rule and the
token budgets. It also holds the exploratory anchored models, which show that a low
$B$ does not guarantee that a drilled model lasts.

\subsection{The task and its models}

The task is a variant of the Countdown game \citep{gandhi2024sos}: the model
searches for an arithmetic expression, and a verifier checks it exactly. The
starting model is the base model with a format adapter. The teacher is trained
from it by reinforcement learning, with group-relative policy gradients
\citep{shao2024deepseekmath}. The clone imitates \nv{d.S.samples} of the teacher's
samples, each seen \nv{d.S.passes} times on average. Two drilled models cycle
\nv{d.drill.n} of the teacher's samples to \nv{d.pass.high} passes: the unfiltered
drilled model uses them whether right or wrong, and the correct-only drilled model
only the correct ones (Appendix~\ref{app:strict} gives the scoring rule). The
clone learned from \nv{d.S.samples} samples over \nv{d.S.updates} updates and saw
each prompt \nv{d.S.presentations} times over eight distinct solutions. So the
drilled models differ from it in repetition, in the prompts they cover and in
completions per prompt. Table~\ref{tab:arms} lists the codes that the tables use:
M0 is the starting model, T the teacher, S the clone, and U and F the unfiltered
and correct-only drilled models.

The question of this paper first arose on this task. A model trained by
reinforcement learning, and a clone trained on thousands of its samples, each seen
about once, kept their accuracy. A model trained by replaying \nv{d.drill.n}
successful traces from a reinforcement-learning run for \nv{d.pass.high} passes
lost nearly all of its accuracy within twenty updates of an unrelated later stage
(the trace-replay model below). Because the replayed model had seen each of its
traces many times and the clone each sample about once, we tested the effect of
repetition directly, first on this task and then on mathematics.

\subsection{Two models from an earlier replay study}
\label{app:earlier}

An earlier study \citep{sheikh2026duraseed} trained two further models on the
same \nv{d.drill.n} prompts in the same order, and we reuse its second
realization. The trace-replay model (R-P) copies, for each prompt, one trace that
a reinforcement-learning run wrote during its own training, which the verifier
accepted and which ended before the token cap. The solver-trained model (R-S)
copies a deterministic solver's derivation. Both use completion-only cross-entropy
averaged per example. So the trace-replay model differs from the two drilled models
in its source, in success filtering and in loss normalization, and we compare it
with them by passes only.

The trace-replay model breaks as the drilled models do. After
\nv{d.drill.updates} updates of replay it solves \nv{tces4.RP.u0}\% forced within
\nv{d.budget.short} tokens, and twenty updates of the answer-only stress test take
it to \nv{tces4.RP.u20}\%. The solver-trained model does not, and so it is a
counterexample to reading $B$ as a warning. At \nv{d.RS.passes} passes it is comparably far from the starting model ($B=$~\nv{n1x.b.RS} over the
full samples, against \nv{n1.b.U140} and \nv{n1.b.F140} for the unfiltered and
correct-only drilled models). Yet it loses \nv{tces4loss.RS} points in the same
twenty updates, from \nv{tces4.RS.u0}\% to \nv{tces4.RS.u20}\%, and keeps writing
derivations. Its text is short and unlike the starting model's, it teaches a
procedure the starting model lacks, and its loss is a per-example mean. Which of
these matters is untested.

\subsection{The drilled models search without finding}
\label{app:search}

On mathematics the broken drilled models stop early (Appendix~\ref{app:mathfail});
here they keep searching without finding. After twenty updates of the answer-only
stress test, \nv{tces16cap.S.u20}\% of the clone's forced searches reach the
\nv{d.budget.long}-token cap, against \nv{tces16cap.U140.u20}\% and
\nv{tces16cap.F140.u20}\% for the unfiltered and correct-only drilled models
(\nv{tces16capn.U140} and \nv{tces16capn.F140} of their \nv{tces16.draws}
responses). A four-times larger budget recovers little
(Table~\ref{tab:budget-tces}). A scan of the capped texts for any expression the
verifier accepts finds one in only \nv{tces16scan.U140} of the unfiltered model's
and \nv{tces16scan.F140} of the correct-only model's.

\subsection{The decision to search}
\label{app:tcesdecision}

On this task we read the decision to reason from short openings. We sample four
\nv{d.decision.tokens}-token openings per problem, and an opening skips reasoning
if it begins with the answer tag. The half-skip update is the update, interpolated
between probes, at which half of them skip. In the answer-only stress test the
teacher, the clone and the starting model reach it after \nv{tces.hs.T},
\nv{tces.hs.S} and \nv{tces.hs.M0} updates; a second run of the first six updates
gave \nv{tces.hs.rep.T} and \nv{tces.hs.rep.S} for the teacher and the clone
(registered; held). The unfiltered drilled model, drilled \nv{d.pass.high} passes,
reaches it sooner, after \nv{tces.hs.U140} updates, and the correct-only one a
little later than the unfiltered one (Figure~\ref{fig:clocks}a; registered for both drilled models against the clone; failed in part, holding for the unfiltered one only). The instruction stress test changes the
decision in another way: no checkpoint learns to answer first, but all stop
searching and answer with one-line derivations (Appendix~\ref{app:budgets}).

\subsection{The long horizon}

We continued the answer-only stress test past twenty updates, to \nv{d.horizon},
from the arithmetic checkpoints, one run per starting point and within
\nv{d.budget.short} tokens. The registered predictions that forced accuracy would
hold through update 80, and that the trained checkpoints would keep their margin
over the starting model, failed (Table~\ref{tab:ledger-arith}, C-1 to C-3 and C2);
the records in the repository give these probes. What repeats across runs is the
half-skip update and the drilled models' break within twenty updates, which the
main text claims. The main text does not claim the late half-lives (the updates at which
accuracy has halved).

\subsection{The anchor, an exploratory test}
\label{app:anchor}

\paragraph{Anchor and dilution.} If drilling breaks a model by moving it away from
its starting model, pulling it back during the drilling might protect it. Three
further models keep the correct-only drilled model's drilling and add something to
every batch. The two anchored models add \nv{d.anchor.extra} fresh forced samples
of the starting model, weighted \nv{d.anchor.whigh} (the fully weighted model, FA1)
or \nv{d.anchor.wlow} (the quarter-weighted model, FA025) in the loss. Fitting the
starting model's samples follows the sampled gradient of $B$
(Section~\ref{sec:signature}), so the anchor should pull the model toward the
starting model without removing any repetition. The diluted model (FN) adds
\nv{d.anchor.extra} fresh No Robots examples instead, which dilutes the drilling
equally without that pull. It is not neutral, since instruction data alone wear
down drilled competence (Appendix~\ref{app:budgets}). A fourth model, the second
unfiltered drilled model (U2), repeats the unfiltered drilling on a second subset.

The anchor rehearses the starting model's forced reasoning, so a protected anchored
model could be keeping the starting model's rehearsed competence rather than
drilled competence. The two weights bound this reading only partly. The diluted
model, meant to separate dilution from the anchor's pull, cannot be read, because
most of its drilling did not take hold (below).

\paragraph{What the anchor did} (Table~\ref{tab:anchor}). Adding the starting
model's samples kept more of the search. After the answer-only stress test the
correct-only drilled model solves \nv{tces16.F140.u20}\%, and the anchored models
weighted \nv{d.anchor.whigh} and \nv{d.anchor.wlow} solve \nv{anc.a16.FA1}\% and
\nv{anc.a16.FA025}\%, differences of \nv{anc.d16.FA1} points \nv{anc.d16ci.FA1}
and \nv{anc.d16.FA025} \nv{anc.d16ci.FA025}. Their $B$ falls from \nv{n1.b.F140}
to \nv{n1.b.FA025} (weight \nv{d.anchor.wlow}) and \nv{n1.b.FA1} (weight
\nv{d.anchor.whigh}), so the registered dose order holds for both accuracy and $B$.
The drilling still happened, as the checks below show. The second unfiltered
drilled model breaks as the first did, falling to \nv{anc.a16.U2}\% with
$B=$~\nv{n1.b.U2}.

\paragraph{Why the test of protection cannot be read.} The registered test of
protection is nevertheless not interpretable, because one of its registered checks
failed. First, the drilling had to happen: negative log-likelihood on the
correct-only model's drilled texts had to fall by at least \nv{d.anchor.fit}\% of
that model's own fall (weight \nv{d.anchor.whigh}: \nv{anc.fit.FA1}\%, weight
\nv{d.anchor.wlow}: \nv{anc.fit.FA025}\%, diluted: \nv{anc.fit.FN}\%). Second,
forced accuracy at handoff within \nv{d.budget.short} tokens had to be at parity,
within \nv{d.anchor.parity} points of the correct-only drilled model's
\nv{anc.h4.F140}\%. Third, $B$ had to move, and it did for both anchored models. The
parity check failed: the anchored models trail by \nv{anc.gap4.FA1} and
\nv{anc.gap4.FA025} points.

They trail because their searches run longer. The anchor texts are long searches,
\nv{anc.samples.capped}\% of them reaching the \nv{d.budget.short}-token cap. From
\nv{d.budget.short} to \nv{d.budget.long} tokens, in the same probes, the
correct-only drilled model gains \nv{anc.gain.F140} points and the fully weighted
model \nv{anc.gain.FA1}. Within \nv{d.budget.long} tokens the handoff gaps shrink
to \nv{anc.gap16.FA1} and \nv{anc.gap16.FA025} points, so the quarter-weighted
model would pass the margin and the fully weighted one would not: at handoff the
correct-only drilled model solves \nv{tces16.F140.u0}\%, the quarter-weighted model
\nv{anc.h16.FA025}\% (level with it) and the fully weighted one \nv{anc.h16.FA1}\%
(still behind).

With a check failed, the registered analysis reads the protection prediction as not
interpretable. The dose order still holds for accuracy and $B$, and the second
unfiltered drilled model repeats the break. The diluted model solves
\nv{anc.a16.FN}\%, close to the quarter-weighted model, but cannot be read, because
the added data kept most of the drilling from happening.

\paragraph{A low $B$ is not sufficient.} Anchoring also shows that a low $B$ does
not mean a model will last. The fully weighted anchored model's $B$ is
\nv{n1.b.FA1}, and it keeps \nv{n1.alt.FA1} of the starting model's probability on
the alternatives the starting model sampled, as a checkpoint that lasts does. Yet
within \nv{d.budget.long} tokens the starting model, the teacher and the clone lose
\nv{tces16.lastingloss.min}--\nv{tces16.lastingloss.max} points, and the fully
weighted model loses \nv{anc.loss16.FA1}. The anchor trains on samples of the very
distribution $B$ is measured on, so its low $B$ is partly by construction, and some
of what drilling does lies outside what $B$ measures.

\begin{table}[t]
\centering
\caption{The anchored, diluted and second unfiltered drilled models of the
arithmetic task: forced accuracy (\%) at both budgets, free accuracy at handoff,
and $B$ over the full samples. F-u140: the correct-only drilled model; FA1 and
FA025: anchored with weights \nv{d.anchor.whigh} and \nv{d.anchor.wlow}; FN:
diluted; U2: the second unfiltered drilled model. At cap: share of responses at
the longer budget's cap; Strict: the strict score
(Appendix~\ref{app:strict}). Differences from F-u140 after the answer-only stress
test within \nv{d.budget.long} tokens, with paired intervals: FA1 \nv{anc.d16.FA1}
\nv{anc.d16ci.FA1}, FA025 \nv{anc.d16.FA025} \nv{anc.d16ci.FA025}, FN
\nv{anc.d16.FN} \nv{anc.d16ci.FN}, U2 \nv{anc.d16.U2} \nv{anc.d16ci.U2}.}
\label{tab:anchor}
\small
\setlength{\tabcolsep}{5pt}
\begin{tabular}{@{}l r r r r r r r r@{}}
\toprule
& \multicolumn{3}{c}{At handoff} & \multicolumn{4}{c}{After the answer-only stress test} & \\
\cmidrule(lr){2-4}\cmidrule(lr){5-8}
Model & Forced, \nv{d.budget.short} & Forced, \nv{d.budget.long} & Free, \nv{d.budget.short} & \nv{d.budget.short} & \nv{d.budget.long} & At cap & Strict & $B$ \\
\midrule
F-u140 & \nv{anc.h4.F140} & \nv{tces16.F140.u0} & \nv{anc.free.F140} & \nv{anc.a4.F140} & \nv{tces16.F140.u20} & \nv{tces16cap.F140.u20} & \nv{tces16s.F140.u20} & \nv{n1.b.F140} \\
FA1 & \nv{anc.h4.FA1} & \nv{anc.h16.FA1} & \nv{anc.free.FA1} & \nv{anc.a4.FA1} & \nv{anc.a16.FA1} & \nv{anc.cap.FA1} & \nv{anc.strict.FA1} & \nv{n1.b.FA1} \\
FA025 & \nv{anc.h4.FA025} & \nv{anc.h16.FA025} & \nv{anc.free.FA025} & \nv{anc.a4.FA025} & \nv{anc.a16.FA025} & \nv{anc.cap.FA025} & \nv{anc.strict.FA025} & \nv{n1.b.FA025} \\
FN & \nv{anc.h4.FN} & \nv{anc.h16.FN} & \nv{anc.free.FN} & \nv{anc.a4.FN} & \nv{anc.a16.FN} & \nv{anc.cap.FN} & \nv{anc.strict.FN} & \nv{n1.b.FN} \\
U2 & \nv{anc.h4.U2} & \nv{anc.h16.U2} & \nv{anc.free.U2} & \nv{anc.a4.U2} & \nv{anc.a16.U2} & \nv{anc.cap.U2} & \nv{anc.strict.U2} & \nv{n1.b.U2} \\
\bottomrule
\end{tabular}
\end{table}

\subsection{Strict and resumed scores}
\label{app:strict}

\paragraph{The rule.} The resumed and the strict score differ only on responses that restate the
placeholder \texttt{<answer>EXPRESSION</answer>} from the instructions before
answering. In the resumed score, which we use for accuracy throughout, a response ends at its first
answer tag that contains a number, and that tag is scored. A tag holding only the
placeholder is the model restating the instructions rather than answering, so
generation continues past it, within the same token budget. The strict score ends
a response at any answer tag and counts a restated placeholder as a wrong answer.

How often a model restates the placeholder is a habit of format that the later
stages change. After the answer-only stress test \nv{tces16r.S.u20}\% of the
clone's responses restate it within \nv{d.budget.long} tokens, against
\nv{tces16r.S.u0}\% at handoff, while the instruction stress test lowers the
starting model's rate from \nv{tces4r.M0.u0}\% to \nv{tces4ir.M0}\% within
\nv{d.budget.short} tokens. So strict scores move with the habit as well as with
competence. In the instruction stress test the starting model's strict score rises
from \nv{tces4s.M0.u0}\% to \nv{tces4is.M0}\% while its resumed score stays level,
and in the answer-only stress test the clone's strict score falls from
\nv{tces16s.S.u0}\% to \nv{tces16s.S.u20}\% while its resumed score holds. The
resumed rule was adopted after only the first long-horizon probes had been read,
and every later probe records both scores.

\paragraph{What survives the strict rule} (Tables~\ref{tab:strict-handoff},
\ref{tab:strict-after} and~\ref{tab:strict-16k}). The correct-only drilled model
breaks under both rules, from \nv{tces16s.F140.u0}\% to \nv{tces16s.F140.u20}\%
strict within \nv{d.budget.long} tokens. At handoff, though, strict accuracy points
the wrong way. Within \nv{d.budget.short} tokens the checkpoints that last (the
starting model, the teacher, the clone and the checkpoints drilled
\nv{d.pass.low} passes) score
\nv{tces4s.handoff.lasting.min}--\nv{tces4s.handoff.lasting.max}\%, and the
correct-only drilled model, which later breaks, scores \nv{tces4s.F140.u0}\%.
Within \nv{d.budget.long} tokens the strict rule also weakens the contrast between
lasting and drilled models: the clone goes from \nv{tces16s.S.u0}\% to
\nv{tces16s.S.u20}\% and the unfiltered drilled model from \nv{tces16s.U140.u0}\%
to \nv{tces16s.U140.u20}\%, both with high restatement rates.

On held-out problem families at handoff, within \nv{d.budget.short} tokens, the
resumed and strict scores are \nv{e2.M0}\% and \nv{e2s.M0}\% for the starting
model, \nv{e2.T}\% and \nv{e2s.T}\% for the teacher, \nv{e2.S}\% and \nv{e2s.S}\%
for the clone, \nv{e2.RP}\% and \nv{e2s.RP}\% for the trace-replay model, and
\nv{e2.RS}\% and \nv{e2s.RS}\% for the solver-trained model.

\begin{table}[htbp]
\centering
\caption{Arithmetic accuracy (\%) at handoff within \nv{d.budget.short} tokens:
forced, resumed and strict, the share of forced responses resumed, and free
accuracy. Free accuracy on the arithmetic task was probed only at this budget.
M0: the starting model; T-u30: the teacher trained by reinforcement learning;
S-u230: its clone; U and F: drilled on all of the teacher's samples (unfiltered)
and on its correct ones only, after 20, 60 or 140 updates; R-P and R-S: the
trace-replay and solver-trained models.}
\label{tab:strict-handoff}
\small
\begin{tabular}{@{}l r r r r@{}}
\toprule
State & Resumed & Strict & Share resumed & Free \\
\midrule
M0 & \nv{tces4.M0.u0} & \nv{tces4s.M0.u0} & \nv{tces4r.M0.u0} & \nv{tces4free.M0.u0} \\
T-u30 & \nv{tces4.T.u0} & \nv{tces4s.T.u0} & \nv{tces4r.T.u0} & \nv{tces4free.T.u0} \\
S-u230 & \nv{tces4.S.u0} & \nv{tces4s.S.u0} & \nv{tces4r.S.u0} & \nv{tces4free.S.u0} \\
U-u20 & \nv{tces4.U20.u0} & \nv{tces4s.U20.u0} & \nv{tces4r.U20.u0} & \nv{tces4free.U20.u0} \\
F-u20 & \nv{tces4.F20.u0} & \nv{tces4s.F20.u0} & \nv{tces4r.F20.u0} & \nv{tces4free.F20.u0} \\
U-u60 & \nv{tces4.U60.u0} & \nv{tces4s.U60.u0} & \nv{tces4r.U60.u0} & \nv{tces4free.U60.u0} \\
F-u60 & \nv{tces4.F60.u0} & \nv{tces4s.F60.u0} & \nv{tces4r.F60.u0} & \nv{tces4free.F60.u0} \\
U-u140 & \nv{tces4.U140.u0} & \nv{tces4s.U140.u0} & \nv{tces4r.U140.u0} & \nv{tces4free.U140.u0} \\
F-u140 & \nv{tces4.F140.u0} & \nv{tces4s.F140.u0} & \nv{tces4r.F140.u0} & \nv{tces4free.F140.u0} \\
R-P-u140 & \nv{tces4.RP.u0} & \nv{tces4s.RP.u0} & \nv{tces4r.RP.u0} & \nv{tces4free.RP.u0} \\
R-S-u200 & \nv{tces4.RS.u0} & \nv{tces4s.RS.u0} & \nv{tces4r.RS.u0} & \nv{tces4free.RS.u0} \\
\bottomrule
\end{tabular}
\end{table}

\begin{table}[htbp]
\centering
\caption{Arithmetic forced accuracy (\%) after twenty updates of each later stage,
within \nv{d.budget.short} tokens, as resumed / strict: the answer-only stress
test, the same answer-only data at a third of its rate, and the instruction stress
test. Codes as in Table~\ref{tab:strict-handoff}.}
\label{tab:strict-after}
\small
\setlength{\tabcolsep}{4pt}
\begin{tabular}{@{}l r r r@{}}
\toprule
State & Answer-only, \nv{d.stage.lr} & Answer-only, \nv{d.stage.lrlow} & Instruction stress test \\
\midrule
M0 & \nv{tces4.M0.u20} / \nv{tces4s.M0.u20} & -- & \nv{tces4i.M0} / \nv{tces4is.M0} \\
T-u30 & \nv{tces4.T.u20} / \nv{tces4s.T.u20} & -- & -- \\
S-u230 & \nv{tces4.S.u20} / \nv{tces4s.S.u20} & -- & -- \\
U-u20 & \nv{tces4.U20.u20} / \nv{tces4s.U20.u20} & \nv{tces4g.U20} / \nv{tces4gs.U20} & \nv{tces4i.U20} / \nv{tces4is.U20} \\
F-u20 & \nv{tces4.F20.u20} / \nv{tces4s.F20.u20} & \nv{tces4g.F20} / \nv{tces4gs.F20} & \nv{tces4i.F20} / \nv{tces4is.F20} \\
U-u60 & \nv{tces4.U60.u20} / \nv{tces4s.U60.u20} & -- & -- \\
F-u60 & \nv{tces4.F60.u20} / \nv{tces4s.F60.u20} & -- & -- \\
U-u140 & \nv{tces4.U140.u20} / \nv{tces4s.U140.u20} & \nv{tces4g.U140} / \nv{tces4gs.U140} & \nv{tces4i.U140} / \nv{tces4is.U140} \\
F-u140 & \nv{tces4.F140.u20} / \nv{tces4s.F140.u20} & \nv{tces4g.F140} / \nv{tces4gs.F140} & \nv{tces4i.F140} / \nv{tces4is.F140} \\
R-P-u140 & \nv{tces4.RP.u20} / \nv{tces4s.RP.u20} & -- & -- \\
R-S-u200 & \nv{tces4.RS.u20} / \nv{tces4s.RS.u20} & -- & -- \\
\bottomrule
\end{tabular}
\end{table}

\begin{table}[htbp]
\centering
\caption{Arithmetic forced accuracy (\%) within \nv{d.budget.long} tokens, as
resumed / strict (share resumed). Codes as in Table~\ref{tab:strict-handoff}.}
\label{tab:strict-16k}
\small
\setlength{\tabcolsep}{4pt}
\begin{tabular}{@{}l r r@{}}
\toprule
State & Handoff & After the answer-only stress test \\
\midrule
M0 & \nv{tces16.M0.u0} / \nv{tces16s.M0.u0} (\nv{tces16r.M0.u0}) & \nv{tces16.M0.u20} / \nv{tces16s.M0.u20} (\nv{tces16r.M0.u20}) \\
T-u30 & \nv{tces16.T.u0} / \nv{tces16s.T.u0} (\nv{tces16r.T.u0}) & \nv{tces16.T.u20} / \nv{tces16s.T.u20} (\nv{tces16r.T.u20}) \\
S-u230 & \nv{tces16.S.u0} / \nv{tces16s.S.u0} (\nv{tces16r.S.u0}) & \nv{tces16.S.u20} / \nv{tces16s.S.u20} (\nv{tces16r.S.u20}) \\
U-u140 & \nv{tces16.U140.u0} / \nv{tces16s.U140.u0} (\nv{tces16r.U140.u0}) & \nv{tces16.U140.u20} / \nv{tces16s.U140.u20} (\nv{tces16r.U140.u20}) \\
F-u140 & \nv{tces16.F140.u0} / \nv{tces16s.F140.u0} (\nv{tces16r.F140.u0}) & \nv{tces16.F140.u20} / \nv{tces16s.F140.u20} (\nv{tces16r.F140.u20}) \\
\bottomrule
\end{tabular}
\end{table}

\subsection{Token budgets}
\label{app:budgets}

A sample counts as correct within a budget only if it finished correct inside it
(Appendix~\ref{app:eval}). The arithmetic probes were first read within
\nv{d.budget.short} tokens and later to \nv{d.budget.long}.

\paragraph{Handoff and the answer-only stress test} (Table~\ref{tab:budget-tces}).
At handoff every state solves \nv{tces16.gain.min}--\nv{tces16.gain.max} points
more within \nv{d.budget.long} tokens than within \nv{d.budget.short}, and at
\nv{d.budget.long} all five are within \nv{tces16.handoff.spread} points of one
another. After twenty updates of the answer-only stress test the budget separates
the drilled states from the rest. Within \nv{d.budget.long} tokens the starting
model, the teacher and the clone lose
\nv{tces16.lastingloss.min}--\nv{tces16.lastingloss.max} points, the drilled states
about \nv{tces16loss.F140}, and a four-times larger budget recovers little. Read
within \nv{d.budget.short} tokens, these probes agree with the earlier probes capped
there, within sampling error: for the clone after the answer-only stress test, for
example, \nv{tces16w4.S.u20}\% against \nv{tces4.S.u20}\%.

\begin{table}[t]
\centering
\caption{Arithmetic forced accuracy (\%) within each budget, from the
\nv{d.budget.long}-token probes; ``Over'' and ``Cap'' are the shares of responses
longer than \nv{d.budget.short} tokens and at the \nv{d.budget.long} cap. M0: the
starting model; T-u30: the teacher trained by reinforcement learning; S-u230: its
clone; U-u140 and F-u140: drilled \nv{d.pass.high} passes on all of the teacher's
samples (unfiltered) and on its correct ones only.}
\label{tab:budget-tces}
\small
\setlength{\tabcolsep}{5pt}
\begin{tabular}{@{}l r r r | r r r r r@{}}
\toprule
& \multicolumn{3}{c|}{At handoff} & \multicolumn{5}{c}{After the answer-only stress test} \\
State & \nv{d.budget.short} & \nv{d.budget.mid} & \nv{d.budget.long} & \nv{d.budget.short} & \nv{d.budget.mid} & \nv{d.budget.long} & Over & Cap \\
\midrule
M0 & \nv{tces16w4.M0.u0} & \nv{tces16w8.M0.u0} & \nv{tces16.M0.u0} & \nv{tces16w4.M0.u20} & \nv{tces16w8.M0.u20} & \nv{tces16.M0.u20} & \nv{tces16over4.M0.u20} & \nv{tces16cap.M0.u20} \\
T-u30 & \nv{tces16w4.T.u0} & \nv{tces16w8.T.u0} & \nv{tces16.T.u0} & \nv{tces16w4.T.u20} & \nv{tces16w8.T.u20} & \nv{tces16.T.u20} & \nv{tces16over4.T.u20} & \nv{tces16cap.T.u20} \\
S-u230 & \nv{tces16w4.S.u0} & \nv{tces16w8.S.u0} & \nv{tces16.S.u0} & \nv{tces16w4.S.u20} & \nv{tces16w8.S.u20} & \nv{tces16.S.u20} & \nv{tces16over4.S.u20} & \nv{tces16cap.S.u20} \\
U-u140 & \nv{tces16w4.U140.u0} & \nv{tces16w8.U140.u0} & \nv{tces16.U140.u0} & \nv{tces16w4.U140.u20} & \nv{tces16w8.U140.u20} & \nv{tces16.U140.u20} & \nv{tces16over4.U140.u20} & \nv{tces16cap.U140.u20} \\
F-u140 & \nv{tces16w4.F140.u0} & \nv{tces16w8.F140.u0} & \nv{tces16.F140.u0} & \nv{tces16w4.F140.u20} & \nv{tces16w8.F140.u20} & \nv{tces16.F140.u20} & \nv{tces16over4.F140.u20} & \nv{tces16cap.F140.u20} \\
\bottomrule
\end{tabular}
\end{table}

\paragraph{The dose within \nv{d.budget.short} tokens.} The drop in forced accuracy
over twenty updates of the answer-only stress test grows with passes. We give it as
the change in accuracy, so a drop is negative. The clone changes by
\nv{tces4.dose.S} points, the teacher by \nv{tces4.dose.T} and the starting model
by \nv{tces4.dose.M0}. At \nv{d.pass.low}, \nv{d.pass.mid} and \nv{d.pass.high}
passes the unfiltered drilled model changes by \nv{tces4.dose.U.low},
\nv{tces4.dose.U.mid} and \nv{tces4.dose.U.high} points, and the correct-only one
by \nv{tces4.dose.F.low}, \nv{tces4.dose.F.mid} and \nv{tces4.dose.F.high}. The
\nv{d.pass.high}-pass checkpoint's change minus the \nv{d.pass.low}-pass
checkpoint's is \nv{tces4.dmo.U.b} \nv{tces4.dmo.U.b.ci} for the unfiltered and
\nv{tces4.dmo.F.b} \nv{tces4.dmo.F.b.ci} for the correct-only drilled model. On the
same answer-only data at \nv{d.stage.lrlow} it is \nv{tces4.dmo.U.g} and
\nv{tces4.dmo.F.g}, and in the instruction stress test \nv{tces4.dmo.U.i} and
\nv{tces4.dmo.F.i}.

\paragraph{The instruction stress test and the lower learning rate.} These
arithmetic probes were read only within \nv{d.budget.short} tokens. In the
instruction stress test no checkpoint learns to answer first, but all stop
searching. Answering freely, the checkpoints drilled only \nv{d.pass.low} passes
write one-line derivations (median \nv{tces4i.median.min}--\nv{tces4i.median.max}
tokens); forced, the starting model and those checkpoints still solve
\nv{tces4i.keep.min}--\nv{tces4i.keep.max}\%. In the instruction stress test and on
the answer-only data at \nv{d.stage.lrlow}, the checkpoints drilled
\nv{d.pass.low} passes lose at most \nv{tces4.once.maxloss} point, while those
drilled \nv{d.pass.high} passes lose far more (Table~\ref{tab:strict-after}).

\paragraph{\nv{d.pass.mid} passes on both tasks.} \nv{d.budget.short} tokens is
the one budget at which both tasks were read at \nv{d.pass.mid} passes. There the
drilled 9B math model keeps its accuracy: it goes from \nv{mbud.4096.h.D60}\% at
handoff to \nv{mbud.4096.b.D60}\% after the answer-only stress test and
\nv{mbud.4096.it.D60}\% after the instruction stress test. The arithmetic
checkpoints at \nv{d.pass.mid} passes lose \nv{tces4loss.U60} (unfiltered) and
\nv{tces4loss.F60} (correct-only) points in the answer-only stress test. So at this
number of passes the arithmetic models already lose accuracy where the math model
does not.

%% file: sections/app-i-ledger.tex
\section{Registered predictions and outcomes}
\label{app:ledger}

This appendix lists every prediction we registered and what happened to it, so
that a reader can see which results were predicted and which predictions failed.
Before each experiment we registered its design, its predictions with numerical
thresholds, and what each possible result would mean for our claims. The records
are in the repository, and Section~\ref{sec:setup} gives the counts.

For the original experiments we registered \nv{reg.total} predictions. Of these,
\nv{reg.held} held, \nv{reg.failed} failed (\nv{reg.failed.partial} of them in
part) and \nv{reg.unread} could not be read. Of the failures, \nv{reg.failed.arith}
are on the arithmetic task. The experiments added in revision
(Appendices~\ref{app:released}, \ref{app:masking} and~\ref{app:replay};
\texttt{records/revision-2.md}) registered \nv{rb.reg.total} more predictions, of
which \nv{rb.reg.failed} failed (Table~\ref{tab:ledger-revision2}).

There are four tables. Table~\ref{tab:ledger-revision2} holds the
predictions added in revision. Tables~\ref{tab:ledger-math}
and~\ref{tab:ledger-revision} list the \nv{reg.math} confirmatory predictions on
mathematics, with the rival reading T2 beside T1, and Table~\ref{tab:ledger-arith}
lists the \nv{reg.arith} predictions of the exploratory line on arithmetic. Each
row gives the prediction with its registered threshold, what we observed and the outcome. In the last three tables a final column names the claim that the
prediction bears on.

\begin{table}[htbp]
\centering
\caption{Predictions registered for the experiments added in revision, with their
outcomes. Extra loss: the drilled model's over the once-trained model, in points.
Replay, stress tests: both the instruction and the answer-only stress test; full fine-tuning, stress: the instruction stress test.}
\label{tab:ledger-revision2}
\footnotesize
\setlength{\tabcolsep}{4pt}
\begin{tabular}{@{}p{0.23\linewidth} p{0.29\linewidth} p{0.21\linewidth} l@{}}
\toprule
Experiment & Prediction & Observed & Outcome \\
\midrule
Replay, stress tests & loss at most half that without replay, both stages & \nv{rb.rh.it.loss} vs \nv{rb.rh.it.loss.without}; \nv{rb.rh.b.loss} vs \nv{rb.rh.b.loss.without} & failed (answer-only) \\
New adapter, one-epoch stage & extra loss interval above zero & \nv{rb.rc.epoch.ex.D} \nv{rb.rc.epoch.ex.D.ci} & failed \\
& extra loss at least 5 & \nv{rb.rc.epoch.ex.D} & failed \\
& fresh-solution within 3 of once-trained & \nv{rb.rc.epoch.ex.P} & held \\
End-of-turn mask & recovered share under 0.5, both stages & \nv{rb.em.it.share}; \nv{rb.em.b.share} & held \\
Instruct student & gain at handoff at least 5 & \nv{rb.is.gain.O} & failed \\
& drilled extra loss interval above zero & \nv{rb.is.forced.epoch.ex.D}; \nv{rb.is.forced.it-stress.ex.D} & failed \\
& fresh-solution within 3 of once-trained & \nv{rb.is.forced.it-stress.ex.P} & failed \\
Five passes & weight change in the gap, about \nv{rb.p5.pred}\% & \nv{rb.wc.D-u91}\% & held \\
Later updates & drilled $U$ factors at least 1.5 times & \nv{rb.am.ratioB} times & held \\
& drilled weights move at least twice as far & \nv{rb.am.ratio5} times & held \\
Full fine-tuning, stress & extra loss interval above zero & \nv{rb.rc.it-stress-full.ex.D} \nv{rb.rc.it-stress-full.ex.D.ci} & held (amended) \\
Instruct, continued adapter & extra loss interval above zero & \nv{rb.ic.it-stress.ex.D} \nv{rb.ic.it-stress.ex.D.ci} & held (floor) \\
Fresh solutions, full fine-tuning & within 5 of once-trained & \nv{rb.rc.it-stress-full.ex.P} & failed \\
Continued adapter, new software & drilled loses at least 5 & \nv{rb.rc.epoch-cont-peft.loss.D} \nv{rb.rc.epoch-cont-peft.loss.D.ci} & held \\
New adapter with output head & drilled loses at most 3 & \nv{rb.rc.epoch-const-head.loss.D} \nv{rb.rc.epoch-const-head.loss.D.ci} & held \\
Once-trained, continued, standard schedule & loses at most 3 & \nv{rb.cs.once-standard.loss} \nv{rb.cs.once-standard.loss.ci} & \nv{rb.co.outcome} \\
\bottomrule
\end{tabular}
\end{table}

\begin{table}[p]
\centering
\caption{Registered predictions on mathematics (confirmatory), one per row.
Threshold: the registered criterion; Observed: the result, in points of forced
accuracy unless marked. Outcomes: held; failed; failed* (in part); rival, not
supported (T2 was registered as the alternative to T1, so at most one could
hold). Claim: break (the drilled model breaks where the once-trained one does
not), cause (repeated text is the cause), teacher (teacher traces), competence
(whether the competence is lost or suppressed), stage (dependence on the later stage), signature ($B$ and
over-sharpening), mechanism (what the later stage does), decision (the decision
to reason), remedy (replay). ``As MP1'': the same criterion as MP1. IT is the
instruction stress test and AO the answer-only stress test; ``a third of the rate''
(R2.1, R4.1) is the gentler stage; excess: extra loss. Codes (Table~\ref{tab:arms}): D, O and P are the
drilled, once-trained and fresh-solution models, D2 the second drilled set, -T marks gpt-oss-120b's
traces, and -u20, -u60 and -u140 give the updates.}
\label{tab:ledger-math}
\scriptsize
\setlength{\tabcolsep}{2.5pt}
\renewcommand{\arraystretch}{1.08}
\begin{tabularx}{\linewidth}{@{}l >{\raggedright\arraybackslash}X >{\raggedright\arraybackslash}p{1.2in} >{\raggedright\arraybackslash}p{1.1in} >{\raggedright\arraybackslash}p{0.5in} >{\raggedright\arraybackslash}p{0.6in}@{}}
\toprule
ID & Prediction & Threshold & Observed & Outcome & Claim \\
\midrule
\multicolumn{6}{@{}l}{\emph{First 9B run}} \\
MP1 & Drilling over-sharpens & $B$(D) $-$ $B$(O) $\geq$ \nv{reg.MP1.thr}\% of the base's NLL, CI $>0$ & \nv{mb.frac.D140}\% against \nv{mb.frac.O140}\% & held & signature \\
MP2 & $B$ rises with passes & D-u20 $<$ D-u60 $<$ D-u140 & \nv{mb.frac.D20}, \nv{mb.frac.D60}, \nv{mb.frac.D140}\% & held & signature \\
MP3 & The drilled model loses more, each stage & more than D-u20 and O, CIs $>0$ & \nv{mex.O.it} (IT), \nv{mex.O.b} (AO) over O & held & break \\
MP5 & Free accuracy falls more than forced & base and O, both stages & all four & held & decision \\
\multicolumn{6}{@{}l}{\emph{Fresh solutions (P)}} \\
M3.1 & P keeps $B$ low, at parity & $B$(P) $\leq$ $B$(D)/2; handoff within \nv{d.parity.math} & $B$ \nv{mm.P.B} against \nv{mb.D140.3} & held & signature \\
M3.2 & Repeated text is the cause & D $-$ P $\geq$ \nv{reg.M3.2.thr}, CIs $>0$ & \nv{mm.DP.it}, \nv{mm.DP.b} & held & cause \\
\multicolumn{6}{@{}l}{\emph{35B}} \\
M1.1 & Drilling over-sharpens & as MP1 & \nv{m35.bfrac.D140}\% against \nv{m35.bfrac.O140}\% & held & signature \\
M1.2 & $B$ rises with passes & as MP2 & rises & held & signature \\
M1.3 & The drilled model loses more & as MP3 & \nv{m35.ex.it}, \nv{m35.ex.b} over O & held & break \\
M1.5 & Free accuracy falls more than forced & as MP5 & not for the base under IT & failed* & decision \\
M1.6 & At least half of $B$ at kept top tokens & at least half & \nv{m35.kept}\% & held & signature \\
M1.7 & On AIME, D loses more after IT (descriptive) & direction & D \nv{m35.aime.h.D140} to \nv{m35.aime.it.D140}, O \nv{m35.aime.h.O140} to \nv{m35.aime.it.O140} & held & break \\
\multicolumn{6}{@{}l}{\emph{Nemotron}} \\
M5.1 & Drilling over-sharpens & as MP1 & \nv{mnem.bfrac.D140}\% against \nv{mnem.bfrac.O140}\% & held & signature \\
M5.2 & $B$ rises with passes & as MP2 & rises & held & signature \\
M5.3 & The drilled model loses more & as MP3 & \nv{mnem.ex.it}, \nv{mnem.ex.b} over O & held & break \\
M5.5 & Free accuracy falls more than forced & as MP5 & not for the start model under AO & failed* & decision \\
M5.6 & At least half of $B$ at kept top tokens & at least half & \nv{mnem.kept}\% & held & signature \\
M5.7 & On AIME, D loses more after IT (descriptive) & direction & D \nv{mnem.aime.h.D140} to \nv{mnem.aime.it.D140}, O \nv{mnem.aime.h.O140} to \nv{mnem.aime.it.O140} & held & break \\
\multicolumn{6}{@{}l}{\emph{Second subset and rank}} \\
M2.1 & A second drilled subset breaks & D2 $-$ O $\geq$ \nv{reg.M2.1.thr}, CIs $>0$ & \nv{mm.D2O.it}, \nv{mm.D2O.b} & held & break \\
M2.2 & It over-sharpens & as MP1 & \nv{mm.D2.bfrac}\% & held & signature \\
N2 & Rank \nv{d.rank.alt} breaks too & at least half of rank \nv{d.rank}'s loss & \nv{mm.R.ratio.it}$\times$, \nv{mm.R.ratio.b}$\times$ & held & break \\
\multicolumn{6}{@{}l}{\emph{Re-forcing}} \\
RF1 & Competence is lost & re-forced excess at least half the first, CIs $>0$ & \nv{rfb.ex.it}, \nv{rfb.ex.b} & held & competence \\
\multicolumn{6}{@{}l}{\emph{Teacher traces}} \\
T1 & Drilled traces break & excess $\geq$ \nv{reg.T1.thr} over O-T and D-T-u20, CIs $>0$ & \nv{mt.ex.O.it}, \nv{mt.ex.O.b} over O-T & held & teacher \\
T2 & Rival: consolidation & excess $<$ \nv{reg.T2.thr}, upper bound $<$ \nv{reg.T2.thr2} & \nv{mt.ex.O.it}, \nv{mt.ex.O.b} & rival, not supported & teacher \\
T4 & Fresh traces fix it & P-T $-$ O-T $\leq$ \nv{reg.T4.thr} & \nv{mt.PO.it}, \nv{mt.PO.b} & held & teacher \\
T5 & Narrowing out of sample: (a) own texts, (b) the base's texts & (a) $\geq$ \nv{reg.T5.thr}$\times$ O-T's; (b) within $\pm$\nv{mt.p5.band} nats & (a) \nv{mt.d.ratio}$\times$; (b) \nv{mt.d.D.base} & failed* & mechanism \\
\multicolumn{6}{@{}l}{\emph{Later stages and what they do}} \\
R1.1 & The stage moves D most on its own texts & $\geq$ \nv{reg.R1.1.thr}$\times$ O's and P's, CI $>0$ & \nv{r1.d.D.u2} against \nv{r1.d.O.u2}, \nv{r1.d.P.u2} nats & held & mechanism \\
R1.2 & Displacement comes before the collapse & D forced at update 2 $>$ \nv{reg.R1.2.thr}\% & \nv{reg.R1.2.obs}\% & held & mechanism \\
R1.3 & Fresh solutions move like once-trained ones & P $\leq$ \nv{reg.R1.3.thr}$\times$ O at updates 1, 2, 5 & \nv{reg.R1.3.obs}$\times$ & held & mechanism \\
R1.4 & Every model fits the new task equally & loss spread $\leq$ \nv{reg.R1.4.thr}\% at updates 2--5 & \nv{reg.R1.4.obs.min}--\nv{r1.lossspread.max}\% & held & mechanism \\
R2.1 & The break survives a third of the rate & excess $\geq$ \nv{reg.R2.1.thr}, CI $>0$ & \nv{r2.ex} & held & stage \\
R3a & The loss persists to update 60 & excess $\geq$ \nv{reg.R3a.thr}, CI $>0$ & \nv{r3.ex60} & held & stage \\
R3b & The stage moves D most on the base's texts & $\geq$ \nv{reg.R3b.thr}$\times$ O's and P's, CI $>0$ & \nv{reg.R3b.obs} nats & failed & mechanism \\
R4.1 & The teacher-trace break survives a third of the rate & excess $\geq$ \nv{r4.req}, CI $>0$ & \nv{r4.ex60} \nv{r4.ex60.ci} & failed & stage \\
R5.1 & The broad mix breaks D & excess CI $>0$ & \nv{r5.ex.D} & held & stage \\
R5.2 & The broad mix breaks D-T & excess CI $>0$ & \nv{r5.ex.DT} & held & stage \\
R5.3 & The broad mix is routine for O and O-T & each loses $\leq$ \nv{r5.req} & \nv{r5.lost.O}, \nv{r5.lost.OT} & held & stage \\
\bottomrule
\end{tabularx}
\end{table}

\begin{table}[htbp]
\centering
\caption{Registered predictions on mathematics, continued: the one-epoch stage (RS),
replay (RP), relearning (RL), sharpening without repetition (SH) and the two further runs
(SD), confirmatory, one per row; O-sharp is the sharpened model; columns, outcomes and claims as in
Table~\ref{tab:ledger-math}; gentler: the gentler stage. SD's secondary prediction, an
extra loss of at least 5 points in each run and stage, also held; secondary
predictions are listed in the records.}
\label{tab:ledger-revision}
\scriptsize
\setlength{\tabcolsep}{2.5pt}
\renewcommand{\arraystretch}{1.08}
\begin{tabularx}{\linewidth}{@{}l >{\raggedright\arraybackslash}X >{\raggedright\arraybackslash}p{1.2in} >{\raggedright\arraybackslash}p{1.1in} >{\raggedright\arraybackslash}p{0.5in} >{\raggedright\arraybackslash}p{0.6in}@{}}
\toprule
ID & Prediction & Threshold & Observed & Outcome & Claim \\
\midrule
RS1 & The one-epoch stage breaks D & excess $\geq$ \nv{reg.RS.D}, CI $>0$ & \nv{rs.ex.D} & held & break \\
RS2 & The one-epoch stage spares P & P $-$ O within $\pm$\nv{reg.RS.P} & \nv{rs.ex.P} & held & cause \\
RP1 & Replay mitigates, own solutions & reduction $\geq$ \nv{reg.RP.mitigate}\% & \nv{rp.cut.D}\%, prevents & held & remedy \\
RL1 & Relearning is fast & $R_5$ of D + IT $\geq$ \nv{reg.RL1.thr} & \nv{rl.D.it.R5} & held & competence \\
RL2 & The ceiling control holds & O + IT within \nv{reg.RL2.thr} at every $k$ & at most \nv{rl.ceil.max} & held & competence \\
SH1 & Sharpening alone does less than half & AO excess of O-sharp $<$ \nv{reg.SH.half} & \nv{sh.ex.b} & held & signature \\
SD1 & D loses more in two new runs & D $-$ O and D $-$ P, CIs $>0$, each run and stage & over O: \nv{sd.s2.ex.it}, \nv{sd.s3.ex.it} (IT); \nv{sd.s2.ex.lr}, \nv{sd.s3.ex.lr} (gentler) & \nv{sd.out.SD1} & break \\
SD3 & P behaves like O in two new runs & P $-$ O within $\pm$3, each run and stage & at most \nv{sd.PO.absmax} & \nv{sd.out.SD3} & cause \\
\bottomrule
\end{tabularx}
\end{table}

\begin{table}[p]
\centering
\caption{Registered predictions on the arithmetic task, an exploratory line; $B$ is over the first \nv{d.tces.cut} tokens,
as registered. Columns, outcomes, IT and AO as in
Table~\ref{tab:ledger-math}; not readable: a registered check failed, or the item
was never evaluated. Claims as in Table~\ref{tab:ledger-math}, plus budget (token
budgets), anchor (the anchored and diluted models), horizon (long-horizon probes) and earlier
study (claims of the earlier replay study, re-examined; its records are in the
repository). Accuracy is forced, in percent. Codes (Table~\ref{tab:arms}): M0 is the starting
model, T the teacher trained by reinforcement learning (RL) and S its clone; U and F are drilled on all of the teacher's samples and
on its correct ones, to \nv{d.pass.high} passes unless marked; U2 is a second unfiltered drilled set;
FA1 and FA025 are the anchored models and FN the diluted one; R-P and R-S are the trace-replay and
solver-trained models of the earlier study (Appendix~\ref{app:earlier}).}
\label{tab:ledger-arith}
\scriptsize
\setlength{\tabcolsep}{2.5pt}
\renewcommand{\arraystretch}{1.08}
\begin{tabularx}{\linewidth}{@{}l >{\raggedright\arraybackslash}X >{\raggedright\arraybackslash}p{1.2in} >{\raggedright\arraybackslash}p{1.1in} >{\raggedright\arraybackslash}p{0.5in} >{\raggedright\arraybackslash}p{0.6in}@{}}
\toprule
ID & Prediction & Threshold & Observed & Outcome & Claim \\
\midrule
\multicolumn{6}{@{}l}{\emph{Depth profile and cap test}} \\
DP1 & Repetition, not training amount, moves the reasoning & $B$(S) below $B$(U) and $B$(F), CIs $<0$ & \nv{b.S} against \nv{b.U140}, \nv{b.F140} & held & signature \\
DP2 & Every lasting $B$ below every drilled $B$ & separation & highest lasting \nv{b.T}, lowest drilled \nv{b.RP} & held & signature \\
DP3 & $B$ rises with passes in U and F & monotone & monotone & held & signature \\
DP4 & $B$ ranks the loss & Spearman $\geq$ \nv{reg.DP4.thr} & \nv{rho.B} & held & signature \\
CT1 & Competence lost, not slowed & U, F $<$ \nv{reg.CT1.thr}\% within \nv{d.budget.long} tokens & \nv{tces16.U140.u20}, \nv{tces16.F140.u20} & held & break \\
CT2 & A longer budget adds little for S & gain $\leq$ \nv{reg.CT2.thr} over \nv{d.budget.short} tokens & \nv{tces16w4.S.u20} to \nv{tces16.S.u20} & failed & budget \\
\multicolumn{6}{@{}l}{\emph{Anchor (exploratory)}} \\
AN1 & The anchor protects & FA1 $\geq$ \nv{reg.AN1.thr}\% after AO, if the checks pass & \nv{anc.a16.FA1}; parity check failed & not readable & anchor \\
AN2 & Dose & FA1 $>$ FA025 $>$ F after AO; $B$ in reverse & \nv{anc.a16.FA1}, \nv{anc.a16.FA025}, \nv{tces16.F140.u20} & held & anchor \\
AN3 & Replication on a second subset & U2 $\leq$ \nv{reg.AN3.thr}\%, $B \geq$ \nv{reg.AN3.thr2} & \nv{anc.a16.U2}\%, $B$ \nv{anc.b.U2} & held & break \\
AN4 & Dilution does not protect & FN $\leq$ \nv{reg.AN4.thr}\%, if the checks pass & drilling check failed & not readable & anchor \\
\multicolumn{6}{@{}l}{\emph{Decision and competence}} \\
DC1 & The RL teacher skips the search later than trace checkpoints & later than R-P at \nv{d.pass.low}, \nv{d.pass.mid} and \nv{d.pass.high} passes & later than R-P at \nv{d.pass.mid} and \nv{d.pass.high} passes, not at \nv{d.pass.low} & failed* & decision \\
DC3 & Forced, trace and RL checkpoints stay above their free accuracy and M0's forced & every such origin & not R-P-u140 & failed* & decision \\
DC4 & Forcing does not lift the solver above M0 & R-S $\leq$ M0 forced after the flip & never evaluated: R-S never flips & not readable & decision \\
C-1 & Forced accuracy holds through update 80 & every origin & T-u30 falls to \nv{tces4.T.u80} & failed & horizon \\
C-2 & Trained checkpoints keep their margin over M0 & every origin & T-u30 \nv{tces4.T.u80}, M0 \nv{tces4.M0.u80} at update 80 & failed & horizon \\
C-3 & Near zero by update 480, halving at least \nv{reg.C-3.thr}$\times$ later than the half-skip update & every origin & R-P-u140 halves at \nv{tces.halves.RP}, half-skip at \nv{tces.hs.RP} & failed* & horizon \\
E1 & The coverage gap is a choice of mode & solver's never-solved count near the trace's & \nv{e1.RS.never} and \nv{e1.RP.never} (M0 \nv{e1.M0.never}) & held & earlier study \\
E2 & Forced on held-out families, trained checkpoints near M0 & close & \nv{e2.S}--\nv{e2.RS} against \nv{e2.M0} & held & earlier study \\
C2 & The clone beats the teacher at update 80 in a rerun & by more than \nv{reg.C2.thr} & \nv{tces4.rerun.S} against \nv{tces4.rerun.T} & failed & horizon \\
C3 & The decision clocks repeat & half-skip in updates \nv{reg.C3.thr}, within \nv{reg.C3.thr2} of the first run & \nv{tces.hs.rep.T}, \nv{tces.hs.rep.S} & held & decision \\
\multicolumn{6}{@{}l}{\emph{Repetition}} \\
RR1 & Drilled U and F stop searching sooner than S & both arms & U only & failed* & decision \\
RR-D & Competence lost at \nv{d.pass.high} passes, kept at \nv{d.pass.low} & most lost; kept & \nv{tces4.dose.U.high}, \nv{tces4.dose.F.high}; at \nv{d.pass.low}: \nv{tces4.dose.U.low}, \nv{tces4.dose.F.low} & held & break \\
RR-D2 & The same at a third of the learning rate & below half of handoff & U \nv{tces4g.U140} from \nv{tces4.U140.u0}, F \nv{tces4g.F140} from \nv{tces4.F140.u0} & failed & stage \\
RR-D3 & The same under IT & below half of handoff & U \nv{tces4i.U140}, F \nv{tces4i.F140} & failed & stage \\
RR-D3b & Under IT, free accuracy shows it & drilled lose $\geq$ \nv{reg.RR-D3b.thr}, states at \nv{d.pass.low} passes $\leq$ \nv{reg.RR-D3b.thr2} & states at \nv{d.pass.low} passes lose \nv{reg.RR-D3b.obs} & failed & decision \\
RR-D4 & A cliff: most of the loss by update 2 & F $<$ \nv{reg.RR-D4.thr}\% at update 2 & \nv{tces4.cliff.F140} & failed & mechanism \\
\bottomrule
\end{tabularx}
\end{table}

\paragraph{What held and what failed.} The predictions behind the headline claims
held. They cover the break in every cell (MP3, M1.3, M5.3, M2.1, N2, RS1, SD1),
its cause (M3.2, RS2, SD3), the teacher traces (T1, T4), the dependence on the
later stage (R2.1, R3a, R5.1--R5.3), relearning (RL1, RL2), replay (RP1) and
sharpening (SH1). One prediction held but is not the paper's reading. RF1 predicted
that the drilled model's competence is lost, and it held as registered; but the
relearning experiment, registered later, shows that the competence is suppressed
rather than lost, and the paper follows that reading. The failures that qualify
the claims (M1.5, M5.5, R3b, T5(b) and R4.1) are stated where the claims are made.
Registered conditions, alternative readings, descriptive items and secondary
predictions are listed in the records.

\paragraph{Analyses that were not registered.} The paper also reports the
following analyses, which were not registered and are given as observations:
\begin{itemize}
\item the weight-change measure and its split of the models
(Section~\ref{sec:signature});
\item the movement of the weights under a new adapter and, as a mean over
matrices, under continued adapters (Section~\ref{sec:signature});
\item the split of that movement at the output head, its direction and the
adapters' factor norms without the output head (Appendix~\ref{app:released});
\item the comparison of the fresh-solution with the once-trained model in the
answer-only stress test with a new adapter (Section~\ref{sec:released});
\item the test-time probe that forbids ending the turn (the end-of-turn mask),
scored by the first boxed answer (Appendix~\ref{app:masking});
\item the reading of the instruct student after the untrained student's run, and
its free accuracy after the later stages (Appendix~\ref{app:released});
\item the fresh-solution model's lead over the drilled one on the synthetic skill
(Appendix~\ref{app:skill});
\item the 35B base's collapse in the instruction stress test, which it starts from
a new adapter (Appendix~\ref{app:scale});
\item the drop of the drilled model's own texts below the base's likelihood, its
movement toward the base's reasoning and the broad mix's training losses
(Appendices~\ref{app:process} and~\ref{app:stage});
\item the synthetic skill's extra loss from handoff and its losses as shares of
handoff accuracy (Appendix~\ref{app:skill});
\item the forms the failures take, the AIME scores and the strict scores
(Appendices~\ref{app:mathfail}, \ref{app:aime} and~\ref{app:strict});
\item the loss jump, the rise in the drilled checkpoints' loss on the next batch one update into the arithmetic task's answer-only stress test (Appendix~\ref{app:profile});
\item the anchored models, which are exploratory (Appendix~\ref{app:anchor}).
\end{itemize}